\documentclass{article}
\usepackage{arxiv}

\usepackage[utf8]{inputenc} % allow utf-8 input
\usepackage[T1]{fontenc}    % use 8-bit T1 fonts
\usepackage{hyperref}       % hyperlinks
\usepackage{url}            % simple URL typesetting
\usepackage{booktabs}       % professional-quality tables
\usepackage{nicefrac}       % compact symbols for 1/2, etc.

\usepackage{microtype}      % microtypography
\usepackage{lipsum}		% Can be removed after putting your text content
\usepackage{graphicx}
\usepackage{doi}
\usepackage{endnotes}
\let\footnote=\endnote

\def\E{{\mathbb E}}

\def\argmin{\mathop{\rm arg\,min}}%
\newtheorem{theorem}{Theorem}
\newtheorem{lemma}{Lemma}
\newtheorem{proposition}{Proposition}
\newtheorem{corollary}{Corollary}

\newtheorem{assumption}{Assumption}

\newtheorem{definition}{Definition}

\usepackage{colortbl}
\usepackage{comment}
\usepackage{graphicx}
\usepackage{bbm}
\usepackage{subcaption}
\usepackage[title]{appendix}
\usepackage{amsmath}
\usepackage{amsfonts}
\usepackage{amssymb}
\usepackage{multirow}
\usepackage{cleveref}
\usepackage{wrapfig}
\usepackage{cases}

\usepackage[ruled]{algorithm2e}

\usepackage{natbib}
 \bibpunct[, ]{(}{)}{,}{a}{}{,}%

\title{Policy Optimization in Dynamic Bayesian Network Hybrid Models of Biomanufacturing Processes}

\date{} 					% Or removing it

\author{Hua Zheng, Wei Xie\thanks{Corresponding author. Email: w.xie@northeastern.edu} \\
Department of Mechanical and Industrial Engineering \\
Northeastern University\\
Boston, MA 02115\\
	\And
	Ilya O. Ryzhov \\
Robert H. Smith School of Business\\
University of Maryland \\
College Park, MD 20742\\
	\AND
	Dongming Xie \\
Department of Chemical Engineering\\
University of Massachusetts Lowell\\
Lowell, MA 01854\\
}

\renewcommand{\shorttitle}{\textbf{Zheng et al.:} \textit{Policy Optimization in Dynamic  Bayesian Networks for Biomanufacturing}}

\begin{document}

\maketitle

\begin{abstract}
Biopharmaceutical manufacturing is a rapidly growing industry with impact in virtually all branches of medicines. Biomanufacturing processes require close monitoring and control, in the presence of complex bioprocess dynamics with many interdependent factors, as well as extremely limited data due to the high cost of experiments as well as the novelty of personalized bio-drugs. We develop a novel model-based reinforcement learning framework that can achieve human-level control in low-data environments.
The model uses a dynamic Bayesian network to capture causal interdependencies between factors and predict how the effects of different inputs propagate through the pathways of the bioprocess mechanisms. This enables the design of process control policies that are both interpretable and robust against model risk. We present a computationally efficient, provably convergence stochastic gradient method for optimizing such policies. Validation is conducted on a realistic application with a multi-dimensional, continuous state variable.
%overcome the key challenges of biopharmaceutical %manufacturing, i.e., high complexity, high uncertainty, and %very limited process data.
% IOR: Limited data were already mentioned earlier in this paragraph. There is no need to repeat the same phrase. 
\end{abstract}

%\KEYWORDS{deterministic inventory theory; infinite linear programming duality; 
%  existence of optimal policies; semi-Markov decision process; cyclic schedule}


%%%%%%%%%%%%%%%%%%%%%%%%%%%%%%%%%%%%%%%%%%%%%%%%%%%%%%%%%%%%%%%%%%%%%%

% Samples of sectioning (and labeling) in IJOC
% NOTE: (1) \section and \subsection do NOT end with a period
%       (2) \subsubsection and lower need end punctuation
%       (3) capitalization is as shown (title style).
%
%\section{Introduction.}\label{intro} %%1.
%\subsection{Duality and the Classical EOQ Problem.}\label{class-EOQ} %% 1.1.
%\subsection{Outline.}\label{outline1} %% 1.2.
%\subsubsection{Cyclic Schedules for the General Deterministic SMDP.}
%  \label{cyclic-schedules} %% 1.2.1
%\section{Problem Description.}\label{problemdescription} %% 2.

% Text of your paper here

\section{Introduction}

This work is motivated by the problem of process control in biomanufacturing, a rapidly growing industry { valued at about \$300 billion in 2019 \citep{martin2021brief}}. Over 40\% of the products in the pharmaceutical industry's development pipeline are bio-drugs \citep{PharmaAnnualReview2019}, designed for prevention and treatment of {cancers, autoimmune disorders, and infectious diseases \citep{tsopanoglou2021moving}.}
These drug substances are manufactured in living {organisms (e.g., cells) % and yeasts)
} 
% {(or other living organisms, such as bacteria and yeast)}
whose biological processes are complex and highly variable. Furthermore, a typical biomanufacturing %production 
process consists of numerous unit operations, e.g., cell culture, purification, and formulation, each of which directly impacts the outcomes of successive steps. Thus, in order to improve productivity and ensure drug quality, the process must be controlled as a whole, from end to end.

{Historical process observations are often very limited, due to the long analytical testing times for biopharmaceuticals with complex molecular structure 
\citep{%jiang2017opportunities, 
hong2018challenges}. In this field, it is very common to work with 3-20 process observations \citep{OBrien_2021}. Moreover, emerging biotherapeutics (e.g., cell and gene therapies) are becoming more and more personalized. Every protein therapy is unique, which often forces R\&D efforts to work with just 3-5 batches; see for example \cite{martagan2016optimal,martagan2018performance}.}
%In this industry, production processes are very complex, and experimental data are very limited, because analytical testing times for complex biopharmaceuticals are very long. Unfortunately, ``big data'' do not exist in this industry: in a typical application, a process controller may have to make decisions based on 10 or fewer prior experiments. 
Additionally, human error is frequent in biomanufacturing, accounting for 80\% of deviations \citep{Ci15}. In this paper, we propose an optimization framework, based on reinforcement learning (RL), which demonstrably improves process control in the small-data setting. Our approach incorporates domain knowledge of the underlying process mechanisms, as well as experimental data, to provide effective control policies that outperform both existing analytical methods as well as human domain experts.

As discussed in an overview by \cite{hong2018challenges}, biomanufacturing has traditionally relied on deterministic kinetic models, based on ordinary or partial differential equations (ODEs/PDEs), to represent the dynamics of bioprocesses. Classic control approaches for such models, such as feed-forward, feedback, and proportional-integral-derivative {(PID)} control, tend to focus on the short-term impact of actions while ignoring their long-term effects. They also do not distinguish between different sources of uncertainty, and are vulnerable to model misspecification. Furthermore, while these physics-based models have been experimentally shown to be valuable in %various types of 
biomanufacturing processes, %\citep{kyriakopoulos2018kinetic,lu2015control}, 
an appropriate model may simply not be available when dealing with a new biodrug.

These limitations have led to recent interest in models based on Markov decision processes or MDPs \citep{%peroni2005optimal,
liu2013modelling,martagan2018performance}, as well as more sophisticated deep RL approaches \citep{spielberg2017deep,Treloar2020deep}. In a sense, however, these techniques go too far in the opposite direction: they fail to incorporate domain knowledge, and therefore require much greater volumes of %training 
{historical data for fitting input-output relations} before they can learn a useful control policy. Additionally, they have limited interpretability, and their ability to handle uncertainty is also limited because they do not consider \textit{model risk}, or error introduced into the policy by misspecification of {the underlying stochastic process }model.
%resulting from using a small volume of data for calibration. 
Finally, existing studies of this type are limited to individual unit operations and do not consider the entire process from end to end.

% IOR: There is an issue with using the abbreviation KG since there are currently many more simulation papers where it means "knowledge gradient." Those papers are also Bayesian, so the phrase "Bayesian KG" is best avoided in my opinion.

Our proposed framework takes the best from both worlds. Unlike existing papers on RL for process control, which use %general-purpose general 
{black-box}
techniques such as neural networks, %our approach is model-based:
% IOR: No one will know what a "hybrid model" is. A hybrid of what with what? This has to be explicitly explained.
we explicitly represent causal interactions within and between different unit operations, using a \textit{probabilistic knowledge graph}. This is a \textit{hybrid} model, in the sense that it uses both real-world data as well as structural information from existing kinetic models (domain experts often know which factors influence each other, even if the precise effects are difficult to measure). We use Bayesian statistics to separately quantify model uncertainty (introduced by misspecification) and stochastic uncertainty (variability inherent in the system). 
%The graph structure of the resulting dynamic Bayesian %network (DBN) incorporates information 
%from existing biological, physical, or chemical models: domain experts often know 
%on which factors influence each other, and the issue is to %precisely measure these effects (the weights of certain %edges).
Thus, the graph structure of the resulting dynamic Bayesian network (DBN) describes the dynamics, interactions, and stochastic variation within the biomanufacturing process. {This hybrid model allows us to predict how the effects of different inputs propagate through the pathways of the bioprocess mechanism. We show how these effects can be quantified and ranked using a Shapley value analysis, providing interpretable insight into the relative importance of different inputs.}

%The RL problem can be viewed as a search for a policy that optimally adjusts the inputs to different unit operations based on their dynamically evolving states.

% IOR: Please do not repeat long phrases like "model-based RL scheme on the Bayesian KG". Abbreviations do not make them more manageable. Also, up to this point we haven't described RL, only the knowledge graph, so the reader doesn't know what "the proposed model-based RL scheme" is. %key challenges of biopharmaceutical manufacturing, i.e., %high complexity, high uncertainty, and very limited process %data.}
% IOR: We already talked about these challenges. There is no need to repeat them here.
%While some of the edge weights in the graph represent the effects of various chemical or physical reactions intrinsic to the biomanufacturing process, others model decisions made in different unit operations (for example, the nutrient feeding strategy for working cells) and thus are directly controllable. The RL problem can then be viewed as a search for a policy that optimally adjusts these weights based on the dynamic state of each unit operation. 

{Our main goal is to design effective process control policies that are robust against model risk.} We conduct policy search using a novel projected stochastic gradient method which can save computational effort by reusing computations associated with similar input-output mechanism pathways. We prove the convergence of this method to a local optimum (the objective is highly non-convex) in the space of policies and give the convergence rate. {We validate our approach in a realistic case study for fed-batch production %fermentation 
of \textit{Yarrowia lipolytica}, a yeast with numerous biotechnological uses \citep{BaKuZi09}. The results indicate that DBN-RL can achieve human-level process control much more data-efficiently than a state-of-the-art model-free RL method.}
%As is typical in the biomanufacturing domain, only eight prior experiments were conducted, and these are the only data available for calibration and optimization of the production process. Nonetheless, by combining these data with information from existing kinetic models, our proposed DBN-RL framework achieves human-level process control with a fairly small number of training iterations, and outperforms the human experts when allowed to run longer. Our approach is far more data-efficient than a state-of-the-art model-free RL method.

In sum, our work makes the following contributions. 1) We propose a Bayesian knowledge graph hybrid model which overcomes the limitations of existing process models by explicitly capturing causal interdependencies between factors in the underlying multi-stage stochastic decision process. 2) We develop a model-based RL scheme on the Bayesian knowledge graph to find effective and robust process control policies, thus mitigating much of the challenge posed by limited process data. 3) We develop an efficient implementation of the RL procedure which exploits the graph structure to reuse computations associated with similar input-output pathways.
4) We demonstrate the efficacy of our approach against both human experts and a state-of-the-art benchmark, in a case application with real biomanufacturing data. The results show that DBN-RL can be very effective even with a very small amount of process data. {The source code and data needed to reproduce our empirical results are available at: \url{https://github.com/zhenghuazx/Policy-Optimization-in-Dynamic-Bayesian-Network}}.

% IOR: Putting Shapley value in the list of contributions looks very strange. This was not discussed anywhere in the introduction at all. Instead, I mentioned it earlier where the RL problem was introduced.

% --------------------------------------------------

\section{Literature Review}
\label{sec:literatureReview}

Traditionally, biomanufacturing has used physics-based, first-principles models (so-called ``kinetic'' or ``mechanistic'' models) of bioprocess dynamics {\citep{mandenius2013measurement}}. One example is the work by {\cite{liu2013modelling}}, which uses an ODE model of cell growth in a fed-batch fermentation process. However, not all bioprocesses are equally well-understood, and first-principles models may simply not be available for certain quality attributes or unit operations \citep{rathore2011chemometrics}. In certain complex operations, such models may oversimplify the process and make a poor fit to real production data \citep{teixeira2007hybrid}.

For this reason, domain experts are increasingly adopting data-driven methods.
% for process understanding \citep{mercier2013multivariate,kirdar2007application}, monitoring \citep{teixeira2009situ} and control \citep{martagan2018performance}.
Such methods are often used when prediction is the main goal: for example, \cite{gunther2009process} aim to predict quality attributes at different stages of a biomanufacturing process using such statistical tools as partial least squares. Statistical techniques can also be used in conjunction with first-principles models, as in \cite{lu2015control}, which uses design of experiments to obtain data for a system of ODEs. The main drawback of purely data-driven methods is that they are not easily interpretable, and fail to draw out causal interdependencies between different input factors and process attributes. %\cite{hong2018challenges} gives examples of certain attributes that are known to indicate product quality, but can neither be measured directly nor accurately predicted by existing statistical models.

Process control %, as the name indicates, 
seeks not only to monitor or predict the process, but to maintain various process parameters within certain acceptable levels in order to guarantee product quality \citep{jiang2016integrated}. Researchers and practitioners have used various standard techniques such as feed-forward, feedback-proportional, and model predictive control \citep{hong2018challenges}. The work by \cite{lakerveld2013model} shows how such strategies may be deployed at various stages of the production processes and evaluated hierarchically for end-to-end control of an entire pharmaceutical plant. These strategies, however, are usually derived from deterministic first-principles models, and do not consider either stochastic uncertainty or process model risk. {As a result, hybrid models are becoming increasingly popular, as they overcome the limitations of purely statistical and mechanistic models \citep{tsopanoglou2021moving}. For example, \cite{zhang2019hybrid} proposed a hybrid modeling and control framework, using pre-trained kinetic models, to generate high quality input to a data-driven model, which is capable of simulating process dynamics over a broad operational range as well as optimizing control in the presence of uncertainty.}
%by exploiting stochastic optimization.}

%Recent work by \cite{martagan2018performance} has sought to address this issue using Markov decision processes, which consider uncertainty to an extent and also allow deeper structural analysis of the optimal control policy. However, these MDP models suffer from the curse of dimensionality (real bioprocesses typically have multidimensional, continuous state and action spaces), and thus focus on a single unit operation under a simplified process model, which exacerbates model risk.

% \textcolor{blue}{Reviewer: The discussion of Martagan et al. (2017) is too generic. The curse of dimensionality is ubiquitous with MDP models. In fact, it also applies to the presented work. The key differences are that Martagan et al. (2017) deals (i) with discrete actions and (ii) not with a process control problem of continuous CPPs."}

Reinforcement learning (RL) has been shown to attain human-level control in challenging problems in healthcare \citep{zheng2021personalized}, competitive games \citep{silver2016mastering} and other applications. However, these successes were made possible by the availability of large amounts of training data, and the control tasks themselves were well-defined and conducted in a fixed environment. Unfortunately, none of these factors holds in biomanufacturing process control. Much of the recent work on RL for this domain  \citep{spielberg2017deep,Treloar2020deep,petsagkourakis2020reinforcement}, requires large training datasets. {\cite{pandian2018control} avoids this problem by using an artificial neural network to learn an initial approximate control policy, then refining that policy with RL. This approach showed good performance, but was developed for a relatively low-dimensional setting in which the policy is stationary (with respect to time).} In any case, general-purpose RL techniques, such as neural networks, make it difficult to obtain interpretable insight from the results. Furthermore, a substantial part of the process dynamics does not need to be guessed from the data, but can be structured according to first-principles models.

%, but strictly speaking should not be \textit{necessary}, because a substantial part of the process dynamics is structured (e.g., according to first-principles models), and one can simply build this structure into the model without having to try to guess it from data.

Our approach captures the strengths of both first-principles and data-driven models through a DBN hybrid model, which combines both expert knowledge and data.
The applicability of Bayesian networks to biomanufacturing was first investigated by \cite{xie2020bayesian}, which used such a model to capture causal interactions between {and within} various unit operations. 
%We build on this work in the present paper. %, but the model we develop here is much more powerful: first, \cite{xie2020bayesian} infers the interactions from data, whereas we show how to use first-principles models to create a prior; second, 
However, the study in \cite{xie2020bayesian} only sought to describe the process, and did not consider the control problem to any extent. The core contributions of the present paper are the introduction of a control policy into the Bayesian network, and the optimization of this policy using gradient ascent. These developments require considerable new developments in modeling, computation, and theory, all of which are new to this paper.

% \textcolor{blue}{Reviewer: ``process control literature, especially on similar approaches that combine kinetic models with data-driven methods"}

% IOR: Since the content has been rewritten, I think this title is the most accurate description of the section.
\section{{Modeling Bioprocess Dynamics, Rewards, and Policies}}
%Dynamic Bayesian Network based Probabilistic Knowledge Graph Modeling Biomanufacturing Process Mechanisms}
\label{sec:hybridModeling}

{Section~\ref{sec:lineargaussian} presents %how one can create
a dynamic Bayesian network (DBN) model for biomanufacturing stochastic decision processes.}
%Section \ref{subsec:hybridModel} shows how the network structure and dynamics can be extracted from domain-specific kinetic models (when such models are available).
%Section \ref{sec:lineargaussian} shows how the information obtained from kinetic models can be represented using linear Gaussian dynamics, and further improved using Bayesian updating with experimental data. 
Section~\ref{sec:rewardspolicies} augments this model with a linear control policy which is evaluated using a linear reward function. Section~\ref{sec:structprop} presents structural results that will be used in later algorithmic developments, {and Section~\ref{eq.SV} provides an interpretive mechanism which uses the Shapley value to quantify and rank the influence and propagation of different inputs through the bioprocess pathways.}
%optimal control policies that are interpretable, and robust against both inherent stochasticity and model uncertainty.} 

%For a complete understanding of the model, some technical terms and details from the biomanufacturing domain will be required. For the sake of readability, we have made an effort to present only the most essential biomanufacturing concepts, while emphasizing the modeling innovations.

%\subsection{Extracting Graph Structure From Kinetic Models}\label{subsec:hybridModel}

\subsection{{Dynamic Bayesian Network (DBN)
%with Linear Gaussian Dynamics
}}
\label{sec:lineargaussian}

A typical biomanufacturing system consists of distinct unit operations, including upstream fermentation for drug substance production and downstream purification to meet quality requirements; see \cite{Pauline_2013} for more details. 
%which take place sequentially. 
The final output metrics of the process (e.g., drug quality, productivity) are impacted by many interacting factors. In general, these factors can be divided into
critical process parameters (CPPs) and critical quality attributes (CQAs). Precise definitions can be found in ICH Q8(R2) \citep{guideline2009pharmaceutical}. For the purpose of this discussion, one should think of CQAs as the ``state'' of the process, including, e.g., the concentration of biomass and/or impurity at a point in time. CPPs should be viewed as ``actions'' that can be monitored and controlled to influence CQAs and thus to optimize the output metrics. For example, acidity of the solution, temperature, and feed rate are all controllable and thus may be included among the CPPs. We use $\pmb{s}_t=\left(s^1_t,...,s^n_t\right)$ and $\pmb{a}_t =\left(a^1_t,...,a^m_t\right)$ to denote, respectively, the state and action vectors at time $t$ (thus, there are $m$ CPPs and $n$ CQAs).

Denote by $s_t^k$ (respectively, $a^k_t$) the $k$th component of the state (decision) variable at time $t$. %In the language of Markov decision processes, $s^k_t$ is the $k$th dimension of the state variable, while $a^k_t$ is the $k$th dimension of the action or decision variable. %We suppose that there are $n$ CQAs and $m$ CPPs in each time period 
%Suppose the dimensions $|\pmb{s}_t|=n$ and $|\pmb{a}_t|=m$
The dimensions $n$ and $m$ of the state and action variables may be time-dependent, but for notational simplicity we keep them constant in our presentation. Suppose there are $H$ time periods in all.
Let $s_{1}^k \sim \mathcal{N}(\mu^k_{1},(v^k_{1})^2)$
% with $\pmb{s}_1$ 
model variation in the initial state (e.g., raw material variation).  We model $a_{t}^k \sim \mathcal{N}(\lambda^k_{t},(\sigma^k_{t})^2)$ for each CPP $a^k_t$. Though $\pmb{a}_t$ is controllable in practice, we model it as a random variable for two main reasons.  First, the kinetic models used in biomanufacturing (which we use to design the dynamic Bayesian network) focus on modeling bioprocess % environment and 
dynamics, rather than control. Second, in practice, the specifications of the production process (which ensure that the final product meets quality requirements) treat CPPs as ranges of values, within which variation is possible. The problem of actually choosing a control policy will be revisited in Section \ref{sec:rewardspolicies}.

In general, the input-output relationship at time step $t$ can be modeled as $\pmb{s}_{t+1}=f(\pmb{s}_t,\pmb{a}_t) + \pmb{e}_{t+1}$, where the {residual $\pmb{e}_{t+1}$ is a random variable representing the impact of many uncontrollable factors. We assume that it follows a Gaussian distribution, i.e., 
$\pmb{e}_{t+1}\sim \mathcal{N}(0, V_{t+1}^2)$ by applying the central limit theorem, where $V_{t+1}\triangleq\mbox{diag}(v_{t+1}^{k})$ is a diagonal matrix.} 
The function $f$ is obtained from a mechanistic model based on domain knowledge of biological, physical, and chemical mechanisms. In this paper, we assume that $f$ is linear; further down, we will discuss why this can be a reasonable assumption on small time scales. Denote by $\pmb\beta_{t}^s$ the $n\times n$ matrix whose $\left(j,k\right)$th element is the linear coefficient $\beta^{jk}_t$ %in (\ref{eq.CQA})
corresponding to the effect of state $s^j_t$ on the next state $s^k_{t+1}$. Similarly, let $\pmb\beta_t^a$ be the $m\times n$ matrix of analogous coefficients representing the effects of each component of $\pmb{a}_t$ on each component of $\pmb{s}_{t+1}$. 
Let $\pmb{e}_{t+1}=V_{t+1}\pmb{z}_{t+1}$, where $\pmb{z}_{t+1}$ is an $n$-dimensional standard normal random vector. 
%and $V_{t+1}\triangleq\mbox{diag}(v_{t+1}^{k})$ is a diagonal matrix of residual {standard deviations}. 
Then, the stochastic process dynamics can be written as
\begin{equation}
    \pmb{s}_{t+1} = \pmb{\mu}_{t+1}^s + \left(\pmb{\beta}_{t}^s\right)^\top(\pmb{s}_t-\pmb\mu_t^s) + \left(\pmb\beta_{t}^a\right)^\top(\pmb{a}_t-\pmb\mu_t^a) +V_{t+1}\pmb{z}_{t+1},
    \label{eq: LGBN matrix form}
\end{equation}
where $\pmb\mu_{t}^s=(\mu_t^{1},\ldots, \mu_t^n)$ and $\pmb\mu_{t}^a=(\lambda_t^{1},\ldots, \lambda_t^m)$. Letting $\pmb\sigma_t=(\sigma_t^1,\ldots,\sigma_t^m)$ and $\pmb{v}_t=(v_t^1,\ldots,v_t^n)$, the list of parameters for the entire model can be denoted by $\pmb{w} = (\pmb{\mu}^s,\pmb\mu^a,\pmb{\beta},\pmb\sigma,\pmb{v})= \{(\pmb{\mu}_{t}^s,\pmb\mu_t^a,\pmb{\beta}_{t}^s,\pmb\beta_{t}^a,\pmb\sigma_t,\pmb{v}_t)| 0\leq t\leq H\}$, where $\pmb\beta = (\pmb\beta^a,\pmb\beta^s)$. The model parameters $\pmb{w}$ are unknown, but can be estimated from data.

\begin{comment}
It is easy to see that (\ref{eq: LGBN matrix form}) has the same form as (\ref{eq: linearization}) if we let
%. Indeed, if we let
\begin{eqnarray*}
\pmb\mu_{t+1}^s &=& \pmb{\mu}_{t}^s + \Delta t\cdot \pmb{f}(\pmb\mu_t^s,\pmb\mu_t^a),\\
\pmb\beta_{t}^s &=& \Delta t \cdot J_f(\pmb\mu_t^s)+1,\\
\pmb\beta_{t}^a &=& \Delta t \cdot J_f(\pmb\mu_t^a),
\end{eqnarray*}
and treat $R_{t+1}$ as the residual. 
\end{comment}

%This linear state transition model in (\ref{eq: LGBN matrix form}) is valid if we have more frequent CPPs/CQAs measurements than the dynamics of bioprocesses. 
% IOR: This sentence does not make sense. I think what you mean is that CPPs/CQAs are measured on a faster time scale.

The linear state transition model in (\ref{eq: LGBN matrix form}) is valid if CPPs/CQAs are monitored on a faster time scale than the evolution of the bioprocess dynamics. This is often the case for heavily instrumented biomanufacturing processes, where online sensor monitoring technologies are used to facilitate real-time process control. For example, %Raman spectroscopy optical sensor measures the amount of light scattered inelastically at different frequencies by molecular vibrations, which results in a linear combination of individual molecular fingerprints (aka single-component spectra) with high chemical specificity. 
  an in-situ Raman spectroscopy optical sensor can monitor cell culture processes, % parameters simultaneously and continuously,
  including cell nutrients, metabolites, and cell density
  %(e.g., glucose, lactate, ammonium, viable cell density, and whole cell density) 
  at intervals of 6 minutes. Bioprocess dynamics are typically much slower (on the order of hours); see for example \cite{craven2014glucose}.

The following simple example illustrates how known bioprocess mechanisms can be incorporated into (\ref{eq: LGBN matrix form}). Consider a simple exponential cell growth mechanism \citep{Pauline_2013}, given by $x_t=x_0e^{\mu t}$, where $x_t$ denotes the cell density at time $t$ and $\mu$ is the unknown growth rate. 
%This is a commonly used mechanism model for cell growth in biomanufacturing industry; see more information in \cite{Pauline_2013}.
%Suppose that there is a fixed cell culture duration $t$. 
Using a log transformation $s_t=\log(x_t)$, we arrive at a hybrid probabilistic model, % for the exponential growth phase in fermentation or cell culture process, 
$s_{t+1} = \mu \Delta t +s_t + e_t$, where %$\beta_0= \mu t$ and 
$\Delta t$ represents the time step for each period, and the residual term $e_t$ characterizes the combined effect from other factors. In this way, given any ODE-based
%and it follows a Gaussian distribution by following central limit theorem (CLT).
mechanistic model $d \pmb{s}_t/d t = \pmb{f}(\pmb{s}_t,\pmb{a}_t)$, we can construct a linear hybrid probabilistic model by applying a first-order Taylor approximation to the function $\pmb{f}(\pmb{s}_t,\pmb{a}_t)$; see Appendix~\ref{sec:TaylorApproximation} for a more detailed exposition of this technique. The approximation error becomes negligible in the setting of online monitoring.

Then, the distribution of the entire trajectory $\pmb\tau=(\pmb{s}_1,\pmb{a}_1,\pmb{s}_2,\pmb{a}_2,\ldots,\pmb{s}_{H})$ of the stochastic decision process (SDP) can be written as a product
\begin{eqnarray*}
p(\pmb{\tau}) = p(\pmb{s}_1) 
\prod_{t=1}^{H-1} p(\pmb{s}_{t+1}|\pmb{s}_t,\pmb{a}_t) p(\pmb{a}_t)
%p(\pmb\tau)= \prod_{t=1}^{H}\left[\prod_{k=1}^{m}\mathcal{N}(\lambda^k_{t},(\sigma^k_{t})^2)\prod_{k=1}^{n}\mathcal{N}\left(\mu^k_{t+1} + \sum_{X^j_t\in Pa(s^k_{t+1})}\beta^{jk}_{t}(X^j_t - \mu^j_t),(v^k_{t+1})^2\right)\right]
    \label{eq: dbn joint distritbuion}
\end{eqnarray*}
of conditional distributions. Given a set of real-world observations denoted by $\mathcal{D} = \left\{\pmb\tau^{(n)}\right\}^R_{n=1}$, we quantify the model parameter estimation uncertainty using the posterior distribution $p(\pmb{w}|\mathcal{D})$. 
%The posterior is not expressible in closed form, but it is possible to generate samples from it using the method of Gibbs sampling \citep{Ge00}.
While this distribution is not easy to compute, we can generate samples from it using the method of Gibbs sampling \citep{Ge00}.
Our implementation of this technique closely follows the study in \cite{xie2020bayesian}, so we do not devote space to it here; for completeness, a brief description is given in Appendix \ref{sec:gibbs}.

\subsection{DBN Augmented with Linear Rewards and Policies}
\label{sec:rewardspolicies}

The process trajectory $\pmb\tau$ is evaluated in terms of revenue {depending on productivity, product CQAs,} 
{and}
%(based on how much bioproduct was harvested at the end of the process),
 {production costs}, including the fixed cost of operating and maintaining the facility, 
%the bioreactor, running the sensors, and maintaining the facilities, 
and variable manufacturing costs related to raw materials, labor, quality control, and purification. 
%Variable costs depend on the values of {actions and states}. %the CPPs and CQAs.
%across the entire production process: for example, purification costs depend on the quantity of unwanted metabolic wastes, which can be one of the CQAs evolving over time, while manufacturing costs depend on CPPs such as the amount of raw materials and media used.
The biomanufacturing industry often uses {(see, e.g., \citealp{martagan2016optimal,petsagkourakis2020reinforcement})} linear reward functions such as $r_t\left(\pmb{s}_t,\pmb{a}_t\right) = m_t + \pmb{b}^\top_t\pmb{a}_t + \pmb{c}^\top_t\pmb{s}_t$, {where the coefficients $m_t,\pmb{b}_t,\pmb{c}_t$ are nonstationary and represent both rewards and costs (that is, they can be either positive or negative). Thus, we can model situations that commonly arise in practice, where rewards are collected only at the final time $H$, while costs are incurred at each time step.}   %uncertain model parameters $\pmb{w}$, which represent the intrinsic attributes of the bioprocess.

Our goal is to use the reward function to guide control policies; however, we are only interested in controlling models whose parameters $\pmb{w}$ fall into some realistic range. Bioprocesses are subject to certain physical and biochemical laws: for example, in fermentation, cell growth rate and oxygen/substrate uptake rate generally do not fall outside a certain range. We let $\mathcal{W}$ be the set of all model parameters $\pmb{w}$ that satisfy these fundamental laws, and modify the reward structure to be
\begin{equation}
    r_t(\pmb{s}_t,\pmb{a}_t;\pmb{w})=\begin{cases}
     m_t + \pmb{b}_t^\top \pmb{a}_t+\pmb{c}_t^\top \pmb{s}_t & \pmb{w}\in\mathcal{W},\\
      m_c, & \pmb{w}\notin	\mathcal{W},
      \end{cases} \label{eq:reward structure}
\end{equation}
where $m_c$ is a negative constant. Essentially this means that, if the underlying model is outside the range of interest, then it does not matter how we set CPPs. This will prevent us from {assessing} %training 
control policies based on unrealistic dynamics. To streamline the notation, we often omit the explicit dependence of $r_t$ on $\pmb{w}$ in the following, except when necessary.

We can now formally define control policies and the optimization problem solved in this work. At each time $t$, the decisions are set according to $\pmb{a}_t = \pi_t\left(\pmb{s}_t\right)$, where the policy $\pi_t$ maps the state vector $\pmb{s}_t$ into the space of all possible action values. The %{(nonstationary)} 
policy $\pi = \left\{\pi_t\right\}^{H}_{t=1}$ is the collection of these mappings over the entire planning period. Let
\begin{equation}
    J(\pi;\pmb w)=\E_{\pmb\tau}\left[\left.\sum_{t=1}^H r_t(\pmb{s}_t,\pmb{a}_t)
\right|\pmb{\pi},\pmb{s}_1,\pmb{{w}}\right] \label{eq: objective-simple}
\end{equation}
be the expected cumulative reward earned by policy $\pi$ during the planning period. The expected value is taken over the \textit{conditional} distribution of the process trajectory $\pmb\tau$ given the model parameters $\pmb{w}$. Thus, the same policy will perform differently under different models.
Ideally, given a set $\mathcal{P}$ of candidate policies, we would like to find $\arg\max_{\pi\in\mathcal{P}} J\left(\pi;\pmb{w}^{\text{true}}\right)$, where $\pmb{w}^{\text{true}}$ contains the true parameters describing the underlying bioprocess. However, the true model is unknown, and optimizing with respect to any fixed $\pmb{w}$ runs the risk of poor performance due to model misspecification. In other words, the optimal policy with respect to $\pmb w$ may be very suboptimal with respect to $\pmb{w}^{\text{true}}$, especially under the situations with very limited real-world experimental data { and high stochasticity}. To mitigate model uncertainty, we instead search for a policy that performs well, on average, across many different posterior samples of process model with $\pmb{w}\sim p\left(\pmb w|\mathcal{D}\right)$, %Formally, we wish to solve
\begin{equation}
\label{eq:optimalpolicy}
\pi^* = \arg\max_{\pi\in\mathcal{P}} \mathcal{J}\left(\pi_{\pmb\theta}\right)
\end{equation}
where
\begin{equation}\label{eq:scriptJ}
\mathcal{J}\left(\pi_{\pmb\theta}\right) = \E_{\pmb{w}\sim p\left(\pmb w|\mathcal{D}\right)} \left[ J\left(\pi_{\pmb\theta};\pmb w\right) \right],
\end{equation}
with the expectation taken over the posterior distribution of $\pmb w$ given the real experimental data $\mathcal{D}$. Recall from (\ref{eq: objective-simple}) that $J\left(\pi;\pmb w\right)$ is an expected value over the stochastic uncertainty, % in the bioprocess, 
i.e., the uncertainty inherent in the trajectory $\pmb\tau$. Equation~(\ref{eq:scriptJ}) takes an additional expectation to account for model risk, thus hedging against this additional source of uncertainty.

Although the expectation in (\ref{eq:scriptJ}) cannot be computed in closed form, even for a fixed policy $\pi$, one could potentially estimate it empirically by averaging over multiple samples from the posterior $p\left(\pmb w|\mathcal{D}\right)$. As was discussed in Section \ref{sec:lineargaussian}, Gibbs sampling can be used to generate such samples. The complexity of this computation, however, makes it difficult to optimize over all possible policies, especially since each $\pi_t$ is defined on the set of all possible (continuous) CQA state values. The intractability of this problem is well-known as the ``curse of dimensionality'' \citep{Po11,powell2010merging}. In this work, we will focus on optimizing over the class of \textit{linear} policy functions. That is, we assume
\begin{equation}
   \pmb{a}_t =  {\pi}_{\pmb\vartheta_t}(\pmb{s}_t) = \pmb\mu^a_t + \pmb{\vartheta}_t^\top(\pmb{s}_t - \pmb\mu^{s}_t), \label{eq: linear policy func}
\end{equation}
where $\pmb\mu^a_t$ is the mean action value and $\pmb{\vartheta}_t$ is an $n\times m$ matrix of coefficients. To be consistent with the model of Section~\ref{sec:lineargaussian}, we use $\pmb\mu^a_t = \left(\lambda^1_t,...,\lambda^m_t\right)$. The linear policy $\pi_{\pmb\vartheta}$ is thus characterized by $\pmb\theta=\left\{\pmb{\vartheta}_t\right\}^{H-1}_{t=1}$, and the set $\mathcal{P}$ of possible policies can be replaced by a suitable closed convex set $\mathbb{C}$ of candidate values for the parameters $\pmb\vartheta$. The remainder of this paper is devoted to the problem of optimizing these policy parameters.

%As argued before for linear state transition functions, 
The assumption of a linear policy is reasonable if the process is monitored and controlled on a sufficiently small time scale. This is because, on such a time scale, {the effect of %CQAs 
state $\pmb{s}_t$ on action $\pmb{a}_t$ %CPPs 
is monotonic and approximately linear. For example, the biological state of a working cell does not change quickly, so the cell's generation rate of protein/waste and uptake rate of nutrients/oxygen is roughly constant in a short time interval; thus, the feeding rate can be proportional to the cell density.
%higher cell growth rate is typically linearly associated with higher feeding rate in the cell culture growth phase.
}

\begin{figure}[t]
	\centering
	\includegraphics[width=0.95\textwidth]{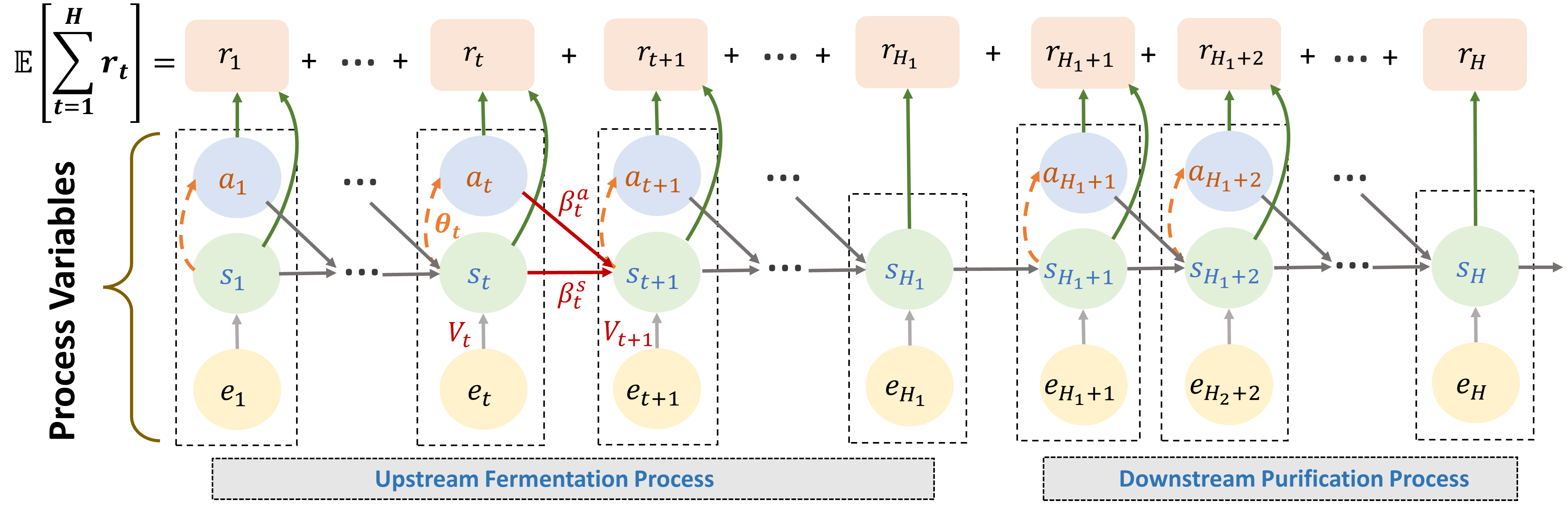}
	\caption{{An illustration of a policy augmented Bayesian network (PABN) with arrows representing interactions.}}
	\label{fig: dbn with RL}
\end{figure}

%With the introduction of rewards and policies, the KG network structure is illustrated in Figure \ref{fig: dbn with RL} {using an example of integrated biomanufacturing process consisting fermentation for drug substance production followed by chromatogrpahy purification.} %is modified from its original form as laid out in Section~\ref{sec:lineargaussian}. An illustrative example is given in Figure \ref{fig: dbn with RL}. 
%As in Figure \ref{fig:hybridexample}, 
%The integrated biomanufacturing process consists of fermentation for protein production followed by chromatography purification.

{
%The state transition $p(\pmb{s}_{t+1}|\pmb{s}_t,\pmb{a}_t)$ in (\ref{eq: LGBN matrix form}) models the process dynamics.
%The process state $\pmb{s}_{t+1}$ depends on the previous %state $\pmb{s}_t$ and action $\pmb{a}_t$, as well as the %residual $\pmb{e}_{t+1}$ from 
%uncontrolled factors (e.g., contamination).
With the introduction of policies and rewards, we create a \textit{policy augmented Bayesian network (PABN)}, as shown in Figure~\ref{fig: dbn with RL}. The edges connecting nodes at time $t$ to nodes at time $t+1$ represent the process dynamics discussed in Section \ref{sec:lineargaussian}. The augmentation consists of additional edges 1) connecting state to action at the same time stage, representing the causal effect of the policy; and 2) connecting action and state to the immediate reward $r_t(\pmb{s}_t,\pmb{a}_t)$.
}
%As before, the network models the causal effects of $\left\{\pmb{s}_t,\pmb{a}_t\right\}$ on future states $\pmb{s}_{t'}$ in later stages $t'>t$, but now includes additional edges connecting CQAs (states) to CPPs (actions) at the same time stage, representing the causal effect of the policy. We use the term \textit{policy augmented Bayesian network (PABN)} to refer to this structure.
To our knowledge, PABNs are a new modeling concept. Structurally, they have certain similarities to neural networks, in that we are choosing parameters to optimize outputs of the network, but a major difference is that {state/action interactions are modeled explicitly at each step. PABNs support \textit{interpretable} predictions of the effects of inputs at different times and their propagation through the bioprocess mechanism pathways.}

%each time period produces its own output, and the goal is to maximize cumulative reward. Furthermore, the output is an economic quantity, not a prediction, and the dynamics connecting nodes at different time periods are an interpretable model of an actual physical process, not simply a black-box transformation of data.

\subsection{Structural Properties {for Predictive Analysis and Policy Optimization}}
\label{sec:structprop}

%We close this section with 
Here we present
two important computational results which will be used later to facilitate {predictive analysis and }policy optimization. The first result relates  state $\pmb{s}_{t+1}$ at any time $t+1$ to an earlier value $\pmb{s}_h$ with $t \ge h $
{connected through mechanism pathways as shown in Figure \ref{fig: dbn with RL}}. 
%This is useful because, as shown in Figure \ref{fig: dbn with RL}, 
There may be multiple paths connecting a given pair of nodes. For example, $\pmb{s}_t$ is connected to $\pmb{s}_{t+1}$ directly through the coefficient $\pmb{\beta}^s_t$, but also through the %CPP 
action $\pmb{a}_t$ via the coefficients $\pmb{\beta}^a_t\pmb{\vartheta}_t$. Thus, $\pmb{s}_{t+1}$ depends on $\pmb{s}_h$ through all possible pathways connecting the two. The nature of these dependencies is characterized in Proposition~\ref{prop: state representation}, whose detailed derivation is given in  Appendix~\ref{appendix:proposition2}.

\begin{proposition}
\label{prop: state representation}
For any given decision epoch $t$ and starting state $\pmb{s}_h$ with $h\in \{1,2,\ldots,t\}$, we have
\begin{eqnarray*}
        \pmb{s}_{t+1}=\pmb{\mu}_{t+1}^s +\mathbf{R}_{h,t}(\pmb{s}_{h} - \pmb\mu^{s}_{h}) +\left[\sum_{i=h}^{t}\mathbf{R}_{i,t}V_i\pmb{z}_i + V_{t+1}\pmb{z}_{t+1}\right],
\end{eqnarray*}
where $V_i$ and $\pmb{z}_i$ are as in (\ref{eq: LGBN matrix form}), and $\mathbf{R}_{i,t} =\prod_{j=i}^t\left[\left(\pmb\beta_{j}^s\right)^\top + \left(\pmb\beta_{j}^a\right)^\top\pmb\vartheta_j^\top\right]$ represents the product of pathway coefficients from time step $i$ to $t$.
\end{proposition}

Using Proposition \ref{prop: state representation}, we can derive an explicit expression for the objective value $J\left(\pi_{\pmb\theta};\pmb{w}\right)$ of the linear policy $\pi_{\pmb{\theta}}$, as well as the mean and variance of the state vector $\pmb{s}_{t+1}$ under such a policy. Again, the derivation can be found in Appendix~\ref{subsec:proof_proposition3}. From this result, we can see the objective function is highly non-convex in the policy parameters $\pmb\theta$.

\begin{proposition}
\label{prop2}
Fix a model {specified by parameters} $\pmb{w} = (\pmb{\mu}^s,\pmb\mu^a,\pmb{\beta},\pmb\sigma,\pmb{v})$, a parametric linear policy $\pi_{\pmb\theta}$, and an initial state vector $\pmb{s}_1$. Then:
\begin{enumerate}%[label=\textbf{P.\arabic*}]
    \item The mean and variance of $\pmb{s}_{t+1}$ are given by
\begin{eqnarray*}
    \E[\pmb{s}_{t+1}]  &= & \pmb{\mu}_{t+1}^s +\mathbf{R}_{h,t}(\E[\pmb{s}_{h}] - \pmb\mu^{s}_{h})
    \label{eq.expState}
    =\pmb{\mu}_{t+1}^s +\mathbf{R}_{1,t}(\pmb{s}_1 - \pmb\mu^{s}_{1}) ~~ \mbox{ for }h\in \{1,2,\ldots,t\}
    \nonumber  \\
    \mbox{Var}[\pmb{s}_{t+1}] &=& \sum_{h=1}^{t}\mathbf{R}_{h,t}V_{h}^2\mathbf{R}_{h,t}^\top+V_{t+1}^2. \nonumber
\end{eqnarray*}
\item Under the linear reward structure of (\ref{eq:reward structure}),
\begin{equation}
    J\left(\pi_{\pmb\theta}; \pmb w\right) = \begin{cases} \sum^{H}_{t=1}\left\{m_t+\pmb b_t^\top\pmb\mu^a_t+\pmb{c}_t^\top\pmb\mu_t^s+(\pmb{b}_t^\top\pmb\vartheta_t^\top+\pmb{c}_t^\top)\mathbf{R}_{1,t}(\pmb{s}_1 - \pmb\mu_1^s)\right\} & \pmb{w} \in \mathcal{W}, \label{eq: objective-simple 2}\\
    Hm_c & \pmb{w} \notin \mathcal{W}.
    \end{cases}
\end{equation}
\end{enumerate}
\end{proposition}

\subsection{{Predictive Analysis Using the Shapley Value}}
\label{eq.SV}

Suppose that the model parameters $\pmb{w}$ are given. We are interested in quantifying the contribution of each input, at any time $h$, to the expected future trajectory $\mbox{E}[\pmb{s}_{t+1}| \pmb{s}_h,\pmb{a}_h]$ with $t \geq h$.
Specifically, %denote $\mathcal{O}_h=\{s^k_h\}_{k=1}^n\cup \{a^k_h\}_{k=1}^m$ impacting on $\E[\pmb{s}_{t+1} | \mathcal{O}_h; \pmb{w}]$ be the combined set of state and action inputs at time $h$. 
{let $\mathcal{O}_h=\{s^k_h\}_{k=1}^n\cup \{a^k_h\}_{k=1}^m$ denote the set of state and action inputs.}
The predicted contribution of any input ${o_h} \in \mathcal{O}_h$ is measured by the Shapley value
%how the effect of each input  ${o}_h^k$ in $ \mathcal{O}_h=\{s^k_h\}_{k=1}^n\cup \{a^k_h\}_{k=1}^m$ propagates through mechanism pathways impacting on
%the prediction of state $\pmb{s}_{t+1}$ from the observed state $\pmb{s}_{h}$ at time $h\leq t$. Given the DBN model parameters $\pmb{w}$, 
%= (\pmb{\mu}^s,\pmb\mu^a,\pmb{\beta},\pmb\sigma,\pmb{v})$,  we quantify the feature importance of each random input (i.e. $s^k_h$ with $k=1,\ldots, n$) through measuring its contribution to 
%$\mbox{E}[\pmb{s}_{t+1} | \pmb{s}_h, \pmb{a}_h]$. 
\begin{equation}
    \mbox{Sh}\left(\pmb{s}_{t+1}| {o_h}; \pmb{w}\right)
	=\sum_{\mathcal{U}\subset \mathcal{O}_h/\{{o_h}\}}\dfrac{(|\mathcal{O}_h| -|\mathcal{U}|-1)! |\mathcal{U}|!} {|\mathcal{O}_h|!}\left[ g(\mathcal{U}\cup\left\{{o_h}\right\}) - g(\mathcal{U}) \right],
	\label{eq.ShapleyEffect1}
	\nonumber 
\end{equation}
where $g(\cdot)=\E[\pmb{s}_{t+1}|\cdot; \pmb{w}]$ represents the expected future trajectory based on the DBN model, and $|\mathcal{U}|$ is the cardinality of $\mathcal{U}\subset \mathcal{O}_h/\left\{o_h\right\}$.
By definition, we have
\begin{equation*}
f(\mathcal{O}_h)=\E[\pmb{s}_{t+1}|\pmb{s}_{h},\pmb{a}_h;\pmb{w}]=\sum_{{o_h}\in \mathcal{O}_h} \mbox{Sh}\left(\pmb{s}_{t+1}| {o_h}; \pmb{w} \right),
\end{equation*}
that is, the predicted mean is the sum of all the contributions. Then, we can use
$\bar{\mbox{Sh}}\left(\pmb{s}_{t+1}| {o_h}\right) = \int \mbox{Sh}(\pmb{s}_{t+1}|{o_h};\pmb{w}) p(\pmb{w}|\mathcal{D}) d\pmb{w}$ to assess the expected contribution of ${o_h}$ over the posterior distribution.
%The similar study can be conducted on the variance; see %\cite{xie2020bayesian}.
Theorem~\ref{thm: fctor imptance to cond. expectation} calculates the Shapley value (conditional on $\pmb{w}$) in closed form for arbitrary $h$ and $t$; see
the derivation in Appendix~\ref{appendix sec: proofs for SV Theorem}.
In this way, we can quantify how the influence of an input at time $h$ propagates far into the future.

\begin{comment}

In game theory, this SV can be interpreted as
the average incremental payoff by including player $k$ over all possible cooperation group formations, i.e., $\mathcal{U}\subset \mathcal{O}_h/\left\{s^k_h\right\}$, and
$\mbox{Sh}(\pmb{s}_{t+1}| s_h^k)$ can be used to measure the contribution of the player $k$. This assessment approach satisfies the “efficiency property"
that the sum of the SVs of all players equals the gain of the grand coalition, i.e., $f(\mathcal{O}_h)=\E[\pmb{s}_{t+1}|\pmb{s}_{h},\pmb{a}_h]= \mathbf{R}_{h,t}\left[\left(\pmb\beta_h^s\right)(\pmb{s}_h-\mu_h) + \left(\pmb\beta_h^a\right)(\pmb{a}_h-\mu_h)\right]=\sum_{{o_h}\in \mathcal{O}_h} \mbox{Sh}\left(\pmb{s}_{t+1}| {o_h}\right)$. We summarize the SV-based feature importance in Theorem~\ref{thm: fctor imptance to cond. expectation}, which has a simple closed form.
\end{comment}

% The definition of feature importance can be generalized to various of choices of function $f$. For example, we may define $f_{m+1}(x)\triangleq \mbox{E}[X_{m+1}\mid x]$ for regression; we may also define $f_{m+1}(x)\triangleq \mbox{P}[X_{m+1} < c\mid x]$ to decompose the probability of evidence.
\begin{theorem}
\label{thm: fctor imptance to cond. expectation}
%Suppose the DBN based predictor is $f_{t+1}(\cdot)\triangleq \E[\pmb{s}_{t+1}|\cdot]$. 
For any input $s^k_h$ and $a^k_h$, the Shapley value contribution is given by 
\begin{eqnarray*}
\mbox{Sh}\left(\pmb{s}_{t+1}| s_h^k; \pmb{w}\right) &=& \mathbf{R}_{h+1,t}\left(\pmb\beta_h^s\right)^{{\top}}(s^k_h-\mu^{k}_h)\mathrm{1}_k\\
\mbox{Sh}\left(\pmb{s}_{t+1}| a_h^k; \pmb{w}\right)&=& \mathbf{R}_{h+1,t}\left(\pmb{\beta}_{h}^a\right)^\top (a^k_h-\lambda^{k}_h)\mathrm{1}_k, 
\end{eqnarray*}
where $\mathrm{1}_k$ is the standard basis column vector whose components are all zero except the $k$th one equal to 1. 
%For any action $a^k_h$, the SV contribution is $\mbox{Sh}\left(\pmb{s}_{t+1}| a_h^k\right)= \mathbf{R}_{h+1,t}\left(\pmb{\beta}_{h}^a\right)^\top (a^k_h-\lambda^{k}_h)\mathrm{1}_k$. 
% Moreover, this importance value doesn't depend on observation set $\mathcal{O}$.
\end{theorem}

{Though the Shapley values are derived for individual state and action inputs, the predictions used to calculate them are influenced by the control policy. Thus, the policy is indirectly reflected in the estimated contributions of various $o_h$. A more direct interpretation of the policy can be obtained by treating their policy parameters as action inputs and calculating their Shapley values. This falls outside the scope of the present paper, but is a subject for future work.}

\section{Policy Optimization with Projected Stochastic Gradients}
\label{sec: optimization}

In this section, we propose a policy optimization method to solve (\ref{eq:optimalpolicy}) with $\mathcal{P}$ restricted to the class of parametric linear policies introduced in (\ref{eq: linear policy func}). Essentially, we use gradient ascent to optimize $\mathcal{J}$ over the space of parametric linear policies, but there are some nuances in the analysis due to the fact that the space of policies is constrained to the set $\mathbb{C}$. Section~\ref{subsec: GD} discusses the computation of the policy gradient, while Section \ref{sec:policyoptimization} states the overall optimization framework in which the gradient computation is used.

\subsection{Gradient Estimation and Computation}
\label{subsec: GD}

For notational convenience, we represent the parametric linear policy $\pi_{\pmb\theta}$ by the parameter vector $\pmb\theta$, and write $J,\mathcal{J}$ in (\ref{eq: objective-simple}) and (\ref{eq:optimalpolicy}) as functions of $\pmb\theta$. We first establish the differentiability of the objective function $\mathcal{J}$, which will allow us to use gradient ascent to optimize the parameters. In the following, we understand $\pmb\theta$ as a vector, i.e., $\pmb\theta = \left(\text{vec}\left(\pmb{\vartheta}_1\right),...,\text{vec}\left(\pmb{\vartheta}_{H-1}\right)\right)^\top$, where $\text{vec}\left(\cdot\right)$ denotes a linear transformation converting an $n\times m$ matrix into a column vector. 
For notation simplification, we write the policy gradient $\nabla_{\pmb{\theta}} \mathcal{J}\left(\pmb\theta\right)$ as $\nabla \mathcal{J}\left(\pmb\theta\right)$ in the following presentation.

\begin{lemma}\label{lemma: L-smooth}
The objective function $\mathcal{J}$ is differentiable and its gradient satisfies
\begin{equation}\label{eq:ipa}
\nabla \mathcal{J}\left(\pmb\theta\right) = \mathbb{E}_{\pmb{w}\sim p\left(\pmb{w}|\mathcal{D}\right)}\left[\nabla_{\pmb\theta} J\left(\pmb\theta;\pmb{w}\right)\right].
\end{equation}
Furthermore, $\nabla\mathcal{J}$ is L-smooth over the closed convex set $\mathbb{C}$, that is,
\begin{equation*}
\Vert\nabla \mathcal{J}(\pmb{x}) -\nabla \mathcal{J}(\pmb{y})\Vert\leq L\lVert \pmb{x}-\pmb{y}\rVert, \qquad \pmb{x},\pmb{y}\in \mathbb{C}.
\end{equation*}
\end{lemma}

Equation (\ref{eq:ipa}) justifies the interchange of derivative and expectation, allowing us to focus on computing the gradient $\nabla_{\pmb\theta} J\left(\pmb\theta;\pmb{w}\right)$. The outer expectation over the posterior distribution of $\pmb{w}$ can be estimated using sample average approximation (SAA) method: we use Gibbs sampling, as discussed in Section \ref{sec:lineargaussian}, to generate posterior samples $\left\{\pmb{w}^{(b)}\right\}^B_{b=1}$ from the distribution $p\left(\pmb{w}|\mathcal{D}\right)$, and calculate
\begin{equation}
    \nabla  \widehat{\mathcal{J}}(\pmb{\theta})=\frac{1}{B}\sum^B_{b=1}\nabla_{\pmb\theta} {J}\left(\pmb{\theta};\pmb{w}^{(b)}\right). \label{eq: gradient}
\end{equation}

In the following discussion, we consider a fixed $h\in\left\{1,...,H-1\right\}$ as well as a fixed model $\pmb{w} \in\mathcal{W}$. Recalling (\ref{eq: objective-simple}), we write
\begin{equation*}
    \nabla_{\pmb\vartheta_h}{J}(\pmb\theta;\pmb w)=\sum^H_{t=1}\nabla_{\pmb\vartheta_h}\E_{\pmb\tau}\left[\left. r_t(\pmb{s}_t,\pmb{a}_t)
\right|\pmb{\pi}_{\pmb{\theta}},\pmb{s}_1,\pmb{{w}}\right].
\end{equation*}
By plugging in the reward and policy functions in (\ref{eq:reward structure}) and (\ref{eq: linear policy func}), the expected reward %can be written as
becomes,
\begin{equation*}
\label{eq: general reward gradient}
    \E_{\pmb\tau}\left[\left.\sum_{t=1}^H r_t(\pmb{s}_t,\pmb{a}_t)
\right|\pmb{\pi}_\theta,\pmb{s}_1,\pmb{{w}}\right] =  m_t+\pmb c_t^\top \E[\pmb{s}_t|\pmb{\pi}_{\pmb\theta},\pmb{s}_1,\pmb{w}]+\pmb b_t^\top\left(\pmb\mu_t^a+\pmb\vartheta_t^\top(\E[\pmb{s}_t|\pmb{\pi}_{\pmb\theta},\pmb{s}_1,\pmb{w}]-\pmb{\mu}_t^s)\right).
\end{equation*}
Similarly, for $\pmb{w}\notin\mathcal{W}$, we have $\mathbb{E}_{\pmb\tau}\left[\left.\sum_{t=1}^H r_t(\pmb{s}_t,\pmb{a}_t)
\right|\pmb{\pi}_{\pmb{\theta}},\pmb{s}_1,\pmb{{w}}\right]=H m_c$.

Let $\bar{\pmb{s}}_{t}=\E[\pmb{s}_{t}|\pmb{\pi}_{\pmb\theta},\pmb{s}_1,\pmb{w}]$. Using Propositions \ref{prop: state representation}-\ref{prop2} and the functional forms of the reward and policy functions, we obtain an expression
\begin{equation}
\bar{r}_t(\pmb\theta;\pmb w)\equiv\E_{\pmb\tau}\left[\left. r_t(\pmb{s}_t,\pmb{a}_t)
\right|\pmb{\pi}_{\pmb\theta},\pmb{s}_1,\pmb{{w}}\right]
=
m_t+\pmb c_t^\top \bar{\pmb{s}}_t+\pmb b_t^\top(\pmb\mu_t^{a}+\pmb\vartheta_t^\top(\bar{\pmb{s}}_t-\pmb{\mu}_t^{s}))
\label{eq.barR}
\end{equation}
for the expected reward at time $t$. This expression is equivalent to the one obtained in Proposition~\ref{prop2}, but functionally it serves a different purpose: instead of expanding out all the pathways leading from time $1$ to time $t$, we instead represent them by $\bar{\pmb{s}}_t$. We can also write a partial expansion of the pathways starting at some earlier time $h \leq t$. Such representations are crucial in policy optimization, where we need to evaluate partial derivatives of $\bar{r}_t$ with respect to $\pmb{\vartheta}_h$ for just such an earlier time $h$. Thus, (\ref{eq.barR}) leads into the following computational result; the proof is given in Appendix~\ref{appendix sec: proofs for Nested Backpropagaation Theorem}.

\begin{theorem}[Nested Backpropagation]
\label{thm: nested backpropagation}
Let $\mathbf{R}_{h,t}$ be as in Proposition \ref{prop: state representation}, with $\mathbf{R}_{t,t-1} = \mathbb{I}_{n\times n}$. % by convention. 
Fix a model $\pmb{w}$ and policy parameter vector $\pmb{\theta}$. For any $h\leq t$, we have
 \begin{equation}
      \frac{\partial \bar{r}_t (\pmb\theta;\pmb w)}{\partial \pmb \vartheta_h} = \begin{cases}
     (\pmb{c}_t+\pmb\vartheta_t\pmb{b}_t)(\bar{\pmb{s}}_{h}-\pmb\mu_{h}^s)^\top\pmb\delta_h^t & \text{if $h<t$}\\
      (\bar{\pmb{s}}_h -\pmb\mu_h^s)\pmb b_t^\top, & \text{if $h=t$}
      \end{cases}\label{eq: nested backpropagation}
  \end{equation}
  where $\pmb\delta_h^t=\mathbf{R}_{h+1,t-1}\left(\pmb\beta_{h}^{a}\right)^\top$.
\end{theorem}

 \begin{figure}[t]
	\centering
	\includegraphics[width=0.75\textwidth]{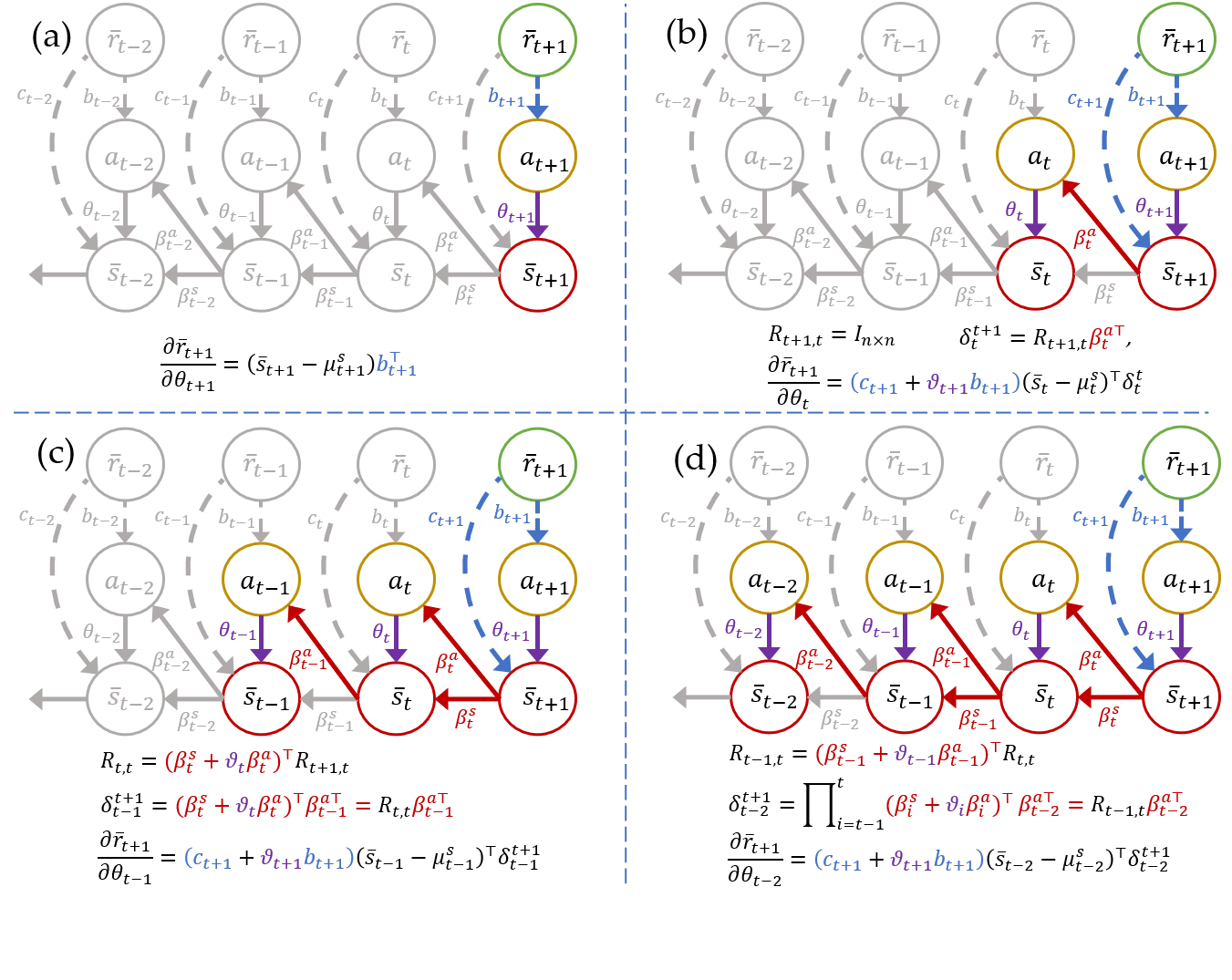}
	\vspace{-0.25in}
	\caption{Illustration showing nested backpropagation computations, highlighting the pathways involved in each policy gradient calculation.}
	\label{fig: nested backpropagation}
	\vspace{-0.1in}
\end{figure}

Theorem \ref{thm: nested backpropagation} gives us a computationally efficient way to update the gradient of $J\left(\pmb{\theta};\pmb{w}\right)$ which we call ``\textit{nested backpropagation}'' to highlight both similarities and differences with the classic backpropagation algorithm used in the calibration of neural networks \citep{Go95}.
Each reward can be backpropagated to any policy parameter (say, $\pmb{\vartheta}_h$) through the network or knowledge graph as shown in Figure \ref{fig: nested backpropagation}. The term ``backpropagation'' refers to the fact that the calculation of gradients proceeds backward through the network, from time $t$ to time $h$. This structure is shared by neural networks. The key distinction, however, is that in our problem we must compute the gradient of $\bar{r}_t$ for \textit{every} time period $t$, whereas in classic neural networks there is a single output function calculated at the final (terminal) node. Thus, there may be overlapping pathways between two such gradients, creating opportunities to save and reuse gradient computations. % made at time $t$ for gradient computations at time $t-1$.
Specifically, the computation of $\frac{\partial \bar{r}_t}{\partial \pmb\theta_h}$ can reuse the propagation pathways $\mathbf{R}_{h+1,t-1}$ through applying the equation
\begin{equation*}
\mathbf{R}_{h,t-1}=(\pmb\beta^s_h+\pmb\vartheta_h\pmb\beta_h^a)^\top\mathbf{R}_{h+1,t-1}.
\end{equation*}
Figure \ref{fig: nested backpropagation} illustrates these reused computations through the examples (a) $\frac{\partial \bar{r}_{t+1}}{\partial \pmb\vartheta_{t+1}}$, (b) $\frac{\partial \bar{r}_{t+1}}{\partial \pmb\vartheta_{t}}$, (c) $\frac{\partial \bar{r}_{t+1}}{\partial \pmb\vartheta_{t-1}}$ and (d) $\frac{\partial \bar{r}_{t+1}}{\partial \pmb\vartheta_{t-2}}$. In Figure \ref{fig: nested backpropagation}(a), the gradient $\frac{\partial \bar{r}_{t+1}}{\partial \pmb\vartheta_{t+1}}$ is computed using only the nodes $r_{t+1}$, $\pmb{a}_{t+1}$ and $\pmb{s}_{t+1}$. In Figure \ref{fig: nested backpropagation}(b), the gradient $\frac{\partial \bar{r}_{t+1}}{\partial \pmb\theta_{t}}$ propagates the reward signal $\bar{r}_{t+1}$ back to $\pmb\vartheta_{t}$, and the computation now uses information from the nodes $r_{t+1}$, $\pmb{a}_{t+1}$, $\pmb{s}_{t+1}$, $\pmb{a}_t$ and $\pmb{s}_{t}$ and associated edges. In Figure~\ref{fig: nested backpropagation}(d), we \textit{reuse} the pathway $\mathbf{R}_{t,t}=(\pmb\beta^s_t+\pmb\vartheta_t\pmb\beta_t^a)^\top\mathbf{R}_{t+1,t}$ from Figure \ref{fig: nested backpropagation}(c) to compute
$\mathbf{R}_{t-1,t}=(\pmb\beta^s_{t-1}+\pmb\vartheta_{t-1}\pmb\beta_{t-1}^a)^\top\mathbf{R}_{t,t}$.

%To compute gradient for $\pmb\vartheta_h$, we need to sum up gradients from all rewards, i.e.,  %$\nabla_{\pmb\vartheta_h}J(\pmb\theta;\pmb{w})=\sum_{t=1}^H\nabla_{\pmb\vartheta_h}\E_{\pmb\tau\sim D_{\pmb{w}}^{\pmb{\theta}}}\left[\left. %r_t(\pmb{s}_t,\pmb{a}_t)
%\right|\pmb{\pi}_{\pmb\theta},\pmb{s}_1,\pmb{{w}}\right] =\sum_{t=1}^H \frac{\partial \bar{r}_t}{\partial \pmb\vartheta_h}$.

The formal statement of the nested backpropagation procedure is given in Algorithm \ref{algo: NBP}. In Step~(1), we first pass through the network to efficiently precompute the propagation pathways $\mathbf{R}_{h,t}$ through reuse, and store $\pmb\delta_h^t$ that will be used multiple times to compute the gradients $ \frac{\partial \bar{r}_t}{\partial \pmb \vartheta_h}$ for $t=1,2,\ldots,H$ and $h=t-1,t-2,\ldots,1$ in Step~(2).
%that will be used multiple times (Step 2), and then pass through it again (Step 2) to compute the gradients, looking up the stored pathways as necessary.
Proposition~\ref{prop: time complexity analysis for Nested Backpropagaation} proves that this approach reduces the cost of computing $\nabla_{\pmb\theta}J\left(\pmb{\theta};\pmb{w}\right)$ by a factor of $\mathcal{O}\left(H\right)$ compared to a brute-force approach that does not reuse pathways. As will be shown in our computational study {in Section~\ref{subsec:empiricalStudyModelRisk}}, the time savings can be quite significant. The proof of Proposition~\ref{prop: time complexity analysis for Nested Backpropagaation} can be found in Appendix~\ref{appendix sec: time complexity analysis for Nested Backpropagaation}.

 \begin{algorithm}[t]
Input: DBN coefficients $\pmb{w}$, policy parameters $\pmb{\vartheta}_t$, reward function $r_t(\bar{\pmb{s}}_t ,\pmb{a}_t)=m_t + \pmb{c}_t \bar{\pmb{s}}_t  + \pmb{b}_t \pmb{a}_t$ for all $t$\; %process state $\bar{\pmb{s}}_t=\E[\pmb{s}_t|\pmb{\pi}_{\pmb\theta},\pmb{s}_1,\pmb{w}]$, $t\in\mathcal{T}$.
% (for BN-MDP-PGA, it's estimated by using Proposition\ref{prop2}\ref{prop: cond exp, var, cov})
% and for BN-MDP-PSGA this conditional expected state can be estimated by sampling, i.e. $\E[\pmb{s}_t|\pmb{\pi}_{\pmb\theta},\pmb{s}_1,\pmb{w}] = \frac{1}{r}\sum_{i=1}^r \pmb{s}_t^{(i)}$).

%Initialize policy weights $\pmb\theta$. Initialize $\mathcal{E}_{1}$, $\pmb{{w}}_{1}$ and $\pmb\vartheta_{1}$ by empty set\;
1. Compute $\mathbf{R}_{h,t}$ and $\pmb\delta^t_h$ with $t=1,2,\ldots,H$ and $h=1,2,\ldots, t$ as follows:

 \For{$t=1,2,\ldots,H$}{
 Set $\mathbf{R}_{t+1,t}=I_{n\times n}$;\\
 \For{$h=t,t-1,\ldots,1$}{
 (a) $\pmb\delta_h^t = \mathbf{R}_{h+1,t}\pmb\beta^{a\top}_h$;\\
  (b) $\mathbf{R}_{h,t}=(\pmb\beta_{h}^s+\pmb\vartheta_h\pmb\beta_{h}^a)^\top\mathbf{R}_{h+1,t}$;\\
 }
 }

 2. Compute policy gradients $ \frac{\partial \bar{r}_t}{\partial \pmb \vartheta_h}$ for $t=1,2,\ldots,H$ as follows:\\
 $\frac{\partial \bar{r}_1}{\partial \pmb \vartheta_1}=(\bar{\pmb{s}}_1 -\pmb\mu_1^s)\pmb b_1^\top$;\\
 \For{$t=2,3,\ldots,H$}{
 $\frac{\partial \bar{r}_t}{\partial \pmb \vartheta_t}=(\bar{\pmb{s}}_t -\pmb\mu_t^s)\pmb b_t^\top$;\\
 \For{$h=t-1,t-2,\ldots,1$}{
  $\frac{\partial \bar{r}_t}{\partial \pmb \vartheta_h}=(\pmb{c}_t+\pmb\vartheta_t\pmb{b}_t))(\bar{\pmb{s}}_{h}-\pmb\mu_{h}^s)^\top\pmb\delta_h^t $;\\
 }
 }
 3. Compute the policy gradientof the cumulative reward:
\begin{equation}
     \nabla_{\pmb{\theta}}J(\pmb\theta;\pmb w) = \left(\mbox{vec}\left(\sum^H_{t=1}\nabla_{\pmb\vartheta_1} \bar{r}_t(\pmb\theta;\pmb w)\right),\ldots,\mbox{vec}\left(\sum^H_{t=1}\nabla_{\pmb\vartheta_{H-1}} \bar{r}_t(\pmb\theta;\pmb w)\right)\right)^\top.
     \nonumber 
\end{equation}
 %$\left\{\frac{\partial \bar{r}_t}{\partial \pmb \vartheta_h} \right\}$, $t=1,2,\ldots, H$ and $h=1,2,\ldots,t$.
\caption{Nested backpropagation procedure for policy gradient computation.}
\label{algo: NBP}
\end{algorithm}

\begin{proposition}
\label{prop: time complexity analysis for Nested Backpropagaation}
Fix a model $\pmb{w}$ and policy parameters $\pmb{\theta}$. The cost to compute $\nabla_{\pmb{\theta}}J(\pmb\theta;\pmb w)$ is $\mathcal{O}\left(H^2n^2m\right)$ for nested backpropagation (Algorithm \ref{algo: NBP}) and $\mathcal{O}\left(H^3n^2m\right)$ for brute force.
\end{proposition}

\subsection{Policy Optimization Algorithm}\label{sec:policyoptimization}

Recall that the gradient of $\mathcal{J}$ is estimated using SAA in (\ref{eq: gradient}), with ${J}\left(\pmb{\theta};\pmb{w}^{(b)}\right)$ computed using Algorithm \ref{algo: NBP}. We can now search for the optimal $\pmb\theta$ using a recursive gradient ascent update. Given $\pmb{\theta}_k$, we compute
\begin{equation}\label{eq:sa}
\pmb{\theta}_{k+1} = \Pi_{\mathbb{C}}\left(\pmb{\theta}_k + \eta_k \nabla\widehat{\mathcal{J}}\left(\pmb{\theta}_k\right)\right),
\end{equation}
where $\eta_k$ is a suitable 
stepsize, $\mathbb{C}$ is a closed convex feasible region as discussed in Section~\ref{sec:rewardspolicies}, and $\Pi_{\mathbb{C}}$ is a projection onto $\mathbb{C}$. 
The projection of a point $\pmb{y}$ onto ${\mathbb{C}}$ is defined as 
$
\Pi_{\mathbb{C}}(\pmb{y}) = \argmin_{\pmb{x}\in {\mathbb{C}}}\frac{1}{2}\lVert \pmb{x} - \pmb{y}\rVert^2.
$
We summarize some previously established properties of the projection, which will be needed for our convergence analysis later; the proofs are given in Section 2.2 of \cite{jain2017non}, and thus are omitted here. Throughout the paper, we use the usual Euclidean norm and inner product.

\begin{comment}
\begin{definition}
%The projection of a point $\pmb{y}$ onto ${\mathbb{C}}$ is defined as
Define the projection of a point $\pmb{y}$ onto ${\mathbb{C}}$ as
$
\Pi_{\mathbb{C}}(\pmb{y}) = \argmin_{\pmb{x}\in {\mathbb{C}}}\frac{1}{2}\lVert \pmb{x} - \pmb{y}\rVert^2.
$
\end{definition}
\end{comment}

\begin{proposition} \label{prop: projection}
For any set ${\mathbb{C}}\subset \mathbb{R}^p$, the following inequalities hold:
\begin{enumerate}
\item If $\pmb{y}\in {\mathbb{C}}$, then $\Pi_{\mathbb{C}}(\pmb{y})=\pmb{y}$.

\item For any $\pmb{y}\in \mathbb{R}^p$ and $\pmb{x}\in {\mathbb{C}}$, $\Vert\Pi_{\mathbb{C}}(\pmb{y}) - \pmb{y} \Vert\leq \Vert\pmb{x} -\pmb{y}
\Vert$.
\label{prop 2-2}

\item If $\mathbb{C}$ is convex, $\Pi_{\mathbb{C}}$ is non-expansive. That is, for any $\pmb{x},\pmb{y}\in \mathbb{R}^p$,
\begin{equation*}
    \Vert \Pi_{\mathbb{C}}(\pmb{x}) -\Pi_{\mathbb{C}}(\pmb{y})\Vert^2 \leq \langle\Pi_{\mathbb{C}}(\pmb{x}) -\Pi_{\mathbb{C}}(\pmb{y}),\pmb{x}-\pmb{y} \rangle\leq \Vert\pmb{x}-\pmb{y}\Vert^2.
\end{equation*}
\item For any convex set ${\mathbb{C}}\in \mathbb{R}^p$, $\pmb{y}\in \mathbb{R}^p$ and $\pmb{x}\in {\mathbb{C}}$,
we have
 $\langle\pmb{x} -\Pi_{\mathbb{C}}(\pmb{y}),\pmb{y}-\Pi_{\mathbb{C}}(\pmb{y}) \rangle\leq 0.
$
\end{enumerate}
\end{proposition}

It is sometimes convenient to rewrite (\ref{eq:sa}) as
\begin{equation}
\label{eq:sawithg}
\pmb{\theta}_{k+1} = \pmb{\theta}_k + \eta_k\widehat{g}_k\left(\pmb{\theta}_k\right)
~~\mbox{ with } ~~
\widehat{g}_k\left(\pmb{\theta}\right) = \frac{1}{\eta_k}\left(\Pi_{\mathbb{C}}\left(\pmb{\theta}+\eta_k\nabla\widehat{\mathcal{J}}\left(\pmb\theta\right)\right)-\pmb\theta\right)
\end{equation}
where $\widehat{g}_k\left(\pmb{\theta}\right)$ can be viewed as a generalized gradient estimator. This representation is identical to (\ref{eq:sa}), except that (\ref{eq:sawithg}) resembles an unconstrained gradient ascent update.

The complete statement of the policy optimization procedure, which we call DBN-RL (``DBN-assisted reinforcement learning''), is given in Algorithm \ref{algo1}. In each iteration, we generate $B$ model parameters, $\left\{\pmb{w}^{(b)}\right\}^B_{b=1}$, by sampling from the posterior distribution $p\left(\pmb{w}|\mathcal{D}\right)$, then use nested backpropagation to calculate gradients for each individual model. We then average over the models and apply (\ref{eq:sawithg}) to update the policy parameters. It is worth noting that the only randomness in this stochastic gradient method comes from the use of SAA to approximate the expectation in (\ref{eq:scriptJ}) with an average over $B$ posterior samples in (\ref{eq: gradient}). 
The computational results of Theorem~\ref{thm: nested backpropagation} and Algorithm~\ref{algo: NBP} allow us to explicitly compute $\nabla_{\pmb\theta} {J}\left(\pmb{\theta};\pmb{w}^{(b)}\right)$ for $b=1,2,\ldots,B$.

%The computational results of Proposition \ref{prop2} allow us to explicitly compute the expectation over the conditional distribution of the process trajectory $\pmb{\tau}$.

 \begin{algorithm}[t]
Input: the maximum number of episodes $K$; the convex constraint set $\mathbb{C}$;
initial parameter $\pmb\theta_0$ for linear policy $\pmb\pi_{\pmb\theta_0}(\pmb{s})$, $\forall \pmb{s} \in \mathcal{S}, \pmb\theta \in\mathbb{C}$;
posterior distribution $p(\pmb{w}|\mathcal{D})$
%$p(\pmb \mu^a, \pmb \mu^s, \pmb \sigma^2, \pmb v^2, \pmb \beta|\mathcal{D})$
for process knowledge graph model\;
%Initialize policy weights $\pmb\theta$. Initialize $\mathcal{E}_{1}$, $\pmb{{w}}_{1}$ and $\pmb\vartheta_{1}$ by empty set\;
 \For{$k=1,2,\ldots,K$}{
%  1. Generate posterior samples $\pmb{{w}}_{k} \sim p(\pmb{{w}}|\mathcal{D}_p)$ and build the transition model with new parameter $\pmb{{w}}_{k}$, i.e., $p(s_{t+1}|s_t,a_t,\pmb{{w}}_{k})$ for $t=1,2,\ldots,H$ \;
 1. Generate $B$ posterior samples of process model parameters $\pmb w_k^{(b)} \sim p(\pmb{w}|\mathcal{D})$ with $b=1,2,\ldots,B$;\\
 \For{$b=1,2,\ldots,B$}{
  %2. Compute gradients of reward with respect to policy parameters for $b$-th DBN model, $\left\{\frac{\partial \bar{r}_t}{\partial \pmb \vartheta_h} \right\}$, $t=1,2,\ldots, H$ and $h=1,2,\ldots,t$ using Algorithm~\ref{algo: NBP}\;
\eIf{$\pmb w_k^{(b)}\in\Omega$}{
   Calculate the policy gradient $\nabla_{\pmb{\theta}}J(\pmb\theta_k;\pmb w_k^{(b)})$ by using Algorithm~\ref{algo: NBP};\
   }{
   $\nabla_{\pmb{\theta}}J(\pmb\theta_k;\pmb w_k^{(b)})=0$;\
  }
 %by Eq.~\eqref{eq: stochastic policy gradient estimator}\;
%  (c) Calculate the gradient mapping $\widehat{g}_{c}^{(b)}(\pmb\theta_k)=\frac{1}{\eta_k}(\Tilde{\pmb\theta}_k^{(b)}-\pmb\theta_k)$, where $\Tilde{\pmb\theta}_k^{(b)}=\Pi_{C}(\pmb\theta_k+\eta_k\widehat{\nabla_{\pmb{\theta}_k}J}(\pmb\theta_k))$\;
 }\
 3. Calculate $\nabla \widehat{\mathcal{J}}(\pmb{\pi}_{\pmb{\theta}})=\frac{1}{B}\sum^B_{b=1}\nabla_{\pmb\theta} J\left(\pmb{\pi}_{\pmb{\theta}};\pmb{w}_k^{(b)}\right)$\;
 %$\nabla_{\pmb\theta} \widehat{\mathcal{J}}(\pmb{\pi}_{\pmb{\theta}})$ by Eq.~\eqref{eq: gradient}
 \ 4. Obtain the gradient mapping $\widehat{g}_{c}(\pmb\theta_k)=\frac{1}{\eta_k}(\Tilde{\pmb\theta}_k-\pmb\theta_k)$, where $\Tilde{\pmb\theta}_k=\Pi_{C}\left(\pmb\theta_k+\eta_k\nabla\widehat{\mathcal{J}}(\pmb\theta_k)\right)$\;
%  Let $\widehat{g}_{c}(\pmb\theta_k)= \sum_{b=1}^B\widehat{g}_{c}^{(b)}(\pmb\theta_k)$\;
 \ 5. Update the policy parameters $\pmb{\theta}_{k+1} \leftarrow \pmb{\theta}_k+ \eta_k\widehat{g}_{c}(\pmb\theta_k)$.
 }
\caption{DBN-RL algorithm for policy optimization. %(PABN)
%(BN-MDP-PGA)
}\label{algo1}
\end{algorithm}

\section{Convergence Analysis}
\label{sec: convergence analysis}

Here we prove the convergence of DBN-RL to a local optimum of $\mathcal{J}$ and characterize the convergence rate. Recall from Proposition \ref{prop2} that $\mathcal{J}$ is non-convex in $\pmb\theta$. Consequently, only local convergence can be guaranteed, as is typical of stochastic gradient ascent methods.

There is a rich literature on the convergence analysis of this class of algorithms. %, covering a wide variety of settings. 
However, the proofs often rely on very subtle nuances in model assumptions, e.g., on the level of noise or the smoothness of the objective function. The classical theory (see, e.g., \citealp{KuYi03}) has focused on almost sure convergence (e.g., of $\nabla \mathcal{J}\left(\pmb{\theta}_k\right)$ to zero). The work by \cite{bertsekas2000gradient} gives some of the weakest assumptions for this type of convergence when the objective is non-convex.
\cite{nemirovski2009robust} pioneered a different style of analysis which derived convergence rates. This initial work required weakly convex objectives, but the non-convex setting was variously studied by \cite{ghadimi2013stochastic}, \cite{jain2017non}, and \cite{li2019convergence}. Our analysis belongs to this overall stream, but derives convergence rates in the presence of a projection operator, with weaker assumptions than past papers studying similar settings.

We first note that the specific objective function optimized by DBN-RL was shown to be $L$-smooth in Lemma \ref{lemma: L-smooth}. A direct consequence of this property (Lemma 1.2.3 of \citealp{nesterov2003introductory}), used in our analysis, is that
\begin{equation}\label{eq:nesterov}
|\mathcal{J}(\pmb{y}) - \mathcal{J}(\pmb{x})-\langle\nabla \mathcal{J}(\pmb{x}), \pmb{y}-\pmb{x}\rangle| \leq L\lVert \pmb{x}-\pmb{y}\rVert^2.
\end{equation}
For notational simplicity, we use $\E$ to refer to the expectation over the posterior distribution $p\left(\pmb{w}|\mathcal{D}\right)$, since this distribution is the only source of randomness in the algorithm. Also let
\begin{equation}\label{eq:eventA}
A_k=\left\{\pmb\theta_k+\eta_k\nabla \widehat{\mathcal{J}}(\pmb\theta_k)\in \mathbb{C}\right\}
\end{equation}
be the event that the updated gradient remains in the feasible region (i.e., we do not need to project it back onto $\mathbb{C}$) after the $k$-th episode.

Three additional assumptions are required for the convergence proof.

\begin{assumption}[Boundary condition]\label{assumption 1}
For any $\pmb\theta\in\partial\mathbb{C}$ and model $\pmb{w}\in \mathcal{W}$, there exists a constant $c_0 >0$ such that, when the stepsize $\eta\leq c_0$, we have
%\begin{equation*}
$
\pmb\theta + \eta \frac{\nabla_{\pmb\theta} {J}(\pmb\theta;\pmb{w})}{\Vert\nabla_{\pmb\theta} {J}(\pmb\theta;\pmb{w})\Vert}\in \mathbb{C}.
%\end{equation*}
$
\end{assumption}

Assumption~\ref{assumption 1} suggests that, at any point on the boundary of the feasible region, the gradient always points toward the interior of $\mathbb{C}$. For example, one might consider a situation where the feasible region is large enough to include all local maxima. This assumption is reasonable for biomanufacturing, because in such an application $\mathbb{C}$ should be defined based on FDA regulatory requirements; any batch outside the required region will induce a massive penalty.

\begin{assumption}[Noise level]\label{assumption 2}
For any $\pmb\theta \in \mathbb{C}$, we have $\E[\Vert\nabla\widehat{\mathcal{J}}(\theta)-{\nabla \mathcal{J}}(\theta)\Vert^4]^{1/4}\leq \sigma$.
\end{assumption}

Classical stochastic approximation theory \citep{KuYi03} assumes uniformly bounded variance, i.e., $\E\left[\Vert\nabla\widehat{\mathcal{J}}(\theta)-{\nabla \mathcal{J}}(\theta)\Vert^2\right]\leq \sigma^2$. This assumption is weaker than Assumption \ref{assumption 2}, but the results will also be weaker since the classical theory is primarily concerned with almost sure convergence only. If we compare against existing work on convergence \textit{rates}, Assumption \ref{assumption 2} is \textit{weaker} than other widely used assumptions such as $\E[\exp{(\Vert\nabla\widehat{\mathcal{J}}(\theta)-{\nabla \mathcal{J}}(\theta)\Vert^2}/\sigma^2)]\leq \exp{(1)}$, used in \cite{nemirovski2009robust} and \cite{li2019convergence}. For non-convex and strongly convex cases, one can also 
find other assumptions, such as boundedness $\Vert\nabla\widehat{\mathcal{J}}(\theta)-{\nabla \mathcal{J}}(\theta)\Vert^2\leq S$ of the stochastic gradient itself \citep{li2019convergence}, or boundedness $\E\left[\Vert\nabla\widehat{\mathcal{J}}(\theta)\Vert^2\right]\leq \sigma^2$ of the second moment of the gradient estimator \citep{shalev2011pegasos,recht2011hogwild}. Assumption~\ref{assumption 2} is weaker than all of these.

\begin{assumption}\label{assumption 3}
The global maximum $\pmb\theta^\star$ lies in the interior of $\mathbb{C}$, that is, $\pmb\theta^\star\in\mathbb{C}\backslash \partial \mathbb{C}$ and $\mathcal{J}(\pmb\theta)\leq \mathcal{J}(\pmb\theta^\star)$ for any $\pmb\theta \in \mathbb{C}$.
\end{assumption}

Assumption~\ref{assumption 3} guarantees that the feasible region is large enough to include the optimal solution, complementing Assumption \ref{assumption 1}, which ensures that the gradient points us back toward the interior if we are ever close to leaving the feasible region.

We can now proceed with the analysis; proofs of the main results and lemmas can be found in  Appendix~\ref{appendix: corollary proof}-\ref{appendix sec: lemma cross grdients}. First, by applying Lemma~\ref{lemma: L-smooth} and Assumption~\ref{assumption 3}, we can show that the true policy gradient $\nabla \mathcal{J}(\pmb{\theta})$ %(not the estimator of this quantity!) 
is bounded on the feasible region.

\begin{corollary}\label{corollary: bounded gradient}
For any  $\pmb{x}\in\mathbb{C}$, we have $\Vert \nabla \mathcal{J}(\pmb{x})\Vert\leq \max_{\pmb{x}\in\mathbb{C}}\Vert \nabla \mathcal{J}(\pmb{x})\Vert\leq G$, where $G = L
\cdot \max_{\pmb{y}\in \mathbb{C}}\Vert\pmb{y} -\pmb{\theta}^\star\Vert^2$.
\end{corollary}

Recalling the definition of $A_k$ in (\ref{eq:eventA}), we let $p\left(A_k\right)$ be the probability that the updated gradient after the $k$-th %episode 
iteration remains in $\mathbb{C}$. We then show that, on almost every sample path, this will happen for all sufficiently large $k$.

\begin{lemma}\label{lemma: p(theta) convergence as} Let Assumptions~\ref{assumption 1}-\ref{assumption 3} hold and suppose $\sum^\infty_{k=1}\eta_k^2<\infty$. Then,
\begin{equation*}
\lim_{k\rightarrow \infty}p(A_k)=\lim_{k\rightarrow \infty}P\left(\pmb\theta_k+\eta_k\nabla \widehat{\mathcal{J}}(\pmb\theta_k)\in \mathbb{C}\right)=1.
\end{equation*}
% In other words, the sequence $\left\{\pmb\theta_k+\eta_k\nabla \widehat{\mathcal{J}} (\pmb\theta_k)\right\}$ converge to feasible region $\mathbb{C}$ in probability.
\end{lemma}

% Then, we can have Lemma~\ref{lemma: cross grdients}.
% In fact, the limit $\lim_{k\rightarrow \infty}p(\pmb\theta_k\in \mathbb{C})=1$ that we have shown in Lemma~\ref{lemma: p(theta) convergence as} implies that for any $\zeta\in (0,1)$, there exist some $k_{\zeta}$ such that all $k\geq  k_0$, $p(\pmb\theta_k)\geq \zeta$. We summarize the statement in the following lemma.
% \begin{corollary}\label{lemma: lower bound of probability} Let $c_0$ be the constant define in \ref{assumption 1} and $\zeta\in (0,1)$.
% There exists a $k_{\zeta}>0$ such that when stepsize is small enough that $\eta_k < \min\{c_0,k_{\zeta}\}$ for all $k>0$, we have $p(\pmb\theta_k)\geq \zeta$.
% \end{corollary}
% \begin{proof}
% Notice that for $\pmb\theta \notin\mathbb{C}$, the projection operator $\Pi_{\mathbb{C}}(\pmb\theta)$ always projects $\pmb\theta$ onto the boundary, i.e. $\Pi_{\mathbb{C}}(\pmb\theta) \in \partial \mathbb{C}$. Then for any $\pmb\theta_i \in\{\pmb\theta_1,\pmb\theta_2,\ldots, \pmb\theta_k,\ldots\}$, if $\pmb\theta_i \notin \mathbb{C}$ then $\pmb\theta_{i+1}\in\mathbb{C}$. It implies that less than half of the estimates from the sequence $\{\pmb\theta_1,\pmb\theta_2,\ldots, \pmb\theta_k\}$ are out of constraint region $\mathbb{C}$. So we have $p(\pmb\theta_k)\geq \frac{1}{2}$.
% \end{proof}

The final convergence result connects $p\left(A_k\right)$ to the averaged expected norm $\frac{1}{K}\sum_{k=1}^K  \E\left[\Vert\nabla \mathcal{J}(\pmb\theta_k)\Vert^2\right]$ of the true policy gradient.

\begin{theorem}\label{theorem:convergence}
Let Assumptions~\ref{assumption 1}-\ref{assumption 3} hold, and suppose that the stepsize satisfies $\sum^\infty_{k=1}\eta_k^2<\infty$. Then,
\begin{equation*}
    \frac{1}{K}\sum_{k=1}^K  \E\left[\Vert\nabla \mathcal{J}(\pmb\theta_k)\Vert^2\right]\leq \frac{\frac{2}{\eta_1} \left(\mathcal{J}^\star -\mathcal{J}(\pmb\theta_1) \right)  +2(G^2+G\sigma)\sum_{k=1}^K(1-p(A_k))^{1/2}+  8L\sigma^2\sum_{k=1}^K\eta_k}{K}.
\end{equation*}
\end{theorem}

Thus, the convergence rate of our algorithm is connected to the behavior of $p\left(A_k\right)$. Combining Lemma \ref{lemma: p(theta) convergence as} with Theorem \ref{theorem:convergence}, we have %it straightforwardly follows that
$ %\begin{equation*}
\frac{1}{K}\sum_{k=1}^K  \E\left[\Vert\nabla \mathcal{J}(\pmb\theta_k)\Vert^2\right]\rightarrow 0.
$ %\end{equation*}

% =============================

\section{Empirical Study}
\label{sec:empiricalStudy}

We present a case application of our approach to multi-phase fermentation of \textit{Yarrowia lipolytica}, a process in which viable cells grow and produce the biological substance of interest. 
%Such operations drive productivity and typically contribute $70-80\%$ of the total production cost \citep{harrison2015bioprocess,straathof2011proportion}. %Section \ref{subsec: mdp formulation} describes the problem context and model specifications; 
{In Section~\ref{subsec: mdp formulation}, we develop a simulator based on real-world experimental data and domain knowledge of bioprocess mechanisms.} In Section \ref{subsec: performance}, we use this simulator to assess the performance of the proposed DBN-RL framework and compare it with a state-of-the art RL benchmark.
{Section~\ref{subsec:empiricalStudyModelRisk} provides additional empirical evidence that there is value in accounting for model risk, especially when the stochastic uncertainty is high and the number of observations is low. In Section \ref{subsec: interpretabiltiy}, we discuss the interpretability of the proposed framework. Finally, in Section \ref{sec:newintegrated}, we present additional empirical results showing how the proposed framework can support integrated biomanufacturing process control.}

%We illustrate the interpretability of the proposed DBN-RL %framework in Section~\ref{subsec: interpretabiltiy} and %further show it can support integrated biomanufacturing %process control in %Appendix~\ref{subsec:integratedUpstreamDownstream}.}
%, with additional issues investigated in Section \ref{subsec:empiricalStudyModelRisk}; finally, Section \ref{subsec: interpretabiltiy} discusses the interpretability of the results.

% ==============
\subsection{{Description of Application and Simulator}}
\label{subsec: mdp formulation}

The simulator is constructed using existing domain knowledge of fermentation process kinetics, as well as 8 batches of real data from lab experiments. Domain knowledge is leveraged by fitting an ODE-based nonlinear kinetic model, with the addition of a Wiener process to represent intrinsic stochastic uncertainty; the details can be found in Appendix~\ref{appendix sec: fermentation kinetics}. Then, to mimic ``real-world data" collection, we generate the data $\mathcal{D}$ consisting of $R$ process observations (replications from the simulator). 

We have a five-dimensional continuous state variable $\pmb{s}=(X_{f},C,S,N,V)$, where $X_f$ represents lipid-free cell mass; $C$ measures citrate, the actual ``product'' to be harvested at the end of the bioprocess, generated by the cells' metabolism; $S$ and $N$ are amounts of substrate (a type of oil) and nitrogen, both used for cell growth and production; and $V$ is the working volume of the entire batch. We also have one scalar continuous CPP action ${a}= F_S$, which represents the feed rate, or the amount of new substrate given to the working cells in one unit of time.

%One can also introduce additional CPPs related to, e.g., {agitation and oxygen flow rates} in the bioreactor, but for simplicity let us omit these from the present discussion (one may assume that these were already well-controlled).

\begin{figure}[t]
	\centering
	\includegraphics[width=1\textwidth]{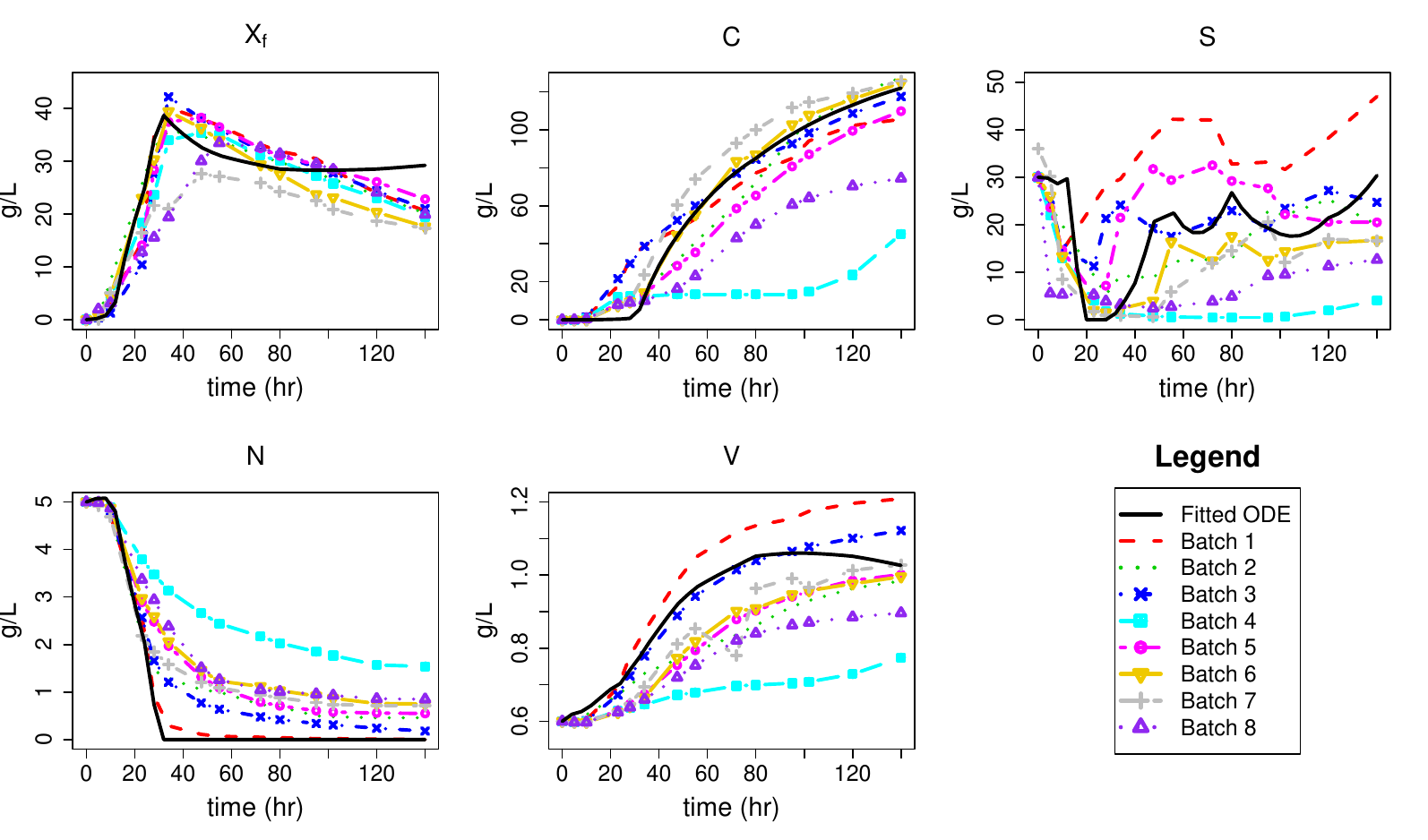}
	\vspace{-0.1in}
	\caption{ODE trajectory (in black) and 8 batches of real lab experiment trajectories. Batches 4 and 8, which have low productivity, are two experiments with low oil feed.}\label{fig:ODE-fitness}
	\vspace{-0.1in}
\end{figure}

%Using 8 batches of real data from fed-batch fermentation %experiments, we first estimate a  nonlinear kinetic model %for the process dynamics, using least squares to fit %model parameters; see the details %. The details of this %model and estimation are given 
%in Appendix~\ref{appendix sec: fermentation kinetics}. %Conceptually, this part of the study proceeds as described in Section \ref{subsec:hybridModel}. 
% IOR: You already said all this earlier.

Figure \ref{fig:ODE-fitness} illustrates the historical trajectories of each CQA from all 8 batches of process observations, as well as the trajectory obtained from the estimated kinetic model.
%the trajectories of the five dimensions of the state variable under each of the 8 real batches as well as under the estimated kinetic model. 
One can see that there is a great deal of variation between real experiments; the given data are not sufficient to estimate intrinsic stochastic uncertainty because each experiment was conducted with very different decision parameters.
Thus, we chose to set the inherent stochastic uncertainty of state $\pmb{s}_t =(X_{ft},C_t,S_t,N_t,V_t)$ at time $t$ as
\begin{equation}\label{eq:kappa}
\sigma(s_t)={\frac{1}{8\kappa}\sum_{i=1}^8 s^{(i)}_t},
\end{equation}
a multiple of the estimated mean over $8$ experiments. In our numerical study, we considered $\kappa\in\left\{10,25,\infty\right\}$, reflecting different levels of stochastic uncertainty which depend on both time and state. We use $\sigma(s_t)$ as the variation parameter in the Wiener process, as described in Appendix~\ref{appendix sec: fermentation kinetics}. %In the empirical study, by setting $\kappa=10, 25, \infty$, we can control the level of process inherent stochastic uncertainty level from high to low.

The initial state $\pmb{s}_1$ was chosen randomly to account for raw material variation. Our process knowledge graph consists of 215 nodes and 770 edges with 36 time measurement steps (proceeding in increments of four hours, up to the harvest time of $140$ hours). There are $35$ transitions, $5$ CQA state nodes (one for each continuous dimension) and one CPP action node per time period, leading to a total of $985$ parameters to estimate using Gibbs sampling. The region $\mathcal{W}$ of valid model parameters was chosen to ensure that each parameter is bounded by a large constant $10^{10}$.

%{The stochastic uncertainty can change with time and also cross different state variables. We incorporate $\sigma(s_t)$ as %variation parameter into the Wiener process; see the simulation stochastic model in Appendix~\ref{appendix sec: fermentation kinetics}. In the %empirical study, by setting $\kappa=10, 25, \infty$, we can control the level of process inherent stochastic uncertainty level from high to low.}

%The dataset $\mathcal{D}$ used to create the posterior distribution was obtained by simulating $R$ trajectories from the kinetic model that was fitted using real data. In this way, we can obtain a closer fit of the Bayesian network to the kinetic model by observing the dynamics of the latter across a wide variety of initializations and feeding profiles. When generating these training data, we chose the feed rate action according to the epsilon-greedy method \citep{SuBa18} to cover more of the state and action space:
This simulator was then used to generate $R$ %additional 
simulated trajectories. In each of these, we chose the feed rate action according to the $\epsilon$-greedy method \citep{sutton2018reinforcement}, which balances exploration and exploitation:
\begin{equation*}
{a}_t = \begin{cases}
    a_t^{h} + \mathcal{N}(0,\bar{a}_t / 10), &  \text{with probability $0.7$},\\
    \mbox{Unif}(0,\max_{t\in\mathcal{T}}\{\bar{ a}_t\}), & \text{with probability $0.3$}.
  \end{cases}
  \end{equation*}
Here, $\bar{{a}}_t$ denotes the maximum feed rate across our 8 real experiments at time $t$, and $a_t^h$ denotes the average feed rate across these experiments. In words, we randomly choose either the average feed rate from the real data (with some noise), or a uniformly distributed quantity between zero and the maximum feed rate.
%Again, these actions are only used to create additional data for training the Bayesian network, and are unrelated to policy optimization.

The reward function at time $t$ was formulated by consultation with biomanufacturing experts, and is based on the citrate concentration $C_t$ and feed rate $a_t$ according to
\begin{equation*}\label{eq: reward function 1}
 r(\pmb{s}_t, {a}_t) = \begin{cases}
 -15 +1.29 C_t, & \text{if $t=H$},\\
 -534.52 {a}_t, & \text{if $0\leq t < H$}.
 \end{cases}
\end{equation*}
This structure reflects the fact that, during the process, the operating cost is primarily driven by the substrate, and %the main source of reward is 
the product revenue is collected at the harvest time.

The linear policy (\ref{eq: linear policy func}) is non-stationary; since there are $35$ state transitions, five state nodes and one action node, the total number of policy parameters is $175$. Based on consultation with biomanufacturing experts, the feasible region $\mathbb{C}$ was chosen by constraining each individual parameter to an interval. The lower and upper bounds are the same for all parameters corresponding to a particular CQA (dimension of the state variable). For cell mass and citrate, we used the interval $[0,0.3]$ as the action (feed rate of substrate) always positively affects cell growth and citrate production. For substrate, we used the interval $[-0.1,0.1]$ because excessively high substrate concentration may have adverse effects. For nitrogen, we used the interval $[-0.1,0.02]$, and for volume, we used $[-0.7,0.5]$. Policy parameters can be negative because excessive nutrients can lead to stresses on living cells, which reduces productivity.

%Finally, we run the DBN-RL method (Algorithm \ref{algo1}). In each episode, $B = 32$ posterior samples were used. The stepsize $\eta_k$ was set using Armijo line search \citep{armijo1966minimization}. We also used the early stopping criterion to terminate the procedure In addition, the early stopping criterion is used to prevent overfitting. Specifically, if the reward received over the past $15$ episodes does not improve, we select the best policy from among these $15$ episodes.

% ===============

\subsection{Performance Comparison}
\label{subsec: performance}

We compare DBN-RL against Deep Deterministic Policy Gradient (DDPG), a state-of-the-art model-free RL algorithm designed for continuous control \citep{lillicrap2015continuous}, as well as with a ``human'' policy inferred from the 8 real experiments (see Figure \ref{fig:ODE-fitness}). Although we are not given an exact policy used by human experts (and it is unlikely that they follow any explicit policy in practice), their observed actions are based on their domain knowledge and so the inferred policy may be quite competitive in some situations. 

% IOR: We haven't mentioned the Q-function anywhere up to this point. No one will understand what this is.

DDPG attempts to learn the value of a policy and simultaneously improve it. The value of a policy is modeled using a deep network with a multi-layer feed-forward architecture which evaluates each state-action pair. {Because the DDPG method was designed to find stationary policies, the time $t$ has to be added as an extra dimension into the state variable.} The deep network contains a 7-dimensional state input layer followed by a dense layer with 64 neurons. 
The 64-dimensional dense layer is concatenated with an (1-dimensional) action input layer and then followed by a 8-dimensional dense layer; the output layer is 1-dimensional with a linear activation function. 
% This critic model has 1,049 parameters in total with 512 in the state input layer, 528 in the second layer and 9 in the last layer. 
The policy improvement step uses another, two-layer neural network with the state input followed by an 8-dimensional intermediate layer and a 1-dimensional action output layer with a ``tanh" activation function. 
% This policy model has 145 parameters in total with 128 in the first layer and 17 in second layer.

%The actions chosen by each policy are evaluated by using the simulator developed in Section~\ref{subsec: mdp formulation}.
%Both DBN-RL and DDPG use the same set of training data. For the human expert policy, we only consider $R=8$ as these are all the human experts we have.

%In biomanufacturing, a practitioner relying on human experts would be unlikely to have access to a much larger number than this. However, as discussed in Sections \ref{subsec:hybridModel}-\ref{sec:lineargaussian}, the two RL algorithms can obtain more training data by simulating from the kinetic model.

\begin{table}[t]
\small
\centering
\caption{Reward and final citrate titer of DBN-RL, DDPG and human policy (30 macro-replications).}
\label{table: policy comparison}
\scalebox{1}{
\begin{tabular}{@{}c|c|cc|cc|cc|cc|cccc@{}}
\toprule
\multicolumn{2}{c|}{Algorithms}                                               & \multicolumn{4}{c|}{DBN-RL}                                    & \multicolumn{4}{c|}{DDPG}                                      & \multicolumn{4}{c}{Human}                                            \\ \midrule
\multicolumn{1}{c|}{\multirow{2}{*}{Sample Size}} & \multirow{2}{*}{$\kappa$} & \multicolumn{2}{c|}{Reward} & \multicolumn{2}{c|}{Titer (g/L)} & \multicolumn{2}{c|}{Reward} & \multicolumn{2}{c|}{Titer (g/L)} & \multicolumn{2}{c|}{Reward}        & \multicolumn{2}{c}{Titer (g/L)} \\
\multicolumn{1}{c|}{}                             &                           & Mean          & SE          & Mean             & SE            & Mean         & SE           & Mean            & SE             & Mean   & \multicolumn{1}{c|}{SE}   & Mean            & SE            \\ \midrule
\multirow{3}{*}{$R=8$}                            & 10                        & 103.44        & 2.50        & 101.15           & 1.11          & \multicolumn{2}{c|}{-}      & \multicolumn{2}{c|}{-}           & 113.47 & \multicolumn{1}{c|}{4.26} & 102.12          & 3.88          \\
                                                  & 25                        & 108.13        & 2.06        & 103.15           & 1.11          & \multicolumn{2}{c|}{-}      & \multicolumn{2}{c|}{-}           & 116.19 & \multicolumn{1}{c|}{4.03} & 103.71          & 3.45          \\
                                                  & $\infty$                  & 114.96        & 1.73        & 105.93           & 0.94          & \multicolumn{2}{c|}{-}      & \multicolumn{2}{c|}{-}           & 118.76 & \multicolumn{1}{c|}{3.56} & 104.86          & 3.10          \\ \midrule
\multirow{3}{*}{$R=15$}                           & 10                        & 119.75        & 1.60        & 110.62           & 0.84          & 17.65        & 11.53        & 39.16           & 8.27           & \multicolumn{2}{c|}{-}             & \multicolumn{2}{c}{-}           \\
                                                  & 25                        & 120.11        & 1.64        & 111.62           & 0.93          & 23.35        & 10.83        & 47.11           & 8.40           & \multicolumn{2}{c|}{-}             & \multicolumn{2}{c}{-}           \\
                                                  & $\infty$                  & 121.35        & 1.75        & 111.98           & 0.90          & 20.65        & 10.24        & 49.04           & 7.12           & \multicolumn{2}{c|}{-}             & \multicolumn{2}{c}{-}           \\ \midrule
\multirow{3}{*}{$R=50$}                           & 10                        & 126.94        & 2.15        & 115.29           & 1.19          & 24.65        & 9.00         & 50.33           & 6.41           & \multicolumn{2}{c|}{-}             & \multicolumn{2}{c}{-}           \\
                                                  & 25                        & 128.34        & 1.23        & 115.13           & 4.52          & 20.44        & 8.04         & 45.93           & 6.10           & \multicolumn{2}{c|}{-}             & \multicolumn{2}{c}{-}           \\
                                                  & $\infty$                  & 131.97        & 0.47        & 116.05           & 0.41          & 21.65        & 7.18         & 49.33           & 5.45           & \multicolumn{2}{c|}{-}             & \multicolumn{2}{c}{-}           \\ \midrule
\multirow{3}{*}{$R=100$}                          & 10                        & 128.90        & 1.25        & 115.89           & 0.60          & 31.36        & 10.98        & 45.21           & 8.24           & \multicolumn{2}{c|}{-}             & \multicolumn{2}{c}{-}           \\
                                                  & 25                        & 130.73        & 0.95        & 116.02           & 0.59          & 32.89        & 10.14        & 50.23           & 8.60           & \multicolumn{2}{c|}{-}             & \multicolumn{2}{c}{-}           \\
                                                  & $\infty$                  & 131.92        & 0.73        & 116.35           & 0.45          & 35.89        & 9.88         & 51.81           & 7.85           & \multicolumn{2}{c|}{-}             & \multicolumn{2}{c}{-}           \\ \midrule
\multirow{3}{*}{$R=400$}                          & 10                        & 130.40        & 0.57        & 116.45           & 0.38          & 37.23        & 9.12         & 50.43           & 8.28           & \multicolumn{2}{c|}{-}             & \multicolumn{2}{c}{-}           \\
                                                  & 25                        & 131.01        & 0.37        & 116.75           & 0.30          & 37.43        & 9.95         & 52.43           & 8.87           & \multicolumn{2}{c|}{-}             & \multicolumn{2}{c}{-}           \\
                                                  & $\infty$                  & 133.04        & 0.12        & 118.12           & 0.14          & 39.77        & 9.23         & 55.16           & 9.29           & \multicolumn{2}{c|}{-}             & \multicolumn{2}{c}{-}           \\ \midrule
$R=3000$                 & $\infty$ & \multicolumn{2}{c|}{-} & \multicolumn{2}{c|}{-} & 119.28 & 0.84 & 104.78 & 0.74 & \multicolumn{2}{c|}{-} & \multicolumn{2}{c}{-}\\
\bottomrule
\end{tabular} }
\end{table}

Table \ref{table: policy comparison} reports the total expected reward, as well as the final citrate titer production (amount of product harvested at the end of $140$ hours) obtained by each policy. A total of 30 macroreplications was conducted. %; this number was sufficient to obtain statistically significant performance estimates. 
We see that, under the smallest sample size $R=8$, the human policy achieves the best performance, as might be expected from domain experts. However, with only a few additional samples from the simulator %kinetic model 
(i.e., $R=15$), DBN-RL can improve on the human policy, with additional improvement as $R$ increases.

%\textcolor{blue}{Although it is certainly possible that the human policy would also improve with more samples, the cost of additional human %experts would be prohibitive for a practitioner (Ilya, please check this sentence)}.
%On the other hand, DBN-RL can use the same starting number of $8$ experts to calibrate the kinetic model, and then use additional trajectories simulated from that model as ``data'' that can help us learn a better policy.

The DDPG policy is much less efficient than DBN-RL, because it is entirely model-free and cannot benefit from domain knowledge of the structured interactions between CPPs and CQAs. For demonstration purposes, we report the performance of DDPG for a very large sample size of $R = 3000$, simply to show that this policy is indeed able to do reasonably well when given enough data. However, even these numbers are not as good as what DBN-RL can achieve with just $R=15$ samples. {In some runs, we observed that DDPG diverged during training, even after several thousand episodes. Similar convergence issues have been reported by \cite{matheron2019problem}. Moreover, in some cases DDPG converged to the boundaries of the action space rather than the true optimum, perhaps because the boundaries are local optima.} 
%In this we see a major advantage of our approach, namely, the ability to use problem structure from kinetic models to learn much more efficiently. 
%The results show that DBN-RL is a robust and %sample-efficient algorithm.
%that can achieve human level control given very small amounts of real experimental data.

%An interesting observation is that during the training, DDPG converges to either true optimum or two boundaries of action space, i.e. 0 and 0.02 %for all $t\in\mathcal{T}$, which indicates that the boundaries are two local optima where DDPG often gets stuck. It is also the reason why DDPG %suffers from high variation in its convergence.

\subsection{Model Risk and Computational Efficiency}
\label{subsec:empiricalStudyModelRisk}

We present additional empirical results demonstrating the benefits of directly building model risk into the DBN, as well as the computational savings obtained by using the proposed nested backpropagation procedure (Algorithm \ref{algo: NBP}).
First, we compare DBN-RL as stated in Algorithm \ref{algo1} with another version of DBN-RL that ignores model risk. In this second version, the posterior samples $\pmb{w}^{(b)}$ are replaced by a single point estimate (i.e., posterior mean) of the model parameters. %This second version ignores model risk. 
The results in Table \ref{table: Posterior  vs point estimator} show that we can obtain a better policy when we account for model risk, especially when the amount of data is small %{($R=8$)} 
and stochastic uncertainty is high.  %{($\kappa=5$)}. 
%This can arise in biomanufacturing when a reliable kinetic model is not available.
%Table \ref{table: Posterior  vs point estimator} shows that posterior sampling is more efficient for low values of $R$. When there is more training data, the accuracy of the point estimate increases and so performance becomes comparable according to a pairwise $t$-test of the respective average performance values.

% \begin{table}[t]
% \centering
% \caption{Performance of DBN-RL with and without model risk ($\kappa=10$, 30 macro-replications).}\label{table: Posterior  vs point estimator}
% \scalebox{0.8}{
% \begin{tabular}{@{}cc|c|c|c@{}}
% \toprule
% Sample size           & Metrics     & DBN-RL with MR & DBN-RL ignoring MR & $p$-value     \\ \midrule
% \multirow{2}{*}{$R=8$}  & Reward      & 103.44 (2.50)  & 95.39 (2.19)       & 0.02             \\
%                       & Titer (g/L) & 101.15 (1.11)  & 92.64 (0.96)       & \textless{}0.001 \\ \midrule
% \multirow{2}{*}{$R=15$} & Reward      & 119.75 (1.60)  & 118.35 (1.81)      & 0.56             \\
%                       & Titer (g/L) & 118.35 (0.84)       & 109.12 (0.94)      & 0.24             \\ \midrule
% \multirow{2}{*}{$R=50$} & Reward      & 126.94 (2.15)  & 129.16 (1.37)      & 0.39             \\
%                       & Titer (g/L) & 115.29 (1.19)  & 115.08 (0.60)      & 0.88             \\ \bottomrule
% \end{tabular}
% }
% \end{table}

\begin{table}[t]
\centering
\caption{Performance of DBN-RL with and without model risk ($\kappa=5,10$ and 30 macro-replications).}\label{table: Posterior  vs point estimator}
\scalebox{1}{
\begin{tabular}{@{}l|l|c|c|c|c@{}}
\toprule
$\kappa$              & Sample size             & Metrics     & DBN-RL with MR & DBN-RL ignoring MR & $p$-value                         \\ \midrule
\multirow{4}{*}{$10$} & \multirow{2}{*}{$R=8$}  & Reward      & 103.44 (2.50)  & 95.39 (2.19)       & \textless 0.001 \\
                      &                         & Titer (g/L) & 101.15 (1.11)  & 96.64 (0.96)       & \textless0.001 \\ \cmidrule(l){2-6} 
                      & \multirow{2}{*}{$R=15$} & Reward      & 119.75 (1.60)  & 118.35 (1.81)      & 0.56                              \\
                      &                         & Titer (g/L) & 118.35 (0.84)  & 109.12 (0.94)      & \textless 0.001 \\ \midrule
\multirow{4}{*}{$5$}  & \multirow{2}{*}{$R=8$}  & Reward      & 102.91 (2.45)  & 94.06 (2.53)       & 0.01                              \\
                      &                         & Titer (g/L) & 100.29 (1.25)  & 95.08 (1.46)       & 0.01                              \\ \cmidrule(l){2-6} 
                      & \multirow{2}{*}{$R=15$} & Reward      & 118.92 (1.39)  & 114.74 (1.68)      & 0.06                            \\
                      &                         & Titer (g/L) & 115.29 (1.19)  & 105.99 (0.98)      & \textless 0.001 \\ \bottomrule
\end{tabular}
}
\end{table}
Next, we compare the computational cost of DBN-RL with and without nested backpropagation. When Algorithm \ref{algo: NBP} is not used, the policy gradient is computed using brute force. Table \ref{table: computation comparison} reports mean computational time in minutes (averaged over 30 macro-replications) for different sample sizes $R$ and time horizons $H$. For reference, we also report the training time for the dynamic Bayesian network (i.e., the Bayesian inference calculations, which are unrelated to policy optimization). It is clear that the brute-force method scales very poorly with the process complexity $H$. In practice, $H$ tends to be much greater than the sample size $R$, meaning that nested backpropagation is much more scalable.
{When $R=15,100,400$ and $H=36$, the computation time (mean $\pm$ SE) of DDPG  %($H=36$) 
is $6.1 \pm 0.4$, $13.1 \pm 0.8$, and $48.2 \pm 3.0$
minutes, % respectively, 
which is comparable with the time taken by DBN-RL with NBP.
%$13.2 \pm 0.5$, $26.7 \pm 0.9$ and $69.8 \pm 1.4$ minutes.
%
%The overall averaged computation time of DBN-RL is slightly higher than that of DDPG by 7.1 ($R=15$), 14.5 ($R=100$) and 21.6 ($R=400$) minutes respectively. The overall computation time of DBN-RL is comp
%our NBP policy optimization is faster than DDPG and the major computation comes from the Bayesian inference of the DBN model.
}

\begin{table}[b]
\centering
\footnotesize
\caption{Computational time (in minutes) of DBN-RL with and without nested backpropagation.}
\label{table: computation comparison}
\scalebox{1}{
\begin{tabular}{@{}l|ccc|ccc|ccc@{}}
\toprule
\multirow{2}{*}{Horizon} & \multicolumn{3}{c|}{$R=15$}                                                                              & \multicolumn{3}{c|}{$R=100$}                                                                       & \multicolumn{3}{c}{$R=400$}                                                                        \\ \cmidrule(l){2-10}
                         & \begin{tabular}[c]{@{}c@{}}Training\end{tabular} & NBP       & \multicolumn{1}{l|}{Brute Force} & \begin{tabular}[c]{@{}c@{}}Training\end{tabular} & \multicolumn{1}{c}{NBP} & Brute Force  & \begin{tabular}[c]{@{}c@{}}Training\end{tabular} & \multicolumn{1}{c}{NBP} & Brute Force  \\ \midrule
$H=8$                      & 1.1 (0.1)                                               & 0.8 (0.1) & 5.9 (0.2)                        & 3.9 (0.1)                                               & 0.8 (0.1)               & 5.6 (0.2)    & 14.9 (0.3)                                              & 0.8 (0.1)               & 6.1 (0.2)    \\
$H=15$                     & 2.4 (0.7)                                               & 2.1 (0.4) & 27.4 (1.2)                       & 8.7 (0.3)                                               & 2.4 (0.3)               & 26.9 (1.2)   & 31.5 (2.3)                                              & 2.0 (0.4)               & 28.1 (1.1)   \\
$H=36$                    & 4.1 (0.2)                                               & 9.1 (0.3) & 302.3 (5.3)                     & 17.9 (0.1)                                              & 9.7 (0.8)               & 312.3 (5.6) & 59.7 (0.6)                                              & 10.1 (0.7)               & 310.5 (6.1) \\ \bottomrule
\end{tabular}  }
\end{table}

% --------------------------------------

\subsection{Interpretability of DBN-RL} \label{subsec: interpretabiltiy}

A well-known limitation of model-free reinforcement learning techniques (such as those based on deep neural networks) is their lack of interpretability. This can be a serious concern, because the average value of a single batch of bio-drug product is worth over $\$1$million. The proposed DBN-RL framework can overcome this limitation.
%Human experts are closely involved in practical implementation, and require explanations of the features used in a model or the decisions made by a policy. 
%These requirements can be met by our approach. 
First, the network structure provides an intuitive visualization of the quantitative associations between CPPs and CQAs. Second, the policy learned from DBN-RL behaves in a way that is understandable to human experts, as demonstrated by the following example.

\begin{figure}[ht]
\begin{subfigure}{0.92\textwidth}
  \centering
  \includegraphics[width=0.92\linewidth]{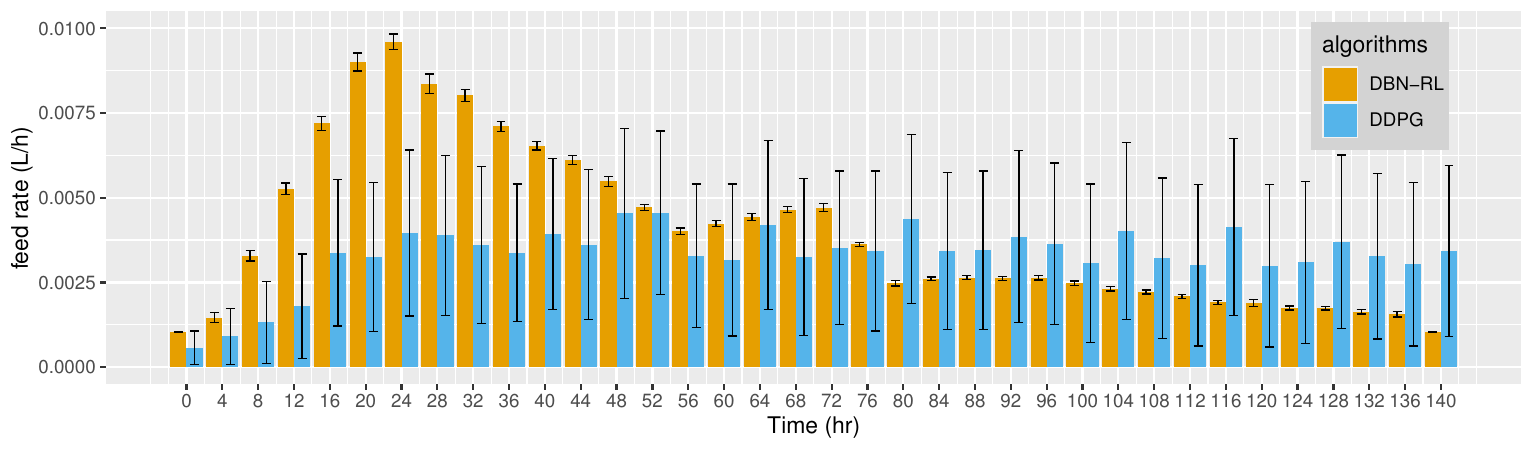}
  \vspace{-0.1in}
  \caption{\small Feeding profiles of DBN-RL and DDPG with 95\% confidence intervals. (R$=400$)}
  \label{fig: RL policy feeding}
\end{subfigure}
\begin{subfigure}{0.92\textwidth}
  \centering
  \includegraphics[width=.92\linewidth]{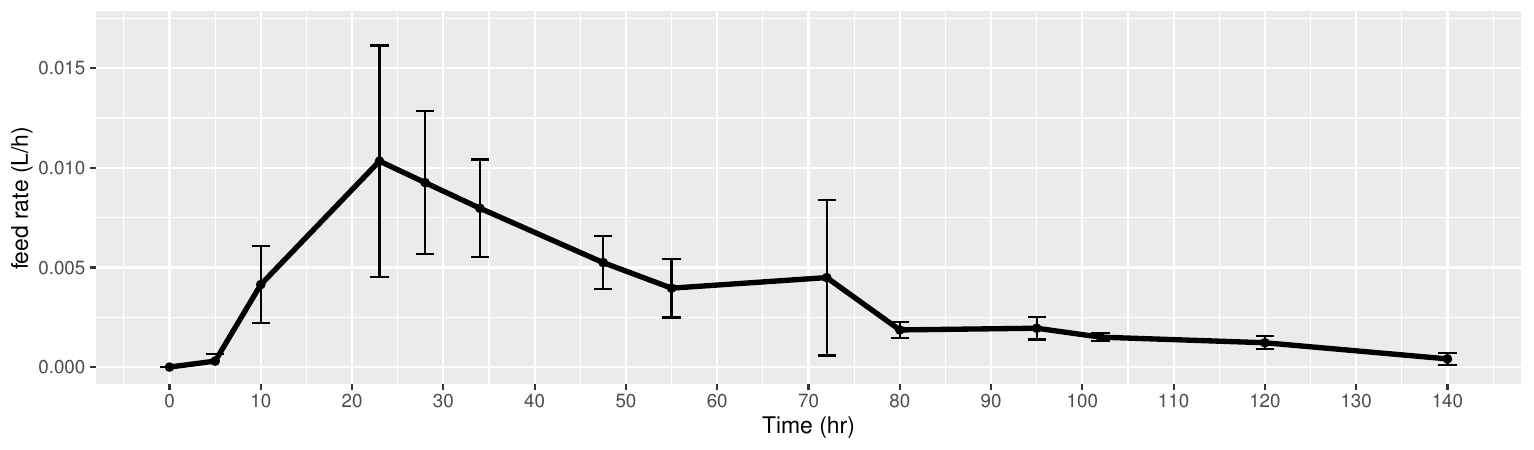}
  \vspace{-0.1in}
  \caption{\small{Feeding profile from human experts: averaged trajectory with 95\% confidence intervals.}}
  \label{fig: human policy fedding}
\end{subfigure}
  \vspace{0.05in}
 \caption{Feeding profiles obtained from (a) RL algorithms and (b) human experts.
 }
\label{fig:feeding profiles}
\vspace{-0.1in}
\end{figure}

We fix an initial state $\pmb{s}_1 =(0.05, 0, 30, 5, 0.6)$ and simulate the actions (feeding rates) taken over time by following the policies obtained from DBN-RL, DDPG, and human experts. For each time $t$, we then average the action $a_t$ and plot these averages over time to obtain a \textit{feeding profile} for each type of policy. These profiles are shown in Figure \ref{fig:feeding profiles}. The policies learned by DBN-RL suggest starting out with low initial feeding rates, because the high initial substrate ($S_1=30$ g/L) is sufficient for biomass development (providing enough carbon sources for cell proliferation and early citrate formation). DBN-RL then ramps up feeding rapidly until the amount of substrate reaches approximately $10$ g/L, when the biomass growth rate is highest. After $24$ hours, DBN-RL begins to reduce the feeding rate, and generally continues doing so until the end of the process. This occurs because the life cycle of cells moves from the growth phase to the production phase at around $24$ hours, and less substrate is required to support production. Comparing with Figure~\ref{fig: human policy fedding}, we find that the feeding profile of DBN-RL has a very similar curve to that of the human expert policy, and even peaks around the same time. On the other hand, DDPG does not supply enough substrate for biomass growth during the first $30$ hours, lowering the cell growth rate; it then attempts to compensate later on, but this is ineffective because of intensive inhibition from high substrate concentration.

\begin{comment}
Finally, Figure \ref{fig:MDP-BN} visualizes the structure and parameters of the policy-augmented Bayesian network during the first $12$ hours of fermentation. Circle nodes represent CQAs while ellipses represent CPPs (feed rates) at different times, with the size of each node indicating the conditional variance of each dimension. Black-colored edges represent the DBN model parameters $\pmb{w}$, and red-colored edges represent the policy parameters $\pmb{\theta}$. The edge weights (coefficients) are reported in the table, and quantify the impact of one factor on another. For example, the coefficient from oil to feed rate is $0.22$, which means that a $1$ unit increase in oil at the $8$th hour is expected to lead to an increase of $0.22$ units in the feed rate at the $12$th hour. Thus, the practitioner is not simply being told how much to feed the biomass, but can also see the reasons for the recommendation, decomposed between different CQAs.

\begin{figure}[t]
	\centering
	\includegraphics[width=1\textwidth]{BN-MDP-v2.png}
	\caption{Visualization of policy-augmented Bayesian network during first 12 hours.
	}\label{fig:MDP-BN}
\end{figure}
\end{comment}

% \subsection{{SV-based Predictive Analysis for Interpretability}}
{
Next, we illustrate how the Shapley value analysis of Section \ref{eq.SV} helps with interpretability. We consider two representative scenarios, namely batches 5 and 8 from Figure \ref{fig:ODE-fitness},
%with state and action inputs $X=(\pmb{s}_t,\pmb{a}_t)$ with historical data size $R=15$ and stochastic uncertainty $\kappa=10$ 
with $R=15$ observations and stochastic uncertainty $\kappa=10$. At the current time $t=15$ (60 hr), we have $\mathcal{O}_t=\{X_{f,t},C_t,S_t,N_t,V_t,F_{St}\}$.
We use the expected Shapley value
$\bar{\mbox{Sh}}\left(C_H|x\right)$ to quantify the contribution of each input $x\in \mathcal{O}_t$ to the predicted expected citrate concentration at the end of the fermentation process.
 %i.e., the concentration of variables in $\mathcal{O}_t=\{X_{ft},C_t,S_t,N_t,V_t,F_{St}\}$, to the expected citrate productivity at end of the fermentation, namely $C_H$; 
%(sample average over the training dataset) to the model output for this prediction. %The DBN-RL are trained using a dataset of size $R=15$ and $\kappa=10$. 
 %Let $X=(\pmb{s}_t,\pmb{a}_t)$ denote the random vector of state and action. 
%Thus the expected value of the model output $\E[f(X)]=\E[\E[\pmb{s}_H|\pmb{s}_t, \pmb{a}_t)]]=\mu_H^C=109.568$ g/L, where $\mu_H^C$ is the mean of citrate concentration at harvest time step $H$. 
}

\begin{figure}[ht]
\begin{subfigure}{0.49\textwidth}
  \centering
  \includegraphics[width=0.96\linewidth]{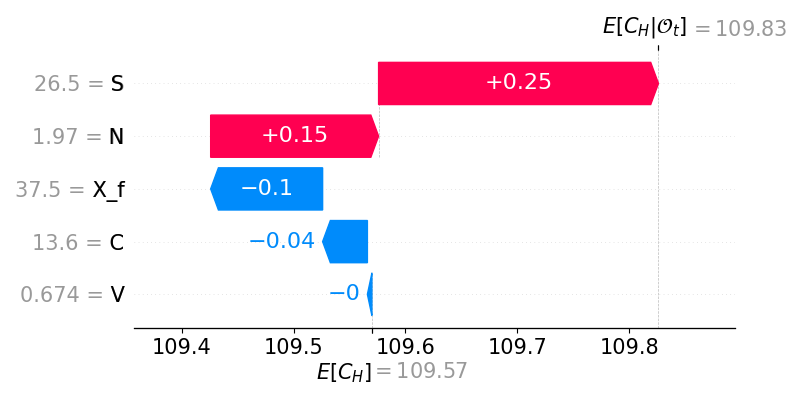}
  \vspace{-0.1in}
  \caption{\small SV score for observation of Batch 5.}
  \label{fig: SV 1}
\end{subfigure}
\begin{subfigure}{0.49\textwidth}
  \centering
  \includegraphics[width=0.96\linewidth]{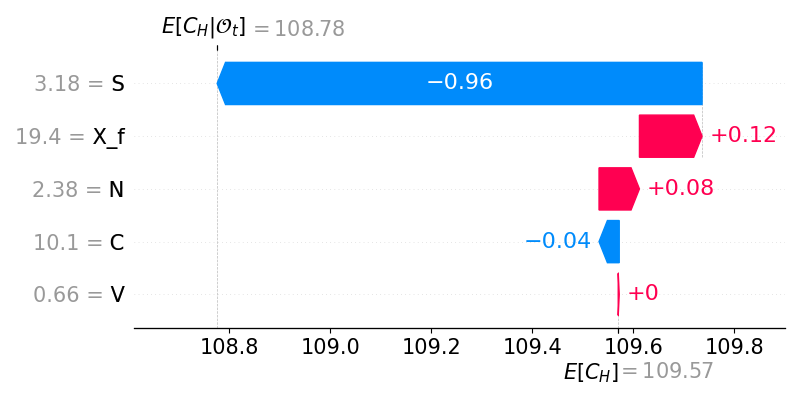}
  \vspace{-0.1in}
  \caption{\small{SV score for observation of Batch 8.}}
  \label{fig: SV 2}
\end{subfigure}
  \vspace{0.05in}
 \caption{{Contribution of each dimension of the state variable to the model output $f(x)$ at 60 hours (corresponding to $t=15$). }
 }
\label{fig: SV importance}
\vspace{-0.1in}
\end{figure}

{
Figure~\ref{fig: SV importance} shows how each input $x\in \mathcal{O}_t$
%CQA and CPP 
contributes to pushing the model output from $\mu_H^C=\E[C_H]$ to $\E[C_H|\mathcal{O}_t]$. Positive contributions are in red, while negative contributions are in blue. In Case (a) corresponding to Batch~5, %(Batch 5 in the real experiment), 
given the input observation
$\mathcal{O}_t= (X_{ft},C_t,S_t,N_t,V_t)=(37.5, 13.6, 26.5, 1.97, 0.674)$
% $(X_f,C,S,N,V,F_S)=(19.4, 10.1, 3.18, 2.38, 0.66,0.0073)$ 
at 60 hours ($t=15$), the model predicts a final citrate concentration of {109.83} g/L under the DBN-RL policy. The main factor that pushes the prediction higher is the current substrate concentration $S_t=26.5$ g/L and nitrogen $N_t=1.97$ g/L. The high cell mass $X_{ft}$ shows a slight negative effect on citrate production because citrate formulation is inhibited in high cell density.
%In fact, the citrate production will be negatively affected when cell density is either too low or too high. Obviously, there are less working cells to produce the citrate when the cell density is low. However, on the opposite, the product formulation will be inhibited in a high cell density environment. 
%Such effect is represented by the term which is in line with the term ($1/(1+X_f/K_{iX})$) in the kinetic model in Appendix~\ref{appendix sec: fermentation kinetics}. 
In Case~(b), the dominant negative contribution comes from low substrate concentration $S_t=3.18$ g/L, which is expected to cause a $0.96$ g/L reduction in the final citrate concentration. According to scientists, this result agrees with their experiment outcome: Batch 8 was known to have low oil feed (see Figure \ref{fig:ODE-fitness}) which negatively impacted citrate productivity. The cell mass $X_{ft}=19.4$ and nitrogen $N=2.38$ increase the expected final citrate concentration by 0.12 g/L and 0.08 g/L respectively. In this way, we see how Shapley value analysis can be integrated with the DBN framework to quantify the effect of each individual component of the state variable on long-term outcomes.
%As we can see, our model use the irregular contribution of some observations to real-time monitor the bioprocess and detect the potential risk. %In both cases, the volume, as expected, does not have much impact on the production of citrate.
}
% --------------------------------------

\subsection{{Applying DBN-RL to Integrated Manufacturing Process Control}}\label{sec:newintegrated}

We present additional numerical results to illustrate how DBN-RL can support integrated biomanufacturing process control consisting of both upstream (fermentation) and downstream (purification) operations. The output of fermentation is a batch mixture containing both the product of interest and a significant amount of undesired impurity derived from the host cells and fermentation medium. {This mixture then undergoes purification, which consists of one centrifugation and two precipitation steps.}

The fermentation process is as described in Section~\ref{subsec: mdp formulation}. {The downstream operation uses a two-dimensional state $\pmb{s}=\left(\log P,\log I\right)$, where $ (P, I)=(C,  X_{f}+S+N)$ denote the respective concentration of citrate product and aggregated impurity. The log transformation is used to make it easier to linearize the process dynamics. %The centrifugation step connects the output $(X_{f},C,S,N)$ of fermentation to the input $(P,I)$ of purification. 
Because the particle sizes of whole cells or cell debris are relatively large, centrifugation yields good separation of $X_f$ and has the output state $ (P, I)=(C, \xi  X_{f}+S+N)$, where $\xi=0.01$ denotes the separation efficiency.} 
%The downstream operation uses a two-dimensional continuous state variable $\pmb{s}=(\log P,\log I)$. %These quantities are calculated from the output $(X_f,C,S,N)$ of upstream fermentation; specifically, $P$ is obtained from $C$ and $I$ is calculated from $X_f,S,N$. 
{Both precipitation steps involve} one scalar continuous CPP action $a=\log\zeta$, where $\zeta$ represents ammonium sulphate saturation. {The complete specification of the process dynamics and state transition for purification is deferred to Appendix~\ref{subsec:integratedUpstreamDownstream}.}

The inherent stochastic uncertainty and initial state are set in the same way as in Section~\ref{subsec: mdp formulation}. The new process knowledge graph (encompassing both upstream and downstream operations) consists of 233 nodes and 818 edges with 39 time measurement steps (36 upstream steps and 3 downstream steps). 
A typical biomanufacturing system often has a trade-off between yield and purity \citep{martagan2018performance}, with higher productivity often corresponding to higher impurity which increases the downstream purification cost. To incorporate these competing considerations, we use the linear reward function
 \begin{numcases}{r(\pmb{s}_t, {a}_t) =}
 -15 +1.3 \log P_t - \log I_t, & \text{if $t=39$}
 ~~~ \mbox{(ending profit)},
 \label{eq.profit}
 \\
 -0.05 a_t, & \text{if $37\leq t\leq 38$} 
 ~~~ \mbox{(purification)},
 \label{eq.purification}
 \\
 -534.52 {a}_t, & \text{if $0\leq t \leq 36$}
 ~~~ \mbox{(fermentation)}.\label{eq.fermentation}
 \end{numcases}
Equations (\ref{eq.purification})-(\ref{eq.fermentation})
reflect the fact that the operating cost is primarily driven by the substrate consumption in the fermentation process and the ammonium sulphate consumption in the purification process. Equation (\ref{eq.profit}) maintains the purity quality requirement on the final product via a linear penalty on $\log I_t$.

\begin{table}[h]
\tiny
\centering
\caption{Mean and standard error (SE) of reward, yield and purity obtained by using DBN-RL, DDPG, and practice-based policy. % (30 macro-replications).
}
\label{table: policy comparison 2}
\begin{tabular}{@{}cc|cccccc|cccccc|cccccc@{}}
\toprule
\multicolumn{2}{c|}{Algorithms}                                                                                         & \multicolumn{6}{c|}{DBN-RL}                                                          & \multicolumn{6}{c|}{DDPG}                                                          & \multicolumn{6}{c}{Practice-based}                                                           \\ \midrule
\multicolumn{1}{c|}{\multirow{2}{*}{\begin{tabular}[c]{@{}c@{}}Sample\\ Size\end{tabular}}} & \multirow{2}{*}{$\kappa$} & \multicolumn{2}{c}{Reward} & \multicolumn{2}{c}{Yield} & \multicolumn{2}{c|}{Purity} & \multicolumn{2}{c}{Reward} & \multicolumn{2}{c}{Yield} & \multicolumn{2}{c|}{Purity} & \multicolumn{2}{c}{Reward} & \multicolumn{2}{c}{Yield} & \multicolumn{2}{c}{Purity} \\ \cmidrule(l){3-20} 
\multicolumn{1}{c|}{}                                                                       &                           & Mean          & SE         & Mean         & SE         & Mean         & SE           & Mean          & SE         & Mean         & SE         & Mean         & SE           & Mean          & SE         & Mean         & SE         & Mean         & SE          \\ \midrule
\multicolumn{1}{c|}{$R=8$}                                                                  & 10                        & -35.05        & 1.65       & 95.08        & 1.99       & 0.79         & 0.02         & -             & -          & -            & -          & -            & -            & -35.32        & 1.35       & 95.48        & 2.26       & 0.81         & 0.01        \\
\multicolumn{1}{c|}{}                                                                       & 25                        & -33.35        & 1.81       & 96.83        & 1.89       & 0.80         & 0.02         & -             & -          & -            & -          & -            & -            & -33.12        & 1.25       & 97.21        & 1.96       & 0.82         & 0.01        \\
\multicolumn{1}{c|}{}                                                                       & $\infty$                  & -32.65        & 1.51       & 97.80        & 1.83       & 0.80         & 0.01         & -             & -          & -            & -          & -            & -            & -33.02        & 1.20       & 97.88        & 2.02       & 0.82         & 0.01        \\
\multicolumn{1}{c|}{\multirow{3}{*}{$R=15$}}                                                & 10                        & -33.05        & 1.70       & 99.08        & 1.41       & 0.80         & 0.02         & -56.50        & 2.01       & 35.13        & 8.01       & 0.64         & 0.05         & -             & -          & -            & -          & -            & -           \\
\multicolumn{1}{c|}{}                                                                       & 25                        & -31.35        & 1.13       & 100.08       & 1.41       & 0.81         & 0.02         & -50.00        & 1.93       & 40.09        & 7.91       & 0.66         & 0.04         & -             & -          & -            & -          & -            & -           \\
\multicolumn{1}{c|}{}                                                                       & $\infty$                  & -31.05        & 1.31       & 99.80        & 1.63       & 0.80         & 0.02         & -48.00        & 2.03       & 44.09        & 7.72       & 0.65         & 0.04         & -             & -          & -            & -          & -            & -           \\
\multicolumn{1}{c|}{\multirow{3}{*}{$R=50$}}                                                & 10                        & -31.88        & 1.37       & 104.88       & 1.63       & 0.81         & 0.02         & -56.50        & 2.01       & 35.13        & 8.01       & 0.64         & 0.05         & -             & -          & -            & -          & -            & -           \\
\multicolumn{1}{c|}{}                                                                       & 25                        & -30.01        & 1.35       & 105.14       & 0.99       & 0.82         & 0.02         & -50.00        & 1.93       & 40.09        & 7.91       & 0.66         & 0.04         & -             & -          & -            & -          & -            & -           \\
\multicolumn{1}{c|}{}                                                                       & $\infty$                  & -29.05        & 1.11       & 106.89       & 1.63       & 0.82         & 0.01         & -48.00        & 2.03       & 44.09        & 7.72       & 0.65         & 0.04         & -             & -          & -            & -          & -            & -           \\
\multicolumn{1}{c|}{\multirow{3}{*}{$R=100$}}                                               & 10                        & -30.91        & 1.24       & 105.78       & 1.41       & 0.81         & 0.01         & -51.04        & 2.05       & 40.13        & 7.91       & 0.65         & 0.05         & -             & -          & -            & -          & -            & -           \\
\multicolumn{1}{c|}{}                                                                       & 25                        & -29.84        & 1.33       & 107.90       & 1.09       & 0.82         & 0.01         & -46.88        & 2.10       & 44.09        & 7.34       & 0.67         & 0.04         & -             & -          & -            & -          & -            & -           \\
\multicolumn{1}{c|}{}                                                                       & $\infty$                  & -29.05        & 1.01       & 107.69       & 0.93       & 0.83         & 0.01         & -46.32        & 1.77       & 45.09        & 7.02       & 0.66         & 0.04         & -             & -          & -            & -          & -            & -           \\
\multicolumn{1}{c|}{\multirow{3}{*}{$R=400$}}                                               & 10                        & -29.79        & 1.52       & 108.13       & 1.48       & 0.84         & 0.01         & -49.04        & 2.08       & 39.91        & 7.91       & 0.63         & 0.05         & -             & -          & -            & -          & -            & -           \\
\multicolumn{1}{c|}{}                                                                       & 25                        & -28.99        & 1.32       & 110.60       & 1.34       & 0.84         & 0.01         & -48.53        & 1.99       & 44.70        & 7.34       & 0.69         & 0.04         & -             & -          & -            & -          & -            & -           \\
\multicolumn{1}{c|}{}                                                                       & $\infty$                  & -28.49        & 1.22       & 113.13       & 1.27       & 0.83         & 0.01         & -45.64        & 1.85       & 45.55        & 7.02       & 0.67         & 0.03         & -             & -          & -            & -          & -            & -           \\ \bottomrule
\end{tabular}
\end{table}

We compare the performance of DBN-RL with DDPG and experimental practice. DDPG was developed for stationary MDP settings and thus does not allow the state space to change from fermentation to purification. To handle this issue, we trained a separate DDPG agent for the purification process using the reward function \eqref{eq.profit}-\eqref{eq.fermentation}. Then, in the integrated biomanufacturing process, the feed rate in fermentation was chosen by following the fermentation DDPG policy obtained in Section~\ref{subsec: performance}, and the purification decision was chosen by following the purification DDPG policy. 
%As we do not have the real experiment data for the purification step, the ``human" policy are unknown. To migrate the problem, we use the practice-based policy as the baseline policy, which choose the purification action based on the literature \cite{varga2001prediction}.

The practice-based policy was designed as follows. In the fermentation step, we use the same ``human'' policy as in Section \ref{subsec: performance}. In the purification step, since we do not have access to laboratory data, we instead implemented the practice-based rule of \cite{varga2001prediction}.

Table \ref{table: policy comparison 2} reports the results of these approaches, including total reward, yield (citrate concentration after purification) and the final product purity ${P_H}/{(P_H+I_H)}$. The results from DBN-RL and DDPG are based on 30 macro-replications. {With} a small sample size $R=8$, DBN-RL is competitive with the practice-based policy, and shows improved performance as $R$ increases. DDPG is much less sample-efficient, similar to what we saw before.

\section{Conclusion}
\label{sec: conclusion}
We have presented new models and algorithms for optimization of control policies on a Bayesian knowledge graph. This approach is especially effective in engineering problems with complex structure derived from physics models. Such structure can be partially extracted and turned into a prior for the Bayesian network. Furthermore, mechanistic models can also be used to provide additional information, which becomes very valuable in the presence of highly complex nonlinear dynamics with very small amounts of available pre-existing data. All of these issues arise in the domain of biomanufacturing, and we have demonstrated that our approach {outperforms a state-of-the-art model-free RL method empirically.}
%can achieve human-level control using as few as $8$ lab experiments, while state-of-the-art model-free methods struggle due to their inability to incorporate known structure in the process dynamics.

As synthetic biology and biotechnology continue to adopt more complex processes for the generation of new drug products, from monoclonal antibodies (mAbs) to cell/gene therapies, data-driven and model-based control will become increasingly important. This work presents compelling evidence that model-based reinforcement learning can provide competitive performance and interpretability in the control of these important systems.

\bibliographystyle{unsrtnat} % outcomment this and next line in Case 1
\bibliography{BN_ref,proj_ref,sensitivity_ref,proposal} % if more than one, comma separated

\begin{thebibliography}{47}
\providecommand{\natexlab}[1]{#1}
\providecommand{\url}[1]{\texttt{#1}}
\expandafter\ifx\csname urlstyle\endcsname\relax
  \providecommand{\doi}[1]{doi: #1}\else
  \providecommand{\doi}{doi: \begingroup \urlstyle{rm}\Url}\fi

\bibitem[Martin et~al.(2021)Martin, Vicente, Beccari, Kellermayer, Koller, Lal,
  Marks, Marova, Mechler, Tapaloaga, et~al.]{martin2021brief}
Donald~K Martin, Oscar Vicente, Tommaso Beccari, Mikl{\'o}s Kellermayer, Martin
  Koller, Ratnesh Lal, Robert~S Marks, Ivana Marova, Adam Mechler, Dana
  Tapaloaga, et~al.
\newblock A brief overview of global biotechnology.
\newblock \emph{Biotechnology \& Biotechnological Equipment}, 35\penalty0
  (sup1):\penalty0 S5--S14, 2021.

\bibitem[Lloyd(2019)]{PharmaAnnualReview2019}
Ian Lloyd.
\newblock Pharma r\&d annual review 2019.
\newblock Technical report, Pharma Intelligence, 2019.

\bibitem[Tsopanoglou and del Val(2021)]{tsopanoglou2021moving}
Apostolos Tsopanoglou and Ioscani~Jim{\'e}nez del Val.
\newblock Moving towards an era of hybrid modelling: advantages and challenges
  of coupling mechanistic and data-driven models for upstream pharmaceutical
  bioprocesses.
\newblock \emph{Current Opinion in Chemical Engineering}, 32:\penalty0 100691,
  2021.

\bibitem[Hong et~al.(2018)Hong, Severson, Jiang, Lu, Love, and
  Braatz]{hong2018challenges}
Moo~Sun Hong, Kristen~A Severson, Mo~Jiang, Amos~E Lu, J~Christopher Love, and
  Richard~D Braatz.
\newblock Challenges and opportunities in biopharmaceutical manufacturing
  control.
\newblock \emph{Computers \& Chemical Engineering}, 110:\penalty0 106--114,
  2018.

\bibitem[O'Brien et~al.(2021)O'Brien, Zhang, Daoutidis, and Hu]{OBrien_2021}
Conor~M. O'Brien, Qi~Zhang, Prodromos Daoutidis, and Wei-Shou Hu.
\newblock A hybrid mechanistic-empirical model for in silico mammalian cell
  bioprocess simulation.
\newblock \emph{Metabolic Engineering}, 66:\penalty0 31--40, 2021.

\bibitem[Martagan et~al.(2016)Martagan, Krishnamurthy, and
  Maravelias]{martagan2016optimal}
Tugce Martagan, Ananth Krishnamurthy, and Christos~T. Maravelias.
\newblock Optimal condition-based harvesting policies for biomanufacturing
  operations with failure risks.
\newblock \emph{IIE Transactions}, 48\penalty0 (5):\penalty0 440--461, 2016.

\bibitem[Martagan et~al.(2018)Martagan, Krishnamurthy, Leland, and
  Maravelias]{martagan2018performance}
Tugce Martagan, Ananth Krishnamurthy, Peter~A Leland, and Christos~T
  Maravelias.
\newblock Performance guarantees and optimal purification decisions for
  engineered proteins.
\newblock \emph{Operations Research}, 66\penalty0 (1):\penalty0 18--41, 2018.

\bibitem[Cintron(2015)]{Ci15}
R.~Cintron.
\newblock \emph{Human Factors Analysis and Classification System Interrater
  Reliability for Biopharmaceutical Manufacturing Investigations}.
\newblock PhD thesis, Walden University, 2015.

\bibitem[Liu et~al.(2013)Liu, Gong, Shen, and Feng]{liu2013modelling}
Chongyang Liu, Zhaohua Gong, Bangyu Shen, and Enmin Feng.
\newblock Modelling and optimal control for a fed-batch fermentation process.
\newblock \emph{Applied Mathematical Modelling}, 37\penalty0 (3):\penalty0
  695--706, 2013.

\bibitem[{Spielberg} et~al.(2017){Spielberg}, {Gopaluni}, and
  {Loewen}]{spielberg2017deep}
S.~P.~K. {Spielberg}, R.~B. {Gopaluni}, and P.~D. {Loewen}.
\newblock Deep reinforcement learning approaches for process control.
\newblock In \emph{2017 6th International Symposium on Advanced Control of
  Industrial Processes (AdCONIP)}, pages 201--206, 2017.

\bibitem[Treloar et~al.(2020)Treloar, Fedorec, Ingalls, and
  Barnes]{Treloar2020deep}
Neythen~J Treloar, Alex~JH Fedorec, Brian Ingalls, and Chris~P Barnes.
\newblock Deep reinforcement learning for the control of microbial co-cultures
  in bioreactors.
\newblock \emph{PLoS Computational Biology}, 16\penalty0 (4):\penalty0
  e1007783, 2020.

\bibitem[Bankar et~al.(2009)Bankar, Kumar, and Zinjarde]{BaKuZi09}
A.~V. Bankar, A.~R. Kumar, and S.~S. Zinjarde.
\newblock Environmental and industrial applications of {Y}arrowia lipolytica.
\newblock \emph{Applied Microbiology and Biotechnology}, 84\penalty0
  (5):\penalty0 847--865, 2009.

\bibitem[Mandenius et~al.(2013)Mandenius, Titchener-Hooker,
  et~al.]{mandenius2013measurement}
Carl-Fredrik Mandenius, Nigel~J Titchener-Hooker, et~al.
\newblock \emph{Measurement, monitoring, modelling and control of
  bioprocesses}, volume 132.
\newblock Springer, Berlin, Heidelberg, 2013.

\bibitem[Rathore et~al.(2011)Rathore, Bhushan, and
  Hadpe]{rathore2011chemometrics}
Anurag~S. Rathore, Nitish Bhushan, and Sandip Hadpe.
\newblock Chemometrics applications in biotech processes: A review.
\newblock \emph{Biotechnology Progress}, 27\penalty0 (2):\penalty0 307--315,
  2011.

\bibitem[Teixeira et~al.(2007)Teixeira, Carinhas, Dias, Cruz, Alves, Carrondo,
  and Oliveira]{teixeira2007hybrid}
A.~P. Teixeira, N.~Carinhas, J.~M.~L. Dias, P.~Cruz, P.~M. Alves, M.~J.~T.
  Carrondo, and R.~Oliveira.
\newblock Hybrid semi-parametric mathematical systems: Bridging the gap between
  systems biology and process engineering.
\newblock \emph{Journal of Biotechnology}, 132\penalty0 (4):\penalty0 418--425,
  2007.

\bibitem[Gunther et~al.(2009)Gunther, Conner, and Seborg]{gunther2009process}
Jon~C Gunther, Jeremy~S Conner, and Dale~E Seborg.
\newblock Process monitoring and quality variable prediction utilizing pls in
  industrial fed-batch cell culture.
\newblock \emph{Journal of Process Control}, 19\penalty0 (5):\penalty0
  914--921, 2009.

\bibitem[Lu et~al.(2015)Lu, Paulson, Mozdzierz, Stockdale, Versypt, Love, Love,
  and Braatz]{lu2015control}
Amos~E. Lu, Joel~A. Paulson, Nicholas~J. Mozdzierz, Alan Stockdale, Ashlee
  N.~Ford Versypt, Kerry~R. Love, J.~Christopher Love, and Richard~D. Braatz.
\newblock Control systems technology in the advanced manufacturing of biologic
  drugs.
\newblock In \emph{Proceedings of the IEEE Conference on Control Applications},
  pages 1505--1515, 2015.

\bibitem[Jiang and Braatz(2016)]{jiang2016integrated}
Mo~Jiang and RD~Braatz.
\newblock Integrated control of continuous (bio)pharmaceutical manufacturing.
\newblock \emph{American Pharmaceutical Review}, 19\penalty0 (6):\penalty0
  110--115, 2016.

\bibitem[Lakerveld et~al.(2013)Lakerveld, Benyahia, Braatz, and
  Barton]{lakerveld2013model}
Richard Lakerveld, Brahim Benyahia, Richard~D Braatz, and Paul~I Barton.
\newblock Model-based design of a plant-wide control strategy for a continuous
  pharmaceutical plant.
\newblock \emph{AIChE Journal}, 59\penalty0 (10):\penalty0 3671--3685, 2013.

\bibitem[Zhang et~al.(2019)Zhang, Del Rio-Chanona, Petsagkourakis, and
  Wagner]{zhang2019hybrid}
Dongda Zhang, Ehecatl~Antonio Del Rio-Chanona, Panagiotis Petsagkourakis, and
  Jonathan Wagner.
\newblock Hybrid physics-based and data-driven modeling for bioprocess online
  simulation and optimization.
\newblock \emph{Biotechnology and Bioengineering}, 116\penalty0 (11):\penalty0
  2919--2930, 2019.

\bibitem[Zheng et~al.(2021)Zheng, Ryzhov, Xie, and
  Zhong]{zheng2021personalized}
Hua Zheng, Ilya~O. Ryzhov, Wei Xie, and Judy Zhong.
\newblock Personalized multimorbidity management for patients with type 2
  diabetes using reinforcement learning of electronic health records.
\newblock \emph{Drugs}, 81\penalty0 (4):\penalty0 471--482, Mar 2021.

\bibitem[Silver et~al.(2016)Silver, Huang, Maddison, Guez, Sifre, Van
  Den~Driessche, Schrittwieser, Antonoglou, Panneershelvam, Lanctot,
  et~al.]{silver2016mastering}
David Silver, Aja Huang, Chris~J Maddison, Arthur Guez, Laurent Sifre, George
  Van Den~Driessche, Julian Schrittwieser, Ioannis Antonoglou, Veda
  Panneershelvam, Marc Lanctot, et~al.
\newblock Mastering the game of go with deep neural networks and tree search.
\newblock \emph{Nature}, 529\penalty0 (7587):\penalty0 484--489, 2016.

\bibitem[Petsagkourakis et~al.(2020)Petsagkourakis, Sandoval, Bradford, Zhang,
  and del Rio-Chanona]{petsagkourakis2020reinforcement}
Panagiotis Petsagkourakis, Ilya~Orson Sandoval, Eric Bradford, Dongda Zhang,
  and Ehecatl~Antonio del Rio-Chanona.
\newblock Reinforcement learning for batch bioprocess optimization.
\newblock \emph{Computers \& Chemical Engineering}, 133:\penalty0 106649, 2020.

\bibitem[Pandian and Noel(2018)]{pandian2018control}
B~Jaganatha Pandian and Mathew~Mithra Noel.
\newblock Control of a bioreactor using a new partially supervised
  reinforcement learning algorithm.
\newblock \emph{Journal of Process Control}, 69:\penalty0 16--29, 2018.

\bibitem[Xie et~al.(2020)Xie, Wang, Li, Xie, and Auclair]{xie2020bayesian}
Wei Xie, Bo~Wang, Cheng Li, Dongming Xie, and Jared Auclair.
\newblock Interpretable biomanufacturing process risk and sensitivity analyses
  for quality-by-design and stability control.
\newblock \emph{Naval Research Logistics (NRL)}, 2020.

\bibitem[Doran(2013)]{Pauline_2013}
Pauline~M. Doran.
\newblock \emph{Bioprocess Engineering Principles}.
\newblock Academic Press, London, 2013.

\bibitem[FDA(2009)]{guideline2009pharmaceutical}
FDA.
\newblock Q8 pharmaceutical development.
\newblock Technical report, U.S. Food \& Drug Administration, Silver Spring,
  MD, 2009.

\bibitem[Craven et~al.(2014)Craven, Whelan, and Glennon]{craven2014glucose}
Stephen Craven, Jessica Whelan, and Brian Glennon.
\newblock Glucose concentration control of a fed-batch mammalian cell
  bioprocess using a nonlinear model predictive controller.
\newblock \emph{Journal of Process Control}, 24\penalty0 (4):\penalty0
  344--357, 2014.

\bibitem[Gelfand(2000)]{Ge00}
A.~E. Gelfand.
\newblock Gibbs sampling.
\newblock \emph{Journal of the American Statistical Association}, 95\penalty0
  (452):\penalty0 1300--1304, 2000.

\bibitem[Powell(2011)]{Po11}
W.~B. Powell.
\newblock \emph{Approximate Dynamic Programming: Solving the curses of
  dimensionality (2nd ed.)}.
\newblock John Wiley and Sons, New York, 2011.

\bibitem[Powell(2010)]{powell2010merging}
Warren~B Powell.
\newblock Merging ai and or to solve high-dimensional stochastic optimization
  problems using approximate dynamic programming.
\newblock \emph{INFORMS Journal on Computing}, 22\penalty0 (1):\penalty0 2--17,
  2010.

\bibitem[Goh(1995)]{Go95}
A.~T.~C. Goh.
\newblock Back-propagation neural networks for modeling complex systems.
\newblock \emph{Artificial Intelligence in Engineering}, 9\penalty0
  (3):\penalty0 143--151, 1995.

\bibitem[Jain and Kar(2017)]{jain2017non}
Prateek Jain and Purushottam Kar.
\newblock Non-convex optimization for machine learning.
\newblock \emph{Foundations and Trends® in Machine Learning}, 10\penalty0
  (3-4):\penalty0 142--363, 2017.
\newblock ISSN 1935-8237.

\bibitem[Kushner and Yin(2003)]{KuYi03}
H.~Kushner and G.~Yin.
\newblock \emph{Stochastic approximation and recursive algorithms and
  applications (2nd ed.)}, volume~35.
\newblock Springer, New York, NY, 2003.

\bibitem[Bertsekas and Tsitsiklis(2000)]{bertsekas2000gradient}
Dimitri~P Bertsekas and John~N Tsitsiklis.
\newblock Gradient convergence in gradient methods with errors.
\newblock \emph{SIAM Journal on Optimization}, 10\penalty0 (3):\penalty0
  627--642, 2000.

\bibitem[Nemirovski et~al.(2009)Nemirovski, Juditsky, Lan, and
  Shapiro]{nemirovski2009robust}
Arkadi Nemirovski, Anatoli Juditsky, Guanghui Lan, and Alexander Shapiro.
\newblock Robust stochastic approximation approach to stochastic programming.
\newblock \emph{SIAM Journal on optimization}, 19\penalty0 (4):\penalty0
  1574--1609, 2009.

\bibitem[Ghadimi and Lan(2013)]{ghadimi2013stochastic}
Saeed Ghadimi and Guanghui Lan.
\newblock Stochastic first-and zeroth-order methods for nonconvex stochastic
  programming.
\newblock \emph{SIAM Journal on Optimization}, 23\penalty0 (4):\penalty0
  2341--2368, 2013.

\bibitem[Li and Orabona(2019)]{li2019convergence}
Xiaoyu Li and Francesco Orabona.
\newblock On the convergence of stochastic gradient descent with adaptive
  stepsizes.
\newblock In \emph{The 22nd International Conference on Artificial Intelligence
  and Statistics}, pages 983--992. PMLR, 2019.

\bibitem[Nesterov(2003)]{nesterov2003introductory}
Yurii Nesterov.
\newblock \emph{Introductory lectures on convex optimization: A basic course}.
\newblock Springer, Boston, MA, 2003.

\bibitem[Shalev-Shwartz et~al.(2011)Shalev-Shwartz, Singer, Srebro, and
  Cotter]{shalev2011pegasos}
Shai Shalev-Shwartz, Yoram Singer, Nathan Srebro, and Andrew Cotter.
\newblock Pegasos: Primal estimated sub-gradient solver for svm.
\newblock \emph{Mathematical Programming}, 127\penalty0 (1):\penalty0 3--30,
  2011.

\bibitem[Recht et~al.(2011)Recht, Re, Wright, and Niu]{recht2011hogwild}
Benjamin Recht, Christopher Re, Stephen Wright, and Feng Niu.
\newblock Hogwild!: A lock-free approach to parallelizing stochastic gradient
  descent.
\newblock In \emph{Advances in Neural Information Processing Systems},
  volume~24, 2011.

\bibitem[Sutton and Barto(2018)]{sutton2018reinforcement}
Richard~S Sutton and Andrew~G Barto.
\newblock \emph{Reinforcement learning: An introduction}.
\newblock MIT Press, Cambridge, MA, 2018.

\bibitem[Lillicrap et~al.(2016)Lillicrap, Hunt, Pritzel, Heess, Erez, Tassa,
  Silver, and Wierstra]{lillicrap2015continuous}
Timothy~P. Lillicrap, Jonathan~J. Hunt, Alexander Pritzel, Nicolas Heess, Tom
  Erez, Yuval Tassa, David Silver, and Daan Wierstra.
\newblock Continuous control with deep reinforcement learning.
\newblock In \emph{4th International Conference on Learning Representations,
  {ICLR} 2016, San Juan, Puerto Rico}, 2016.

\bibitem[Matheron et~al.(2019)Matheron, Perrin, and
  Sigaud]{matheron2019problem}
Guillaume Matheron, Nicolas Perrin, and Olivier Sigaud.
\newblock The problem with ddpg: understanding failures in deterministic
  environments with sparse rewards.
\newblock \emph{arXiv preprint arXiv:1911.11679}, 2019.

\bibitem[Varga et~al.(2001)Varga, Titchener-Hooker, and
  Dunnill]{varga2001prediction}
EG~Varga, NJ~Titchener-Hooker, and P~Dunnill.
\newblock Prediction of the pilot-scale recovery of a recombinant yeast enzyme
  using integrated models.
\newblock \emph{Biotechnology and Bioengineering}, 74\penalty0 (2):\penalty0
  96--107, 2001.

\bibitem[Niktari et~al.(1990)Niktari, Chard, Richardson, and
  Hoare]{niktari1990monitoring}
Maria Niktari, Stephen Chard, Phillip Richardson, and Michael Hoare.
\newblock The monitoring and control of protein purification and recovery
  processes.
\newblock In \emph{Separations for Biotechnology 2}, pages 622--631. Springer,
  Dordrecht, 1990.

\bibitem[Boyd et~al.(2004)Boyd, Boyd, and Vandenberghe]{boyd2004convex}
Stephen Boyd, Stephen~P Boyd, and Lieven Vandenberghe.
\newblock \emph{Convex optimization}.
\newblock Cambridge University Press, 2004.

\end{thebibliography}

% CASE 2: BiBTeX used to generate mypaper.bbl (to be further fine tuned)
%\input{mypaper.bbl} % outcomment this line in Case 2

%% Here starts the e-companion (EC)
%%%%%%%%%%%%%%%%%%%%%%%%%%%%%%%%%%%%%%%%%%%%%%%%%%%%%%%%%%
% \ECSwitch

%\ECDisclaimer
%%%%%%%%%%%%%%%%%%%%%%%%%%%%%%%%%%%%%%%%%%%%%%%%%%%%%%%%%%
\clearpage

%%% Main head for the e-companion

\appendix
\begin{appendices}
\section{Nomenclature}
\label{appendix: nomenclature}

We list the symbols for the proposed DBN-RL framework in Table~\ref{table: nomenclaure}.

\begin{table}[h]
\footnotesize
\caption{List of Symbols} \label{table: nomenclaure}
\begin{tabular}{@{}cl@{}}
\toprule
Variables                                             & Description                                                                                                             \\ \midrule
% {$n$}                                              & {dimension of state space}                                     \\
% {$m$}                                               & {dimension of action space}                                        \\
$s_t^k$                                               & the value of the $k$th state variable at time $t$                                             \\
$a_t^k$                                               & the value of the $k$th decision variable at time $t$                                                                    \\
{$\pmb{s}_t$} & the {$n$-dimensional state vector} \\
{$\pmb{a}_t$} & {the $m$-dimensional action vector} \\
$\pmb\beta^{s}_t$                                     & $n\times n$ matrix of coefficients representing effects of $\pmb{s}_t$ on $\pmb{s}_{t+1}$   \\
$\pmb\beta^{a}_t$                                     & $m\times n$ matrix of coefficients representing effects of $\pmb{a}_t$ on $\pmb{s}_{t+1}$\\
${\pi}_{\pmb\vartheta_t}(\pmb{s}_t)$                  & linear deterministic nonstationary policy                                                                               \\
$\pmb{\vartheta}_t$                                   & an $n\times m$ matrix of policy function coefficients                                                                            \\
$r_t(\pmb{s}_t,\pmb{a}_t)$                             & linear reward function of state and action at time $t$                                                                          \\
$m_t$                                                 &  fixed cost at time $t$                                                                                              \\
$\pmb{b}_t$                                           &  variable manufacturing cost                                                                                         \\
$\pmb{c}_t$                                           &  purification penalty cost and product revenue                                                                              \\
$\pmb{w}$                                             & DBN model parameters                                                                                                    \\
$J\left(\pi;\pmb w\right)$                            &  expected total reward over the stochastic uncertainty given a model $\pmb{w}$                                       \\
$\mathcal{J}\left(\pi_{\pmb\theta}\right)$            &  expected total reward over the stochastic uncertainty over the posterior distribution of $\pmb w$                   \\
$\kappa$                                              & reflecting different levels of stochastic uncertainty                                                                   \\
$\mathcal{D}$                                         & a set of process observations                                                                                           \\
$\eta_k$                                              & stepsize of policy gradient optimization algorithm at the $k$th iteration                                                                       \\
$\mathbb{C}$                                          & closed convex feasible region                                                                                           \\
$B$                                                   & number of posterior samples                                                                                            \\
$R$                                                   & sample size of process observations                                                                                                                                           \\
$\mu^k_{t}$                                           & mean of $k$th state variable at time $t$                                                                                \\
$v^k_{t}$                                             & SD of $k$th state variable at time $t$                                                                                  \\
$\lambda_t^k$                                         & mean of $k$th decision variable at time $t$                                                                             \\
$\sigma^k_{t}$                                        & SD of $k$th decision variable at time $t$                                                                               \\
$\pmb\mu^a_t$                                         &  mean action vector at time $t$                                                                                                  \\
$\pmb\mu^s_t$                                         &  mean state vector at time $t$                                                                                                   \\
$\pmb\tau$                                            & process trajectory                                                                      \\
$\mathbf{R}_{i,t}$                                    & product of pathway coefficients from time step $i$ to $t$.                                                          \\
$H$                                                   & planning horizon                                                                                                                 \\
$P$                                                   & product (citrate) concentration at purification step                                                                    \\
$I$                                                   & impurity concentration at purification step                                                                             \\
$X_f$                                                 & lipid-free cell mass                                                                                                    \\
$C$                                                   & citrate, the actual ``product'' generated by the cells' metabolism                                                      \\
$S$                                                   & amount of substrate                                                                                                     \\
$N$                                                   & amount of nitrogen                                                                                                      \\
$V$                                                   & working volume of the entire batch                                                                                  \\
$F_S$                                                 & feed rate, or the amount of new substrate given to the cell in one unit of time                                     \\
$\mathcal{O}_h$                                       & set of {inputs (i.e., state $\pmb{s}_h$ and $\pmb{a}_h$)} at time step $h$                                    \\
$\mbox{Sh}\left(\pmb{s}_{t+1}| {o_h}; \pmb{w}\right)$ & Shapley value of {any input $o_h \in 
\mathcal{O}_h$} with respective to the expected future state of $\pmb{s}_{t+1}$ with $t\ge h$                                                            \\ \bottomrule
\end{tabular}
\end{table}

\section{Taylor Approximation for ODE-based Kinetic Models}
\label{sec:TaylorApproximation}

This section shows how our proposed DBN model can be built using a first-order Taylor approximation of an ODE-based kinetic model.
The biomanufacturing literature generally uses kinetic models based on PDEs or ODEs to model the dynamics of these variables. Suppose that $\pmb s_t$ evolves according to the ordinary differential equation
\begin{equation}
    \frac{ d \pmb{s}_t}{d t} = \pmb{f}(\pmb{s}_t,\pmb{a}_t)
    \label{eq: differential equations}
\end{equation}
where $\pmb{f}(\cdot)=(f_1,f_2,\ldots,f_n)$ encodes the causal interdependencies between various CPPs and CQAs. One typically assumes that the functional form of $\pmb{f}$ is known, though it may also depend on additional parameters that are calibrated from data. Supposing that the bioprocess is monitored on a small time scale using sensors, let us replace (\ref{eq: differential equations}) by the first-order Taylor approximation
\begin{equation}
\frac{\Delta \pmb{s}_{t+1}}{\Delta t} =\pmb{f}(\pmb\mu_t^s,\pmb\mu_t^a)+J^s_f(\pmb\mu_t^s)(\pmb{s}_t-\pmb\mu_t^s)+J^a_f(\pmb\mu_t^a)(\pmb{a}_t-\pmb\mu_t^a),
\label{eq: rhs expansion 1}
\end{equation}
where $\Delta \pmb{s}_{t+1}=\pmb{s}_{t+1}-\pmb{s}_{t}$, and $J^s_f,J^a_f$ denote the Jacobian matrices of $\pmb{f}$ with respect to $\pmb{s}_t$ and $\pmb{a}_t$, respectively. 
The interval $\Delta t$ can change with time, but we keep it constant for simplicity.
The approximation is taken at a point $\left(\pmb\mu_t^s,\pmb\mu_t^a\right)$ to be elucidated later. We can then rewrite (\ref{eq: rhs expansion 1}) as
\begin{equation}
    \pmb{s}_{t+1} = \pmb{\mu}_{t}^s + \Delta t\cdot \pmb{f}(\pmb\mu_t^s,\pmb\mu_t^a)+(\Delta t \cdot J_f(\pmb\mu_t^s)+1)(\pmb{s}_t-\pmb\mu_t^s)+\Delta t \cdot J_f(\pmb\mu_t^a)(\pmb{a}_t-\pmb\mu_t^a)+R_{t+1},
    \label{eq: linearization}
\end{equation}
where $R_{t+1}$ is a remainder term modeling the effect from other uncontrolled factors. In this way, the original process dynamics have been linearized, with $R_{t+1}$ serving as a residual.
One can easily represent (\ref{eq: linearization}) using a network model. An edge exists from $s^k_t$ (respectively, $a^k_t$) to $s^l_{t+1}$ if the $\left(k,l\right)$th entry of $J^s_f$ (respectively, $J^a_f$) is not identically zero. %As will be seen shortly, 
The linearized dynamics can provide prior knowledge to the state transition model of the dynamic Bayesian network.
Specifically, (\ref{eq: LGBN matrix form}) has the same form as (\ref{eq: linearization}) if we let
%. Indeed, if we let
\begin{eqnarray*}
\pmb\mu_{t+1}^s =\pmb{\mu}_{t}^s + \Delta t\cdot \pmb{f}(\pmb\mu_t^s,\pmb\mu_t^a),
~~~
\pmb\beta_{t}^s = \Delta t \cdot J_f(\pmb\mu_t^s)+1,
~~~
\pmb\beta_{t}^a =\Delta t \cdot J_f(\pmb\mu_t^a),
\end{eqnarray*}
and treat $R_{t+1}$ as the residual.

\section{Details of Gibbs Sampling Procedure}\label{sec:gibbs}

This section briefly describes key computations used in the Gibbs sampling procedure.
Given the data $\mathcal{D}$, the posterior distribution of $\pmb{w}$ is proportional to $p\left(\pmb{w}\right)\prod_{n=1}^Rp\left(\pmb{\tau}^{(n)}|\pmb{w}\right)$. The Gibbs sampling technique can be used to sample from this distribution. Overall, the procedure is quite similar to the one laid out in \cite{xie2020bayesian}, so we will only give a brief description of the computations used in our setting.

For each node $X$ in the network, let $Ch\left(X\right)$ be the set of child nodes (direct successors) of $X$. The full likelihood $p\left(\mathcal{D}|\pmb{w}\right)$ becomes
\begin{equation*}
    p(\mathcal{D}|\pmb{w})=\prod^R_{i=1}p(\pmb{\tau}_i|\pmb{w})=\prod^R_{i=1}\prod_{t=1}^H\left[\prod_{k=1}^{m}\mathcal{N}\left(\lambda^k_{t},(\sigma^k_{t})^2\right)\prod_{k=1}^{n}\mathcal{N}\left(\mu^k_{t+1} + \sum_{X^j_t\in Pa(s^k_{t+1})}\beta^{jk}_{t}(X^j_t - \mu^j_t),(v^k_{t+1})^2\right)\right],
\end{equation*}
For the model parameters $\pmb\mu^a, \pmb \mu^s,\pmb\sigma^2 \pmb v^2, \pmb \beta$, we have the conjugate prior
\begin{equation*}
	p(\pmb \mu^a, \pmb \mu^s,\pmb\sigma^2 \pmb v^2, \pmb \beta) =
	\prod_{t=1}^H\left(\prod_{k=1}^{m} p(\lambda^k_t)p((\sigma_k^{t})^2)\prod_{k=1}^{n} p(\mu^k_t)p((v^k_{t})^2) \prod_{i\neq j} p(\beta^{ij}_t)\right),
\end{equation*}
where
\begin{eqnarray*}
p(\lambda^k_t) &=& \mathcal{N}\left({\lambda}^{k(0)}_t, \left({\delta}_{t,k}^{(\lambda)}\right)^2\right), \quad p(\mu^k_t) = \mathcal{N}\left({\mu}^{k(0)}_t, \left({\delta}_{t,k}^{(\mu)}\right)^2\right), \quad p(\beta_{t}^{ij}) = \mathcal{N}\left({\beta}_{t}^{ij(0)}, \left({\delta}_{t,ij}^{(\beta)}\right)^2\right)\\
p\left((\sigma_t^k)^2\right) &=& \mbox{Inv-}\Gamma\left(\dfrac{\kappa_{t,k}^{(\sigma)}}{2}, \dfrac{\rho_{t,k}^{(\sigma)}}{2}\right), \quad p((v_t^k)^2) = \mbox{Inv-}\Gamma\left(\dfrac{\kappa_{t,k}^{(v)}}{2}, \dfrac{\rho_{t,k}^{(v)}}{2}\right),
\end{eqnarray*}
where $\mbox{Inv-}\Gamma$ denotes the inverse-gamma distribution.

We now state the posterior conditional distribution for each model parameter; the detailed derivations are omitted since they proceed very similarly to those in \cite{xie2020bayesian}. Let $\pmb \mu^a_{-t,k}$, $\pmb \mu^s_{-t,k}$, $\pmb \sigma_{-t,k}^2$, $\pmb v_{-t,k}^2$ and $\pmb \beta_{-t,jk}$ denote the collection of parameters $\pmb{\mu}^a, \pmb{\mu}^s,\pmb{\sigma}, \pmb{v}, \pmb{\beta}$ excluding the $k$th or ${(j,k)}$th element at time $t$. These collections are used to obtain five conditional distributions:

\begin{itemize}
\item $p(\beta^{jk}_t|\mathcal{D}, \pmb \mu^a, \pmb \mu^s,  \pmb\sigma^2, \pmb v^2, \pmb \beta_{-t,jk})= \mathcal{N}( \tilde{\beta}_{t}^{jk}, (\tilde{\delta}_{t,jk}^{(\beta)})^2)$, where
\begin{eqnarray*}
\tilde{\beta}_{t}^{jk} &=& \dfrac{({\delta}_{t,jk}^{(\beta)})^2\sum_{i=1}^{R} \alpha_{t}^{j(i)} m_{t+1,jk}^{(i)} + (v_{t+1}^k)^2{\beta}_{t}^{ij(0)}}{({\delta}_{t,jk}^{(\beta)})^2\sum_{i=1}^{R} (\alpha_{t}^{j(i)})^2 + (v_{t+1}^k)^2}\\
\tilde{\delta}_{t,jk}^{(\beta)2} &=& \dfrac{({\delta}_{t,jk}^{(\beta)})^2(v_{t+1}^k)^2} {({\delta}_{t,jk}^{(\beta)})^2\sum_{i=1}^{R} (\alpha_{t}^{j(i)})^2 + (v_{t+1}^k)^2}
\end{eqnarray*}
with $\alpha_{t}^{j(i)} = x_t^{j(i)} - \mu_t^j$ and $m_{t+1,jk}^{(i)} = (s_{t+1}^{k(i)} - \mu_{t+1}^k) - \sum_{X_t^\ell\in Pa(s_{t+1}^k)/\{x_t^j\}}\beta_t^{\ell k}(x_t^{\ell(i)} - \mu_t^\ell)$.
\item $p\left((v_{t}^k)^2|\mathcal{D},\pmb \mu^a, \pmb \mu^s, \pmb\sigma^2, \pmb v_{-t,k}^2, \pmb \beta\right) = \mbox{Inv-}\Gamma\left(\dfrac{\tilde{\kappa}_{t,k}^{(v)}}{2}, \dfrac{\tilde{\rho}_{t,k}^{(v)}}{2}\right)$, where
\begin{eqnarray*}
\tilde{\kappa}_{t,k}^{(v)} &=& \kappa_{t,k}^{(v)} + R, \qquad \tilde{\rho}_{t,k}^{(v)} = \rho_{t,k}^{(v)} + \sum_{i=1}^{R}u_{t,k}^{(i)2},\\
u_{t,k}^{(i)} &=& (s_t^{k(i)} - \mu_t^k) - \sum_{X_t^j\in Pa(s_t^k)}\beta_t^{jk}(x_t^{j(i)} - \mu_t^j).
\end{eqnarray*}
\item $p\left((\sigma_{t}^k)^2|\mathcal{D}, \pmb \mu^a, \pmb \mu^s, \pmb \sigma_{-t,k}^2, \pmb v^2, \pmb \beta\right) = \mbox{Inv-}\Gamma\left(\dfrac{\tilde{\kappa}_{t,k}^{(\sigma)}}{2}, \dfrac{\tilde{\rho}_{t,k}^{(\sigma)}}{2}\right)$, where
\begin{equation*}
\tilde{\kappa}_{t,k}^{(\sigma)} = \kappa_{t,k}^{(\sigma)} + R, \quad \tilde{\rho}_{t,k}^{(\sigma)} = \rho_{t,k}^{(\sigma)} + \sum_{i=1}^{R}u_{t,k}^{(i)2}, \quad u_{t,k}^{(i)} = s_t^{k(i)} - \mu_t^k.
\end{equation*}
\item $p\left(\mu_t^k|\mathcal{D}, \pmb\mu^a,\pmb \mu^s_{-t,k},  \pmb \sigma^2, \pmb v^2, \pmb \beta\right)
	= \mathcal{N}(\tilde{\mu}_{t}^k, (\tilde{\delta}_{t,k}^{(\mu)})^2)$, where
\begin{eqnarray*}
\tilde{\mu}_{t}^k &=& (\tilde{\delta}_{t,k}^{(\mu)})^2 \left[ \dfrac{\mu_{t}^{k(0)}}{({\delta}_{t,k}^{(\mu)})^2} + \sum_{i=1}^{R} \dfrac{a_{t,k}^{(i)}}{(v_t^k)^2} + \sum_{i=1}^{R}\sum_{s_{t+1}^\ell\in Ch(s_t^k)} \dfrac{\beta_{t}^{k\ell}c_{t,k\ell}^{(i)}}{(v_t^\ell)^2}\right]\\
\dfrac{1}{(\tilde{\delta}_{t,k}^{(\mu)})^2} &=& \dfrac{1}{({\delta}_{t,k}^{(\mu)})^2} + \dfrac{R}{(v_t^k)^2} + \sum_{s_{t+1}^\ell\in Ch(s_t^k)}\dfrac{R(\beta_{t}^{k\ell})^2}{(v_t^\ell)^2},
\end{eqnarray*}
with
\begin{eqnarray*}
a_{t,k}^{(i)} &=& s_t^{k(i)} - \sum_{X_{t-1}^j\in Pa(s_t^k)}\beta_{t-1}^{jk}(x_{t-1}^{j(i)} - \mu_{t-1}^j)\\
c_{t,k\ell}^{(i)} &=& \beta_{t}^{k\ell}s_t^{k(i)} - (s_{t+1}^{\ell(i)} - \mu_{t+1}^\ell) + \sum_{X_t^j\in Pa(s_{t+1}^\ell)/\{s_t^k\}}\beta_{t}^{j\ell}(x_t^{j(i)} - \mu_t^j).
\end{eqnarray*}
\item $p\left(\lambda_t^k|\mathcal{D}, \pmb\mu^a_{-t,k}, \pmb \mu^s, \pmb \sigma^2, \pmb v^2, \pmb \beta\right)
	= \mathcal{N}(\tilde{\lambda}_{t}^k, (\tilde{\delta}_{t,k}^{(\lambda)})^2)$, where
\begin{eqnarray*}
\tilde{\lambda}_{t}^k &=& (\tilde{\delta}_{t,k}^{(\lambda)})^2 \left[ \dfrac{\lambda_{t}^{k(0)}}{({\delta}_{t,k}^{(\lambda)})^2} + \sum_{i=1}^{R} \dfrac{a_t^{k(i)}}{(v_t^k)^2} + \sum_{i=1}^{R}\sum_{s_{t+1}^\ell\in Ch(a_t^k)} \dfrac{\beta_{t}^{k\ell}c_{t,k\ell}^{(i)}}{(v_t^\ell)^2} \right]\\
\dfrac{1}{(\tilde{\delta}_{t,k}^{(\lambda)})^2} &=& \dfrac{1}{({\delta}_{t,k}^{(\lambda)})^2} + \dfrac{R}{(v_t^k)^2} + \sum_{s_{t+1}^\ell\in Ch(a_t^k)}\dfrac{R(\beta_{t}^{k\ell})^2}{(v_t^\ell)^2}
\end{eqnarray*}
with
\begin{equation*}
c_{t,k\ell}^{(i)} = \beta_{t}^{k\ell}a_t^{k(i)} - (s_{t+1}^{\ell(i)} - \mu_{t+1}^\ell) + \sum_{X_t^j\in Pa(s_{t+1}^\ell)/\{a_t^k\}}\beta_{t}^{j\ell}(x_t^{j(i)} - \mu_t^j).
\end{equation*}
\end{itemize}
In Gibbs sampling, we set a prior $p\left(\pmb w\right)$ and sample $\pmb w^{(0)}$ from it. Now, given $\pmb w^{(i-1)}$ we sequentially compute and generate one sample from the above conditional posterior distributions for each parameter, obtaining a new $\pmb w^{(i)}$. By repeating this process, one can arrive at a sample whose distribution is a very close approximation to the desired posterior.

% >>>

\section{Kinetic Modeling of Fed-Batch Fermentation of Yarrowia lipolytica}
\label{appendix sec: fermentation kinetics}

Here we provide additional domain information about the upstream fermentation studied in Section \ref{subsec: mdp formulation}; specifically, the complete details of the kinetic models used to construct the Bayesian network.
Cell growth and citrate production of \textit{Yarrowia lipolytica} take place inside a bioreactor, which requires carbon sources (e.g., soybean oil or waste cooking oil), nitrogen (yeast extract and ammonia sulfate), and oxygen, subject to good mixing and carefully controlled operating conditions (e.g., pH, temperature). During fermentation, the feed rate is adjusted to control the concentration of oil.

In our case study, we treat feed rate as the only control. Other CPPs relevant to the bioprocess are set automatically as follows. The dissolved oxygen level is set at $30\%$ of air saturation using cascade controls of agitation speed between 500 and 1,400 rpm, with the aeration rate fixed at 0.3 L/min. The pH is controlled at 6.0 for the first $12$ hours, then maintained at $7.0$ until the end of the process. The temperature is maintained at $30^\circ$C for the entire run. We stop the fermentation process after 140 hours (or if the volume reaches the bioreactor capacity).

%Before fermentation, the Seed Culture process will prepare the seed inoculation, which will be used to inoculate the bioreactor. The thawed seed %vial solution of \textit{Yarrowia lipolytica} strain will be transferred and grown in a shake flask containing seed culture medium, which %consisted of yeast extract, ammonia sulfate, oil, KH2PO4, and Na2HPO4. The seed culture usually takes 18-24h at $30^\circ$C and 280 rpm until cell %concentration reaches around 2-5 in OD600 (optical density measured at a wavelength of 600 nm). The Fed-Batch Fermentation starts after the seed %culture transferred to the bioreactor, which contains the initial fermentation medium consisted of yeast extract, ammonia sulfate, oil, KH2PO4, %Na2HPO4, thiamin$\cdot$HCl, 100x trace metals, and Antifoam 204; as well as the initial substrate. We use soybem oil or waste cooking oil as the %single substrate, and start the feeding when its concentration in fermentation medium decreased below 20g/L. The concentration of ooil substrate %is maintained
%by adjusting the feed rate based on off-line sample measurement.

% \begin{wrapfigure}{t}{0.6\textwidth}
% \vspace{-0.15in}
% 	\centering
% 	\includegraphics[width=0.6\textwidth]{fermentation.png}
% 	\vspace{-0.22in}
% 	\caption{Illustration of fed-batch fermentation process of Yarrowia lipolytica yeast for citrate production}\label{fig:fermentation}
% 	\vspace{-0.1in}
% \end{wrapfigure}

The kinetics of the fed-batch fermentation of \textit{Yarrowia lipolytica} can be described by a system of differential equations. The biomanufacturing literature uses deterministic ODEs, but we have augmented some of the equations with Brownian noise terms of the form $dB_t$ to reflect the inherent stochastic uncertainty of the bioprocess. Table \ref{table: model parameter description} provides a list of environmental parameters that appear in the equations, together with their values in our case study. The equations themselves can be grouped into eight parts as follows:

\begin{table}[t]\small
\caption{Mechanistic model parameter description and estimation.}\label{table: model parameter description}
\centering
\begin{tabular}{@{}llcc@{}}
\toprule
Parameters    & Descrption                                                                                                                                                       & Estimation & Unit   \\ \midrule
$\alpha_L$     & Coefficient of lipid production for cell growth                                                                                                                  & 0.1273     & -      \\
$C_{max}$       & Maximum citrate concentration that cells can tolerate                                                                                                            & 130.90     & g/L    \\
$K_{iN}$        & Nitrogen limitation constant to trigger on lipid and citrate   production                                                                                        & 0.1229     & g/L    \\
$K_{iS}$        & Inhibition constant for substrate in lipid-based growth kinetics                                                                                                   & 612.18     & g/L    \\
$K_{iX}$        & Constant for cell density effect on cell growth and lipid/citrate   formation                                                                                    & 59.974     & g/L    \\
$K_N$         & Saturation constant for intracellular nitrogen in growth kinetics                                                                                                & 0.0200     & g/L    \\
$K_O$         & \begin{tabular}[c]{@{}l@{}}Saturation constant for dissolved oxygen in kinetics of cell growth, \\ substrate uptake, lipid consumption by $\beta$-oxidation\end{tabular} & 0.3309     & \% Air \\
$K_S$         & Saturation constant for substrate utilization                                                                                                                      & 0.0430     & g/L    \\
$K_{SL}$        & Coefficient for lipid consumption/decomposition                                                                                                                  & 0.0217     & -      \\
$m_s$         & Maintenance coefficient for substrate                                                                                                                              & 0.0225     & g/g/h  \\
$r_L$         & Constant ratio of lipid carbon flow to total carbon flow (lipid + citrate)                                                                                     & 0.4792     & -      \\
$V_{evap}$      & Evaporation rate (or loss of volume) in the fermentation                                                                                    & 0.0026     & L/h    \\
$Y_{cs}$        & Yield coefficient of citrate based on substrate consumed                                                                                                           & 0.6826     & g/g    \\
$Y_{ls}$        & Yield coefficient of lipid based on substrate consumed                                                                                                             & 0.3574     & g/g    \\
$Y_{xn}$        & Yield coefficient of cell mass based on nitrogen consumed                                                                                                        & 10.0       & g/g    \\
$Y_{xs}$        & Yield coefficient of cell mass based on substrate consumed                                                                                                         & 0.2386     & g/g    \\
$\beta_{LCmax}$ & Coefficient of maximum carbon flow for citrate and lipid                                                                                                         & 0.1426     & h-1    \\
$\mu_{max}$      & Maximum specific growth rate on substrate                                                                                                                          & 0.3845     & h-1    \\
$S_F$         & Oil concentration in oil feed                                                                                                                                    & 917.00     & g/L    \\ \bottomrule
\end{tabular}
 \vspace{-0.15in}
\end{table}

\begin{enumerate}
    \item \textbf{Cell mass}: The total cell mass $X$  consists of lipid-free ($X_f$) and lipid ($L$) mass, and is measured by dry cell weight (DCW) in fermentation experiments.
    \begin{align*}
    X &= X_f+L 	%Eq. (1a)
    % \frac{dX}{dt} &=\frac{dX_f}{dt}+\frac{dL}{dt} 	 %Eq. (1b)
    % X^{t_2} &= X^{t_1} + (X_f^{t_2} - X_f^{t_1}) + (L^{t_2} - L^{t_1}) + e_X^{t_2}
    \end{align*}

    \item \textbf{Dilution rate}: The dilution $D$ of the working liquid volume in the bioreactor is caused by feed of base, such as KOH solution ($F_B$) and substrate ($F_S$):
    \begin{align*}
    D &=\frac{F_B+F_S}{V} \nonumber	%Eq. (2)
    \end{align*}

    \item \textbf{Lipid-free cell growth}: Cell growth consumes nutrients, including the substrate (carbon source) $S$, nitrogen $N$, and dissolved oxygen $O$, and is described by coupled Monod equations, with considerations of inhibitions from high oil concentrations and cell densities:
    \begin{align*}
    \mbox{d}X_f &=\mu X_f \mbox{d}t-\left(D-\frac{V_{evap}}{V}\right) X_f \mbox{d}t + \sigma(X_f)\mbox{d}B_t \\	%Eq. (3a)
    % X_f^{t_2} &= X_f^{t_1} + (t_2-t_1) \cdot [\mu^{t_1} X_f^{t_1}-(D^{t_1}-\frac{V^{t_1}_{evap}}{V^{t_1}}) X_f^{t_1} ] + e_{X_f}^{t_2}\\
    \mu &=\mu_{max} \left(\frac{S}{K_S+S} \cdot \frac{1}{1+S/K_{iS}}\right) \frac{N}{K_N+N} \cdot \frac{O}{K_O+O} \cdot \frac{1}{1+X_f/K_{ix}} 	%Eq. (3b)
    \end{align*}

    \item \textbf{Citrate accumulation}: Citrate ($C$) is an overflow of all the carbon introduced to the lipid synthesis pathway ($\beta_{LC}$), which is more active under nitrogen-limited, substrate-rich, and aerobic conditions. Only a proportion $r_L$ of the total carbon flow (citrate plus lipid) in the lipid synthesis pathway goes to production due to the overflow loss in citrate. A tolerance limit $C_{max}$ of citrate, and the effect of cell density on product formation ($1/(1+X_f/K_{iX})$), are also considered in the model.
    \begin{align*}
    {\mbox{d}C}&= \beta_C \cdot X_f \mbox{d}t -\left(D-\frac{V_{evap}}{V}\right)C \mbox{d}t +\sigma(C)\mbox{d}B_t  \\%	Eq. (4a)
    % C^{t_2} &= C^{t_1} + (t_2-t_1) \cdot [q_C^{t_1} \cdot X_f^{t_1}-(D^{t_1}-\frac{V^{t_1}_{evap}}{V^{t_1}})C^{t_1}] + e_{C}^{t_2}\\
    \beta_C &= 2(1-r_L)\beta_{LC} \\	%Eq. (4b)
    \beta_{LC} &= \frac{1}{1+N/K_{iN}} \cdot \left(\frac{S}{K_S+S} \cdot \frac{1}{1+S/K_{iS}}\right) \frac{O}{K_O+O} \cdot \frac{1}{1+X_f/K_{iX}} \left(1-\frac{C}{C_{max}} \right) \beta_{LCmax}  % 	Eq. (4c)
    \end{align*}

    \item \textbf{Lipid accumulation}: Lipid is accumulated under nitrogen-limited, substrate-rich, and aerobic conditions. Lipid production is described using partial growth-association kinetics; a small portion of lipid can be degraded when its concentration is high in the presence of oxygen.
    \begin{align*}
    \mbox{d}{L}&=q_L \cdot X_f\mbox{d}t -\left(D-\frac{V_{evap}}{V}\right)L \mbox{d}t + \sigma(L) \mbox{d}B_t\\
          &=\left(\alpha_L \cdot \mu + \beta_L \right) X_f \mbox{d}t-\left(D-\frac{V_{evap}}{V}\right)L \mbox{d}t + \sigma(L) \mbox{d}B_t\\	%Eq. (5a)
    % L^{t_2} &= L^{t_1} + (t_2-t_1) \cdot [(\alpha_L^{t_1} \cdot \mu^{t_1} + \beta_L^{t_1} ) X_f^{t_1}-(D^{t_1}-\frac{V^{t_1}_{evap}}{V^{t_1}})L^{t_1}] + e_L^{t_2} \\
    \beta_L &=r_L \cdot \beta_{LC}-K_{SL}  \frac{L}{L+X_f} \cdot \frac{O}{K_O+O} 	%Eq. (5b)
    \end{align*}

    \item \textbf{Substrate consumption}: Oil ($S$) is fed during the fed-batch fermentation and used for cell growth, energy maintenance, citrate formation, and lipid production.
    \begin{align*}
    -{dS} &= q_S \cdot X_f\mbox{d}t - \frac{F_S}{V} S_F \mbox{d}t+ \left(D-\frac{V_{evap}}{V}\right)S\mbox{d}t +\sigma(S)\mbox{d}B_t 	\\%Eq. (6a)
    % S^{t_2} &= S^{t_1} - (t_2-t_1) \cdot [ q_S^{t_1} \cdot X_f^{t_1} - \frac{F_S^{t_1}}{V^{t_1}} S_F^{t_1}+ (D^{t_1}-\frac{V^{t_1}_{evap}}{V^{t_1}})S^{t_1}] + e_S^{t_2} \\
    q_S &= \frac{1}{Y_{X/S}} \mu +\frac{O}{K_O+O} \cdot \frac{S}{K_S+S} m_S+ \frac{1}{Y_{C/S}}  \beta_C + \frac{1}{Y_{L/S}}  \beta_L 	%Eq. (6b)
    \end{align*}

    \item \textbf{Nitrogen consumption}: Extracellular nitrogen ($N$) is used for cell growth.
    \begin{align*}
    -{dN} &= \frac{1}{Y_{X/N}}  \mu X_f \mbox{d}t+(D-\frac{V_{evap}}{V}) N\mbox{d}t + \sigma(N) \mbox{d}B_t  %	Eq. (7)
    % N^{t_2} &= N^{t_1} - (t_2-t_1) \cdot [\frac{1}{Y^{t_1}_{X/N}}  \mu^{t_1} X_f^{t_1} + (D^{t_1}-\frac{V^{t_1}_{evap}}{V^{t_1}}) N^{t_1}] + e_N^{t_2}
    \end{align*}

    \item \textbf{Volume change}: The rate $V$ of working volume change of the fermentation is calculated based on the rates of base feed ($F_B$), substrate feed ($F_S$), and evaporation ($V_{evap}$).
    \begin{align*}
    {dV} &= (F_B + F_S - V_{evap})\mbox{d}t + \sigma(V)\mbox{d}B_t \\ %	Eq. (8a)
    % V^{t_2} &= V^{t_1} + (t_2-t_1) \cdot [F_B^{t_1} + F_S^{t_1} - V^{t_1}_{evap}] + e_V^{t_2} \\
    F_B &= \frac{V}{1000} \left(\frac{7.14}{Y_{X/N}}  \mu X_f + 1.59 \beta_C X_f \right) 	%Eq. (8b)
    \end{align*}
\end{enumerate}

Figure \ref{fig:ODE-fitness} in the main text shows the fit of the ODE trajectory to the 8 available batches of experimental data. The fit is not perfect due to the limited sample size and the noticeably large variation between experimental trajectories (a consequence of the fact that scientists conducted these experiments with very different oil feeding profiles).

% -----------------------------

\section{{Purification Process Dynamics in Integrated Process Control}}
\label{subsec:integratedUpstreamDownstream}

This section provides details about the downstream operation added in Section \ref{sec:newintegrated}.
Similarly to Appendix \ref{appendix sec: fermentation kinetics}, we develop a mechanistic model of the purification process during the production of citrate acid. The downstream process described here is a simplified version of what is given in \cite{varga2001prediction}. % to one centrifugation and two successive precipitation steps. The simplified process flow 
It contains multiple stages, including: (C1) disc stack centrifugation for cell debris removal; (P1) a first ammonium sulphate precipitation stage to selectively precipitate undesired impurities; and (P2) a second ammonium sulphate precipitation to selectively precipitate  the desired product. 
% and a disc stack centrifugation stage to recover the solid citrate acid precipitate (C2).
In the downstream setup, sensors are only placed at the entrance and exit of each stage.

As the particle sizes of whole cell or cell debris are relatively large, we expect Stage (C1) to yield good separation of the material. Thus, given the output $(X_{f},C,S,N)$ of the upstream fermentation process, we obtain the following aggregated process state after centrifugation:
\begin{equation}
    \mbox{(C1)} \quad  (P, I)=(C, \xi  X_{f}+S+N)
    \label{eq.centrifiguation}
\end{equation}
where $P$ denotes the total amount of product, $I$ denotes the total amount of impurity, and $\xi=0.01$ denotes the separation efficiency of cell debris in Stage~(C1).

\begin{figure}[t]
	\centering
	\includegraphics[width=0.5\textwidth]{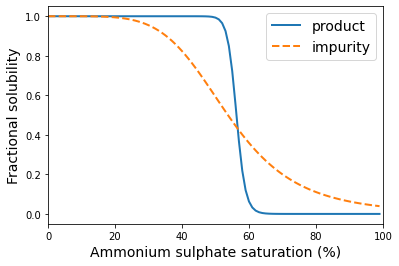}
	\caption{Fractional solubility of product and impurity.
	}\label{fig: fractional solubility}
	    \vspace{-0.3in}
\end{figure}

Next, we discuss the ammonium sulphate precipitation steps (P1) and (P2). Purification through precipitation involves a trade-off between yield and purity, illustrated in Figure~\ref{fig: fractional solubility}. The operator %scientist 
needs to decide the ammonium sulphate saturation to remove as much impurity and as little product as possible. In this example, the operator %scientist can 
selects low ammonium sulphate saturation (40\%-50\%) to precipitate the undesired impurity in Step~(P1) and selects high ammonium sulphate saturation (60\%-80\%) to selectively precipitate the product in Step~(P2). 
%In the simplified process, 
Thus, in (P1), the precipitation process keeps more product but removes less impurity if lower ammonium sulphate saturation is used; and in (P2), the process precipitates or keeps more product but removes more impurity if lower ammonium sulphate saturation is used.

The fractional ammonium sulphate solubility of natural product and impurity from yeast have been described by \cite{niktari1990monitoring} using the equation      \vspace{-0.1in}
\begin{equation}
    F(\zeta)=\frac{1}{1+(\zeta/a)^b}
    \label{eq.solubilityModel}
     \vspace{-0.1in}
\end{equation}
where $F$ is fractional solubility (the fraction of enzyme
or protein that is soluble), $\zeta$ is the ammonium sulphate saturation
and $a$, $b$ are constants. In the simulation model, we set the fractional solubility parameters as $a =56.27$, $b = 42.00$ for product and $a =53.72$, $b=5.23$ for impurity. Thus the nonlinear relationships of process states before and after stages (P1) and (P2) are given by
\begin{align}
    \mbox{(P1)} \quad P_{t+1} &= F_P(\zeta_t)\cdot P_t, \quad I_{t+1} = F_I(\zeta_t)\cdot I_t \label{eq: purification model 1}\\
    \mbox{(P2)} \quad P_{t+1} &= (1-F_P(\zeta_t))\cdot P_t, \quad I_{t+1} = (1-F_I(\zeta_t))\cdot I_t \label{eq: purification model 2}
\end{align}
where $F_P(\zeta)$ and $F_I(\zeta)$ denote the fractional solubility model in (\ref{eq.solubilityModel}) for product and impurity respectively. Then by taking logarithms and including normally distributed residual terms, the nonlinear state transition \eqref{eq: purification model 1}-\eqref{eq: purification model 2} becomes
\begin{align*}
    \mbox{(P1)} \quad \log P_{t+1} &= \log F_P(\zeta_t)+\log P_t + e^P_t, \quad \log I_{t+1} = \log F_I(\zeta_t)+ \log I_t  + e^I_t\\
    \mbox{(P2)} \quad \log P_{t+1} &= \log\left(1-F_P(\zeta_t)\right) + \log P_t  + e^P_t,\quad \log I_{t+1} = \log\left(1-F_I(\zeta_t)\right) +\log I_t  + e^I_t. 
\end{align*}
% where $\log F(S)=-\log(1+(S/a)^{b})=\sum ^\infty_{i=0}\frac{(-1)^{i+1} (S/a)^{bi}}{i+1}$ and $\log \left(1-F(S)\right)=b\log S-b\log a-\log(1+(S/a)^{b})=b\log S-b\log a+\sum ^\infty_{i=0}\frac{(-1)^{i+1} (S/a)^{bi}}{i+1}$. These equations can be viewed as combinations of logarithm and polynomials of $S$.
This gives the state transition model for Phases~(P1) and (P2) from state $(\log P_t, \log I_t)$ to next state $(\log P_{t+1}, \log I_{t+1})$ after taking the action $\log \zeta_t$ in the precipitation processes.

\section{Proofs}\label{sec:allproofs}

In this section, we give the proofs and the derivations of all results stated in the main text.
%Below, we give full proofs for all results that were stated in the main text.

\subsection{Proof for Proposition~\ref{prop: state representation}}
\label{appendix:proposition2}

By plugging in the linear policy function (\ref{eq: linear policy func}) into the state transition in (\ref{eq: LGBN matrix form}), we have
\begin{eqnarray}
    \pmb{s}_{t+1} &=& \pmb{\mu}_{t+1}^s + \left({\pmb{\beta}_{t}^s}\right)^\top(\pmb{s}_t-\pmb\mu_t^s) + \left(\pmb\beta_{t}^a\right)^\top(\pmb{a}_{t}-\pmb\mu_t^a) +V_{t+1}\mathbf{z}
    \nonumber\\
    &=& \pmb{\mu}_{t+1}^s + \left(\pmb{\beta}_{t}^s\right)^\top(\pmb{s}_t-\pmb\mu_t^s) + \left(\pmb\beta_{t}^a\right)^\top\left[\pmb\mu^a_t + \pmb\vartheta_t^\top(\pmb{s}_t - \pmb\mu_{t}^s)-\pmb\mu_t^a\right] +\pmb{e}_{t+1} %\mathcal{N}(0,V_{t+1}^2)
    \nonumber\\
    &=&\pmb{\mu}_{t+1}^s + \left(\left(\pmb{\beta}_{t}^s\right)^\top + \left(\pmb\beta_{t}^a\right)^\top \pmb\vartheta_t^\top\right)(\pmb{s}_t - \pmb\mu^{s}_t) +\pmb{e}_{t+1} \label{eq.mid2}
    \\
    &=&\pmb{\mu}_{t+1}^s + (\left(\pmb{\beta}_{t}^s\right)^\top + \left(\pmb\beta_{t}^a\right)^\top \pmb\vartheta_t^\top)\left(\left(\pmb{\beta}_{t-1}^s\right)^\top + \left(\pmb\beta_{t-1}^{a}\right)^\top \pmb\vartheta_{t-1}^\top\right)(\pmb{s}_{t-1} - \pmb\mu^{s}_{t-1}) +\left[(\left(\pmb{\beta}_{t}^s\right)^\top + \left(\pmb\beta_{t}^a\right)^\top \pmb\vartheta_t^\top)\pmb{e}_{t}+\pmb{e}_{t+1}\right]\nonumber\\
        &=&\pmb{\mu}_{t+1}^s + \mathbf{R}_{t-1,t}(\pmb{s}_{t-1} - \pmb\mu^{s}_{t-1}) +\left[\left(\left(\pmb{\beta}_{t}^s\right)^\top + \left(\pmb\beta_{t}^a\right)^\top \pmb\vartheta_t^\top\right)\pmb{e}_{t}+\pmb{e}_{t+1}\right]\nonumber\\
    &&\ldots\nonumber\\
&=&\pmb{\mu}_{t+1}^s +\prod_{i=1}^t\left(\left(\pmb\beta_{i}^s\right)^\top + \left(\pmb\beta_{i}^a\right)^\top\pmb\vartheta_i^\top\right)(\pmb{s}_{1} - \pmb\mu^{s}_{1}) +\left[\sum_{i=1}^{t}\left(\prod_{j=i}^t\left(\left(\pmb\beta_{j}^s\right)^\top + \left(\pmb\beta_{j}^a\right)^\top\pmb\vartheta_j^\top\right)\pmb{e}_{i}\right)+\pmb{e}_{t+1}\right]\label{eq: state_transition until h}\\
&=&\pmb{\mu}_{t+1}^s +\mathbf{R}_{1,t}(\pmb{s}_{1} - \pmb\mu^{s}_{1}) +\left[\sum_{i=1}^{t}\mathbf{R}_{i,t}\pmb{e}_{i}+\pmb{e}_{t+1}\right]
\label{eq: state_transition}
\vspace{-0.1in}
\end{eqnarray}
where $\pmb{e}_t$ is a $n$-dimensional normal random column vector with mean zero and diagonal covariance matrix $V_{t}\triangleq\mbox{diag}((v_{t}^s)^2)$. Similarly, one can expand $\pmb{s}_{t+1}$ until time $h$ to obtain an analog of (\ref{eq: state_transition until h}), which yields the desired result.

\subsection{Proof for Proposition~\ref{prop2}}
\label{subsec:proof_proposition3}

The conditional mean follows directly from (\ref{eq: state_transition}) above. Because the vectors $\pmb{e}_i,\pmb{e}_j$ are mutually independent, we obtain
\begin{eqnarray*}
\mbox{Var}[\pmb{s}_{t+1}|\pmb{s}_1,\pmb\pi_{\pmb\theta}] &=& \mbox{Var}\left[\sum_{i=1}^{t}\mathbf{R}_{i,t}\pmb{e}_{i}+\pmb{e}_{t+1}\right]= \sum_{i=1}^{t}\mathbf{R}_{i,t}\mbox{Var}\left[\pmb{e}_{i}\right]\mathbf{R}_{i,t}^\top+\mbox{Var}\left[\pmb{e}_{t+1}\right]=\sum_{i=1}^{t}\mathbf{R}_{i,t}V_{i}^2\mathbf{R}_{i,t}^\top+V_{t+1}^2.
\end{eqnarray*}
The objective function can be obtained from (\ref{eq:reward structure}) and (\ref{eq: linear policy func}) as,
\begin{eqnarray*}
J(\pmb\theta;\pmb{w}) &=& \sum_{t=1}^H \E[r_t(\pmb{s}_t,\pmb{a}_t)|\pmb{\pi}_{\pmb\theta},\pmb{s}_1,\pmb{w}] =\sum_{t=1}^H m_t+\pmb{b}_t^\top \pmb{a}_t +\pmb{c}_t^\top\E[\pmb{s}_t|\pmb{\pi}_{\pmb\theta},\pmb{s}_1,\pmb{w}]\\
&=& \sum^{H}_{t=1} m_t+\pmb b_t^\top\pmb\mu^a_t+(\pmb{b}_t^\top\pmb\vartheta_t^\top+\pmb{c}_t^\top)\E[\pmb{s}_t|\pmb{s}_1,\pmb{\pi}_{\pmb\theta}]-\pmb{b}_t^\top\pmb\vartheta^\top \pmb\mu_t^s.
\end{eqnarray*}
Plugging in the conditional mean computed previously, we obtain
\begin{eqnarray*}
J(\pmb\theta;\pmb{w}) &=& \sum^{H}_{t=1} m_t+\pmb b_t^\top\pmb\mu^a_t+(\pmb{b}_t^\top\pmb\vartheta_t^\top+\pmb{c}_t^\top)(\pmb\mu_t^s+\mathbf{R}_{1,t}(\pmb{s}_1 - \pmb\mu_1^s))-\pmb{b}_t^\top\pmb\vartheta_t^\top\pmb\mu^s_t\\
    &=&\sum^{H}_{t=1} m_t+\pmb b_t^\top\pmb\mu^a_t+\pmb{c}_t^\top\pmb\mu_t^s+(\pmb{b}_t^\top\pmb\vartheta_t^\top+\pmb{c}_t^\top)\mathbf{R}_{1,t}(\pmb{s}_1 - \pmb\mu_1^s).
\end{eqnarray*}

% --------------------------------------

\subsection{{Proof of Theorem~\ref{thm: fctor imptance to cond. expectation} }}
\label{appendix sec: proofs for SV Theorem}

At any time $h$, let 
% $\mathbf{A}_h=\{a_h^k\}_{k=1}^m$, $\mathbf{S}_h=\{s_h^k\}_{k=1}^n$ and
{$\mathcal{O}_h= \{s_h^k\}_{k=1}^n\cup\{a_h^k\}_{k=1}^m$ be the set of state and action inputs.
Given any state input $s^k_h\in\mathcal{O}_h$, we have the expected prediction of the future state $\pmb{s}_{t+1}$ with $t\geq h$,}
\begin{align}
    f_{t+1}(s_h^k)&=\E[\pmb{s}_{t+1}|s_h^k]\nonumber\\
    &=\E\left[\E[\pmb{s}_{t+1}|\pmb{s}_{h+1}]|s_h^k\right] \label{eq: shapley value 1}\\
    &=\E\left[\pmb{\mu}_{t+1}^s +\mathbf{R}_{h+1,t}(\pmb{s}_{h+1} - \pmb\mu^{s}_{h+1})|s_h^k \right] \label{eq: shapley value 2}\\
    &=\pmb{\mu}_{t+1}^s +\mathbf{R}_{h+1,t}\E\left[\pmb{s}_{h+1}-\pmb\mu^{s}_{h+1}|s_h^k \right] \nonumber\\
    &=\pmb{\mu}_{t+1}^s +\mathbf{R}_{h+1,t}\E\left[
    \left. \left(\pmb{\beta}_{h}^s\right)^\top(\pmb{s}_t-\pmb\mu_t^s) + \left(\pmb\beta_{h}^a\right)^\top(\pmb{a}_t-\pmb\mu_t^a)
    \right|s_h^k \right]\label{eq: shapley value 3}\\
    &=\pmb{\mu}_{t+1}^s +\mathbf{R}_{h+1,t}\left(\pmb{\beta}_{h}^s\right)^\top(0,\ldots,s^k_h-\mu^{k}_h,\ldots,0)^\top
    % +\mathbf{R}_{h+1,t}\left(\pmb{\beta}_{h}^a\right)^\top\pmb\vartheta_h^\top(0,\ldots,s^k_h-\mu^{k}_h,\ldots,0)^\top
    \label{eq: shapley value 4}
    % &=\pmb{\mu}_{t+1}^s +\mathbf{R}_{h+1,t}\left(\left(\pmb{\beta}_{h}^s\right)^\top + \left(\pmb{\beta}_{h}^a\right)^\top\pmb\vartheta_h^\top\right)(0,\ldots,s^k_h-\mu^{k}_h,\ldots,0)^\top \nonumber\\
    % &=\pmb{\mu}_{t+1}^s +\mathbf{R}_{h,t}(0,\ldots,s^k_h-\mu^{k}_h,\ldots,0)^\top \nonumber
\end{align}
where \eqref{eq: shapley value 1} follows by the tower property of conditional expectation, \eqref{eq: shapley value 2} follows by Proposition~\ref{prop: state representation}, and \eqref{eq: shapley value 3} follows due to \eqref{eq: LGBN matrix form}. Equation~\eqref{eq: shapley value 4} holds because $\E[{s}_h^j]=\mu^{j}_h$ and $\E[{a}_h^j]=\lambda^{j}_h$ for any $j\neq k$. Thus, we have $\E[\pmb{a}_t-\pmb\mu_t^a|{s}_h^k]=\pmb 0$
% $\E[\pmb{a}_t-\pmb\mu_t^a|{s}_h^k]=\pmb\vartheta_t^\top (0,\ldots,s^k_h-\mu^{k}_h,\ldots,0)^\top$ 
and $\E[\pmb{s}_t-\pmb\mu_t^s|{s}_h^k]=(0,\ldots,s^k_h-\mu^{k}_h,\ldots,0)^\top$. Similarly,
for an action $a^k_h\in\mathcal{O}_h$, we have 
\begin{equation}
    f_{t+1}(a_h^k)=\E[\pmb{s}_{t+1}|a_h^k]
    =\pmb{\mu}_{t+1}^s +\mathbf{R}_{h+1,t}\left(\pmb{\beta}_{h}^a\right)^\top(0,\ldots,a^k_h-\lambda^{k}_h,\ldots,0)^\top.
    \label{eq: shapley value 5}
\end{equation}
For any subset $\mathcal{U}\subset\mathcal{O}_h$, based on \eqref{eq: shapley value 4} and \eqref{eq: shapley value 5}, we know that the conditional expectation $f_{t+1}(\mathcal{U})$ is linear in the observed inputs $s_h^k$ and $a_h^k$. Thus, we have
\begin{align}
    f_{t+1}(\mathcal{U}\cup\{s_h^k\})-f_{t+1}(\mathcal{U}\})&=\E[\pmb{s}_{t+1}|\mathcal{U}\cup\{s_h^k\}]-\E[\pmb{s}_{t+1}|\mathcal{U}] 
    %\nonumber\\
    %&= f_{t+1}(\{s_h^k\})-f_{t+1}(\emptyset) \nonumber\\
    = \mathbf{R}_{h+1,t}(\pmb\beta^s_h)^\top(0,\ldots,s^k_h-\mu^{k}_h,\ldots,0)^\top, \nonumber
\end{align}
\begin{align}
    f_{t+1}(\mathcal{U}\cup\{a_h^k\})-f_{t+1}(\mathcal{U}\cup\{a_h^k\})&=\E[\pmb{s}_{t+1}|\mathcal{U}\cup\{a_h^k\}]-\E[\pmb{s}_{t+1}|\mathcal{U}] 
    %\nonumber\\
    %&= f_{t+1}(\{a_h^k\})-f_{t+1}(\emptyset) \nonumber\\
    = \mathbf{R}_{h+1,t}\left(\pmb{\beta}_{h}^a\right)^\top(0,\ldots,a^k_h-\lambda^{k}_h,\ldots,0)^\top.
    \nonumber
\end{align}
%where $f_{t+1}(\emptyset) = \E[\pmb{s}_{t+1}]=\pmb\mu_{t+1}^s$
% For any ${o}^k_h\in\mathcal{O}_h$, we can combine \eqref{eq: shapley value 4} and \eqref{eq: shapley value 5} as follows
% \begin{equation}
%     f_{t+1}(o_h)=\E[\pmb{s}_{t+1}|o_h]
%     =\pmb{\mu}_{t+1}^s +\mathbf{R}_{h+1,t}\left(\pmb{\beta}_{h}^o\right)_{k.}^\top({o}^k_h-\mu_h^{o,k}) \mbox{ where } (\mu^{o,k}_h,\pmb{\beta}_{h}^o)=\begin{cases} (\mu^{a,k}_h,\pmb{\beta}_{h}^a) & \text{if $o_h \in \mathbf{A}_h$}  \\
%     (\mu^{s,k}_h,\pmb{\beta}_{h}^s) & \text{if $o_h \in \mathbf{S}_h$}
%     \end{cases} \label{eq: shapley value 6}
% \end{equation}
% Then it follows that 
% \begin{align}
%   f_{t+1}(\mathcal{U}\cup\left\{s^k_h\right\}) - f_{t+1}(\mathcal{U})&= \left[ \left(\pmb{\mu}_{t+1}^s +\mathbf{R}_{h+1,t}\sum_{o_h^\ell\subset \mathcal{U}\cup\{o^k_h\}}\left(\pmb{\beta}_{h}^o\right)_{.\ell}^\top({o}^\ell_h-\mu_h^{o,\ell})\right) 
% 			- \left(\pmb{\mu}_{t+1}^s +\mathbf{R}_{h+1,t}\sum_{o^\ell_h\in \mathcal{U}}\left(\pmb{\beta}_{h}^o\right)_{.\ell}^\top({o}^\ell_h-\mu_h^{o,\ell})\right)\right]
% \end{align}
Then, the Shapley value of {any input $o_h\in\mathcal{O}_h$} becomes
		\begin{align}
    	 \mbox{Sh}\left(\pmb{s}_{t+1}| o_h\right)
	&=\sum_{\mathcal{U}\subset \mathcal{O}_h/\{o_h\}}\dfrac{(|\mathcal{O}_h| -|\mathcal{U}|-1)! |\mathcal{U}|!} {|\mathcal{O}_h|!}\left[ f_{t+1}(\mathcal{U}\cup\left\{{o}_h\right\}) - f_{t+1}(\mathcal{U}) \right] \nonumber\\
	%&=\sum_{\mathcal{U}\subset \mathcal{O}_h/\{o^k_h\}}\dfrac{(|\mathcal{O}_h| -|\mathcal{U}|-1)! |\mathcal{U}|!} {|\mathcal{O}_h|!}\left( f_{t+1}(\left\{o^k_h\right\}) - f_{t+1}(\emptyset)\right) \nonumber\\
% 			&= \sum_{\mathcal{U}\subset \mathcal{O}_h/\{o^k_h\}}\dfrac{(|\mathcal{O}_h| -|\mathcal{U}|-1)! |\mathcal{U}|!} {|\mathcal{O}_h|!}\left[ \left(\pmb{\mu}_{t+1}^s +\mathbf{R}_{h+1,t}\sum_{o_h^\ell\subset \mathcal{U}\cup\{o^k_h\}}\left(\pmb{\beta}_{h}^o\right)_{.\ell}^\top({o}^\ell_h-\mu_h^{o,\ell})\right) \right. \\
% 			&- \left.\left(\pmb{\mu}_{t+1}^s +\mathbf{R}_{h+1,t}\sum_{o^\ell_h\in \mathcal{U}}\left(\pmb{\beta}_{h}^o\right)_{.\ell}^\top({o}^\ell_h-\mu_h^{o,\ell})\right)\right] \nonumber\\
		%	&= \left( f_{t+1}(\left\{o^k_h\right\}) - f_{t+1}(\emptyset)\right)\sum_{s=0}^{a}\binom{s}{a}\frac{(a-s)!s!}{(a+1)!} \nonumber\\
		%	&=  f_{t+1}(\left\{o^k_h\right\}) - f_{t+1}(\emptyset) \nonumber\\
			&= \begin{cases} \mathbf{R}_{h+1,t}\left(\pmb{\beta}_{h}^a\right)^\top(0,\ldots,0,a^k_h-\lambda^{k}_h,0,\ldots,0)^\top & \text{if $o_h =a_h^k \in {\mathcal{O}_h}$};  \\
    \mathbf{R}_{h+1,t}\left(\pmb\beta^s_h\right)^\top(0,\ldots,0,s^k_h-\mu^{k}_h,0,\ldots,0)^\top & \text{if $o_h =s_h^k\in {\mathcal{O}_h}$},
    \end{cases}\label{eq.shapleyftr}
% 			\mathbf{R}_{h+1,t}\left(\pmb{\beta}_{h}^\ell\right)_{k.}^\top({o}^k_h-\mu_h^{o,k})
		\end{align}
% It is obvious that for any observation set $\mathcal{O}$, the factor importance for $X_k$ is the same.
completing the proof.

% -------------------------------------

\subsection{Proof of Theorem~\ref{thm: nested backpropagation} }
\label{appendix sec: proofs for Nested Backpropagaation Theorem}

From Proposition \ref{prop: state representation}, we have $\bar{\pmb{s}}_{t} = \pmb{\mu}_{t}^s + \mathbf{R}_{h,t-1}(\bar{\pmb{s}}_{h} - \pmb\mu^{s}_{h})$. We then take the gradient of $\bar{\pmb{s}}_{t}$ over the policy parameters $\pmb\vartheta_{h}$, obtaining
\begin{eqnarray}
    \frac{\partial \bar{\pmb{s}}_{t}}{\partial \pmb\vartheta_{h}} &=& \frac{\partial}{\partial \pmb\vartheta_{h}}\mathbf{R}_{h,t-1}(\bar{\pmb{s}}_{h} - \pmb\mu^{s}_{h})\nonumber\\
    &=& (\bar{\pmb{s}}_{h} - \pmb\mu^{s}_{h})^\top\left[\prod_{i=h+1}^{t-1}(\pmb\beta_{i}^s+\pmb\vartheta_i\pmb\beta_{i}^a)^\top\right]\pmb\beta_{h}^{a\top}\nonumber\\
    &=& (\bar{\pmb{s}}_{h} - \pmb\mu^{s}_{h})^\top\mathbf{R}_{h+1,t-1}\left(\pmb\beta_{h}^{a}\right)^\top.\label{eq: expState}
\end{eqnarray}
Combining this derivation with (\ref{eq.barR}), we can obtain $\frac{\partial \bar{r}_t}{\partial \pmb\vartheta_t}$ in the following cases:
\begin{itemize}
    \item When $h=t$,
    \begin{eqnarray}
    \frac{\partial \bar{r}_t}{\partial \pmb\vartheta_t}&=&\frac{\partial}{\partial \pmb\vartheta_t}\left(m_t+\pmb c_t^\top \bar{\pmb{s}}_t +\pmb b_t^\top(\pmb\mu_t^a+\pmb\vartheta_t^\top(\bar{\pmb{s}}_t -\pmb{\mu}_t^s))\right)
    =\left(\bar{\pmb{s}}_t -\pmb\mu_t^s\right)\pmb b_t^\top\label{eq: NB proof 1}
\end{eqnarray}
\item When $h=t-1$,
\begin{eqnarray}
        \frac{\partial \bar{r}_t}{\partial\pmb \vartheta_{t-1}}&=&\frac{\partial}{\partial \pmb\vartheta_{t-1}}\left(m_t+\pmb c_t^\top \bar{\pmb{s}}_t +\pmb b_t^\top(\pmb\mu_t^a+\pmb\vartheta_t^\top(\bar{\pmb{s}}_t -\pmb{\mu}_t^s))\right)\nonumber\\
    &=&\frac{\partial}{\partial \pmb\vartheta_{t-1}}(\pmb c_t^\top \bar{\pmb{s}}_t +\pmb b_t^\top\pmb\vartheta_t^\top\bar{\pmb{s}}_t )\nonumber\\
    &=&(\pmb{c}_t+\pmb\vartheta_t\pmb{b}_t)\frac{\partial\bar{\pmb s}_t}{\partial \pmb\vartheta_{t-1}}\nonumber\\
    &=&(\pmb{c}_t+\pmb\vartheta_t\pmb{b}_t)(\bar{\pmb{s}}_{t-1}-\pmb\mu_{t-1}^s)^\top\left(\pmb\beta^{a}_{t-1}\right)^{\top},
    \label{eq: NB proof 2}
\end{eqnarray}
where (\ref{eq: NB proof 2}) follows by applying (\ref{eq.mid2}) to $\pmb{s}_t$.

% and  $\mathbf{R}_{t,t-1}=\mathbb{I}_{n\times n}$.
\item When $h<t-1$,
\begin{eqnarray}
        \frac{\partial \bar{r}_t}{\partial \pmb\vartheta_k}&=&\frac{\partial}{\partial \pmb\vartheta_t}\left(m_t+\pmb c_t^\top \bar{\pmb{s}}_t +\pmb b_t^\top(\pmb\mu_t^a+\pmb\vartheta_t^\top(\bar{\pmb{s}}_t -\pmb{\mu}_t^s))\right)\nonumber\\
        &=&(\pmb{c}_t+\pmb\vartheta_t\pmb{b}_t)\frac{\partial\bar{\pmb{s}}_t}{\partial \pmb\vartheta_{h}}\nonumber\\
        &=&(\pmb{c}_t+\pmb\vartheta_t\pmb{b}_t)(\bar{\pmb{s}}_{h} - \pmb\mu^{s}_{h})^\top\left[\prod_{i=h+1}^{t-1}(\pmb\beta_{i}^s+\pmb\vartheta_i\pmb\beta_{i}^a)^\top\right]\left(\pmb\beta_{h}^{a}\right)^\top,
        \label{eq: NB proof 3}
\end{eqnarray}
where (\ref{eq: NB proof 3}) follows by applying \eqref{eq: expState}.
\end{itemize}
The desired conclusion follows from (\ref{eq: NB proof 1})-(\ref{eq: NB proof 3}).

\subsection{Proof of Proposition \ref{prop: time complexity analysis for Nested Backpropagaation}}\label{appendix sec: time complexity analysis for Nested Backpropagaation}

Recall that the multiplication of two matrices $A\in\mathbb{R}^{n\times m}$ and $B\in\mathbb{R}^{m\times p}$ requires $\mathcal{O}(nmp)$ time. In our setting, we are given the model parameters $\pmb{\mu}_{t}^s\in\mathbb{R}^n$, $\pmb\mu_t^a \in \mathbb{R}^m$, $\pmb{\beta}_{t}^s\in\mathbb{R}^{n\times n}$, $\pmb\beta_{t}^a\in\mathbb{R}^{m\times n}$, as well as the policy parameters $\vartheta_t\in\mathbb{R}^{n\times m}$, $\pmb{b}_t\in\mathbb{R}^m$ and $\pmb{c}_t \in \mathbb{R}^n$. We also have the expected state $\bar{\pmb s}_t \in\mathbb{R}^n$.

We now consider the computation of $\frac{\partial \bar{r}_t}{\partial \pmb \vartheta_h}$. If $t=h$, both nested backpropagation (NBP) and brute force require $\mathcal{O}(nm)$ time to compute $(\bar{\pmb{s}}_t -\pmb\mu_t^s)\pmb b_t^\top$. For each $t$ and $h < t$, it takes $n^2$ addition operations and $n^2 m$ multiplication operations to compute $\pmb\beta_{i}^s+\pmb\vartheta_i\pmb\beta_{i}^a$. Therefore, the brute-force approach, which directly computes $\pmb\delta_h^t$ in (\ref{eq: nested backpropagation}), takes
\begin{equation*}
\mathcal{O}\left((t-h) (n^2 +n^2m)+n^2m\right)=\mathcal{O}\left((t-h) (n^2 +n^2m)\right)
\end{equation*}
time when $t\neq h$. Then, to compute the gradient with respect to each parameter in the network, it takes
\begin{eqnarray*}
\mathcal{O}\left(\sum^H_{t=1}\sum_{h=1}^{t-1}(t-h) (n^2 +n^2m)\right) &=& \mathcal{O}\left(\sum^H_{t=1}\frac{t(t-1)}{2} (n^2 +n^2m)\right)\\
&=& \mathcal{O}\left(\frac{H^3-H}{2} (n^2 +n^2m)\right)\\
&=& \mathcal{O}\left(H^3 (n^2m)\right).
\end{eqnarray*}
In comparison, by storing and reusing the computation of $\mathbf{R}_{h,t-1}$ and $\pmb\beta_{i}^s+\pmb\vartheta_i\pmb\beta_{i}^a$ (see Step 1 in Algorithm~\ref{algo: NBP}), NBP takes $\mathcal{O}(n^2m)$ time to compute  $\pmb\delta_h^t$ and $\mathcal{O}(n^2+n^2m)$ time to compute $\mathbf{R}_{h,t}$. Thus, the total computational cost of NBP becomes
\begin{equation*}
\mathcal{O}\left(\sum^H_{t=1}\sum^h_{h=1}(n^2 +n^2m)+\sum^H_{t=1}\sum^h_{h=1}n^2m\right)=\mathcal{O}\left(H^2n^2m\right).
\end{equation*}

\subsection{Proof of Lemma~\ref{lemma: L-smooth}} \label{appendix subsec: lemma 1}

In the following, we will use $\Vert A\Vert_F$ to denote the Frobenius norm of a matrix $A$. A useful property of this norm is $\Vert AB \Vert_F \leq \Vert A \Vert_F \cdot \Vert B \Vert_F$ for matrices $A,B$. We also use the notation $\left|A\right|$ to represent a matrix whose elements are equal to the absolute values of the corresponding elements of $A$.

Fix $\pmb{w}\in\mathcal{W}$. From (\ref{eq: objective-simple 2}), the gradient of the objective function can be expressed as
\begin{equation}\label{eq: non-stantionary policy gradient}
    \frac{\partial J(\pmb\theta;\pmb{w})}
 {\partial {\pmb\vartheta_t}}
 = \mathbf{R}_{1,t}(\pmb{s}_1 - \pmb\mu_1^s)\pmb{b}_t^\top+\sum^H_{t^\prime=t} (\pmb{s}_1-\pmb{\mu}_1^s)(\pmb{b}_t^\top\pmb\vartheta_t^\top+\pmb{c}_t^\top)\frac{\partial \mathbf{R}_{1,t^\prime}}{\partial\pmb\vartheta_t},
\end{equation}
where $\frac{\partial \mathbf{R}_{1,t^\prime}}{\partial\pmb\vartheta_t} =\left(\prod^{t^\prime}_{i=1;i\neq t}\left({\pmb\beta_{i}^s}^\top + {\pmb\beta_{i}^a}^\top\pmb\vartheta_i^\top\right)\right){\pmb\beta_{t}^a}^\top$.

The proof proceeds in three steps: we show that $\mathcal{J}$ is differentiable; that $\frac{\partial J\left(\pmb\theta;\pmb{w}\right)}{\partial \vartheta_t}$ is $L$-smooth; and that $\mathcal{J}$ itself is $L$-smooth.

\noindent\textbf{Step 1: $\mathcal{J}$ is differentiable.}

Equation (\ref{eq: non-stantionary policy gradient}) shows that $J\left(\pmb\theta;\pmb{w}\right)$ is differentiable for any $\pmb{w}$ (note that, if $\pmb{w}\notin \mathcal{W}$, the derivative is zero). Since $\mathbb{C}$ is bounded, we can let $B_{\mathbb{C}} \equiv \max_{\pmb{x}\in\mathbb{C}}\Vert\pmb{x}\Vert_F$. Let $B\in\mathbb{R}^{n\times m}$ be a matrix with all entries equal to $B_{\mathbb{C}}$.

The function
\begin{eqnarray*}
g(\pmb{w})&=& \prod_{i=1}^t\left({|\pmb\beta_{i}^s|}^\top + {|\pmb\beta^a_i|}^\top B^\top\right)|\pmb{s}_1 - \pmb\mu_1^s|\cdot |\pmb{b}_t|^\top\\
&\,& +\sum^H_{t^\prime=t} \left|\pmb{s}_1-\pmb{\mu}_1^s\right| \left( |\pmb{b}_t|^\top B^\top+|\pmb{c}_t|^\top\right) \left(\prod^{t^\prime}_{i=1;i\neq t}\left({|\pmb\beta_{i}^s|}^\top + |\pmb\beta_i^a|^\top B^\top\right)\right){|\pmb\beta_{t}^a|}^\top
\end{eqnarray*}
is non-negative and Lebesgue integrable on $\mathcal{W}$. We have $|\frac{\partial}{\partial \pmb\theta}J(\pmb\theta;\pmb{w})|\leq g(\pmb{w})$ for all $(\pmb\theta,\pmb{w}) \in \mathbb{C}\times \mathcal{W}$. Then,
\begin{eqnarray}
\nabla \mathcal{J}(\pmb\theta)&=&\frac{\partial}{\partial \pmb\theta}\int J(\pmb\theta;\pmb{w}) p(\pmb{w}|\mathcal{D})d\pmb{w}\nonumber\\
&=&\int \frac{\partial}{\partial \pmb\theta}J(\pmb\theta;\pmb{w}) p(\pmb{w}|\mathcal{D})d\pmb{w}\label{eq:dct1}\\
&=&\int_{\mathcal{W}} \frac{\partial}{\partial \pmb\theta}J(\pmb\theta;\pmb{w}) p(\pmb{w}|\mathcal{D})d\pmb{w}\nonumber
\end{eqnarray}
where (\ref{eq:dct1}) follows by the dominated convergence theorem. We conclude that $\mathcal{J}$ is differentiable.

\noindent\textbf{Step 2: The gradient of $J\left(\pmb\theta;\pmb{w}\right)$ is $L$-smooth in $\pmb\theta$.}

Fix $\pmb{x}=\left(\pmb{x}_1,\ldots,\pmb{x}_{H-1}\right)\in \mathbb{C}$ and $\pmb{y}=\left(\pmb{y}_1,\ldots,\pmb{y}_{H-1}\right)\in \mathbb{C}$ such that $\pmb{x}_i=\pmb{y}_i$ for $i\neq t$ and $\pmb{x}_t\neq\pmb{y}_t$. Without loss of generality, we may assume $\pmb{w}\in\mathcal{W}$. We derive
\begin{eqnarray}
   &\,&\left\Vert
   \frac{\partial J(\pmb{x};\pmb{w})}
 {\partial {\pmb\vartheta_t}} -\frac{ \partial J(\pmb{y};\pmb{w})}
 {\partial {\pmb\vartheta_t}}
 \right\Vert_F  %= \Vert \left(\mathbf{R}_{1,t}(\pmb{x})-\mathbf{R}_{1,t}(\pmb{y})\right)(\pmb{s}_1 - \pmb\mu_1^s))\pmb{b}_t^\top
 \nonumber \\
  %  &&+\sum^H_{t^\prime=t} (\pmb{s}_1-\pmb{\mu}_1^s)(\pmb{b}_t^\top\pmb{x}_t^\top+\pmb{c}_t^\top)\frac{\partial \mathbf{R}_{1,t^\prime}(\pmb{x})}{\partial\pmb\vartheta_t}-\sum^H_{t^\prime=t} (\pmb{s}_1-\pmb{\mu}_1^s)(\pmb{b}_t^\top\pmb{y}_t^\top+\pmb{c}_t^\top)\frac{\partial \mathbf{R}_{1,t^\prime}(\pmb{y})}{\partial\pmb\vartheta_t}\Vert_F \nonumber\\
    &=& \left\Vert \left(\mathbf{R}_{1,t}(\pmb{x})-\mathbf{R}_{1,t}(\pmb{y})\right)(\pmb{s}_1 - \pmb\mu_1^s))\pmb{b}_t^\top
    +\sum^H_{t^\prime=t} (\pmb{s}_1-\pmb{\mu}_1^s)(\pmb{b}_t^\top\pmb{x}_t^\top+\pmb{c}_t^\top)\frac{\partial \mathbf{R}_{1,t^\prime}(\pmb{x})}{\partial\pmb\vartheta_t}
    \right. \nonumber\\
    && \left. -\sum^H_{t^\prime=t} (\pmb{s}_1-\pmb{\mu}_1^s)(\pmb{b}_t^\top\pmb{y}_t^\top+\pmb{c}_t^\top)\frac{\partial \mathbf{R}_{1,t^\prime}(\pmb{y})}{\partial\pmb\vartheta_t}\right \Vert_F
    \nonumber\\
    &=& \left \Vert \left(\mathbf{R}_{1,t}(\pmb{x})-\mathbf{R}_{1,t}(\pmb{y})\right)(\pmb{s}_1 - \pmb\mu_1^s))\pmb{b}_t^\top
    + (\pmb{s}_1-\pmb{\mu}_1^s)(\pmb{b}_t^\top\pmb{x}_t^\top+\pmb{c}_t^\top)\sum^H_{t^\prime=t}\frac{\partial \mathbf{R}_{1,t^\prime}(\pmb{x})}{\partial\pmb\vartheta_t} \right.
    \nonumber\\
    && \left.
    - (\pmb{s}_1-\pmb{\mu}_1^s)(\pmb{b}_t^\top\pmb{y}_t^\top+\pmb{c}_t^\top)\sum^H_{t^\prime=t}\frac{\partial \mathbf{R}_{1,t^\prime}(\pmb{x})}{\partial\pmb\vartheta_t} \right \Vert_F
    \nonumber\\
     & { \leq } &
     \left\Vert\pmb{s}_1 - \pmb\mu_1^s\pmb{b}_t^\top
     \right\Vert_F
     \left \Vert\mathbf{R}_{1,t}(\pmb{x})-\mathbf{R}_{1,t}(\pmb{y}) \right \Vert_F
     + \Vert(\pmb{s}_1-\pmb{\mu}_1^s)\Vert_F
     \left \Vert\pmb{b}_t^\top(\pmb{x}_t^\top-\pmb{y}_t^\top)\sum^H_{t^\prime=t}\frac{\partial \mathbf{R}_{1,t^\prime}(\pmb{x})}{\partial\pmb\vartheta_t}
     \right \Vert_F \label{eq: lemma 1 triangle inequality}\\
     &\leq&
     \Vert\pmb{s}_1 - \pmb\mu_1^s\Vert_F\Vert\pmb{b}_t\Vert_F \left\Vert\left(\prod^{t-1}_{i=1}\left(\left(\pmb\beta_{i}^s\right)^\top + \left(\pmb\beta_{i}^a\right)^\top\pmb{x}_i^\top\right)\right)\left(\pmb\beta_{t}^a\right)^\top(\pmb{x}_t-\pmb{y}_t)^\top\right\Vert_F
     \nonumber\\
     &&+ \left\Vert\pmb{s}_1-\pmb{\mu}_1^s\right\Vert_F\left\Vert\pmb{b}_t\right\Vert_F\left\Vert\pmb{x}_t-\pmb{y}_t\right\Vert_F\left\Vert \sum^H_{t^\prime=t}\frac{\partial \mathbf{R}_{1,t^\prime} (\pmb{x})}
     {\partial\pmb\vartheta_t}\right\Vert_F. \nonumber\\
    &\leq& \Vert\pmb{s}_1 - \pmb\mu_1^s\Vert_F\Vert\pmb{b}_t\Vert \left\Vert\prod^{t-1}_{i=1}\left(\left(\pmb\beta_{i}^s\right)^\top + \left(\pmb\beta_{i}^a\right)^\top\pmb{x}_i^\top\right)\right\Vert_F\left\Vert{\pmb\beta_{t}^a}\right\Vert_F\left\Vert(\pmb{x}_t-\pmb{y}_t)^\top\right\Vert_F \nonumber\\
    &&+ \left\Vert\pmb{s}_1-\pmb{\mu}_1^s\right\Vert\left\Vert\pmb{b}_t\right\Vert_F\left\Vert\pmb{x}_t-\pmb{y}_t\right\Vert_F\left\Vert \sum^H_{t^\prime=t}\frac{\partial \mathbf{R}_{1,t^\prime}(\pmb{x})}{\partial\pmb\vartheta_t}\right\Vert_F,
     \label{eq: upper bound for L smooth}\nonumber\\
    &\leq& (\Vert\pmb{s}_1\Vert + \Vert \pmb\mu_1^s\Vert)\Vert\pmb{b}_t\Vert_F \left\Vert\prod^{t-1}_{i=1}\left({\pmb\beta_{i}^s}^\top + {\pmb\beta_{i}^a}^\top\pmb{x}_i^\top\right)\right\Vert_F\left\Vert{\pmb\beta_{t}^a}\right\Vert_F\left\Vert(\pmb{x}_t-\pmb{y}_t)^\top\right\Vert_F \nonumber\\
    &&+ (\left\Vert\pmb{s}_1\right\Vert+\left\Vert\pmb{\mu}_1^s\right\Vert)\left\Vert\pmb{b}_t\right\Vert\left\Vert\pmb{x}_t-\pmb{y}_t\right\Vert_F\left\Vert \sum^H_{t^\prime=t}\frac{\partial \mathbf{R}_{1,t^\prime}(\pmb{x})}{\partial\pmb\vartheta_t}\right\Vert_F,
     \label{eq: upper bound for L smooth 2}
\end{eqnarray}
where (\ref{eq: lemma 1 triangle inequality}) and (\ref{eq: upper bound for L smooth 2}) follow from the triangle inequality.

We can rewrite (\ref{eq: upper bound for L smooth 2}) as
\begin{equation} \label{eq: squared upper bound for L smooth 2}
    \left\Vert
   \frac{\partial J(\pmb{x};\pmb{w})}
 {\partial {\pmb\vartheta_t}} -\frac{ \partial J(\pmb{y};\pmb{w})}
 {\partial {\pmb\vartheta_t}}
 \right\Vert_F \leq L_t(\pmb{w})\left\Vert\pmb{x}_t-\pmb{y}_t\right\Vert_F
\end{equation}
where
\begin{equation*}
    L_t(\pmb{w})=(\Vert\pmb{s}_1\Vert + \Vert \pmb\mu_1^s\Vert)\Vert\pmb{b}_t\Vert_F \left\Vert\prod^{t-1}_{i=1}\left({\pmb\beta_{i}^s}^\top + {\pmb\beta_{i}^a}^\top\pmb{x}_i^\top\right)\right\Vert_F\left\Vert{\pmb\beta_{t}^a}\right\Vert_F + (\Vert\pmb{s}_1\Vert + \Vert \pmb\mu_1^s\Vert)\left\Vert\pmb{b}_t\right\Vert_F\left\Vert \sum^H_{t^\prime=t}\frac{\partial \mathbf{R}_{1,t^\prime}(\pmb{x})}{\partial\pmb\vartheta_t}\right\Vert_F \nonumber
\end{equation*}
Since $\mathcal{W}$ is bounded, define $B_{\mathcal{W}} = \max_{\pmb{w}\in\mathcal{W}}\Vert\pmb{w}\Vert$ to be a constant bounding $\pmb{w}\in\mathcal{W}$. Recall that we also have $\Vert\pmb{x}_t\Vert_F\leq B_{\mathbb{C}}$. Then, by applying the triangle inequality, we have
\begin{eqnarray}
    \left\Vert\frac{\partial \mathbf{R}_{1,t^\prime}(\pmb{x})}{\partial\pmb\vartheta_t} \right\Vert_F
    % =\left(\prod^{t^\prime}_{i=1;i\neq t}\left\Vert{\pmb\beta_{i}^s}^\top + {\pmb\beta_{i}^a}^\top\pmb\vartheta_i^\top\right\Vert_F\right)\left\Vert\pmb\beta_{t}^a\right\Vert_F
    &\leq& \left(\prod^{t^\prime}_{i=1;i\neq t}\left(\Vert\pmb\beta_{i}^s\Vert_F + \Vert{\pmb\beta_{i}^a}\Vert_F \Vert\pmb\vartheta_i\Vert_F\right)\right)\Vert\pmb\beta_{t}^a\Vert_F\nonumber\\
    &\leq& \left(\prod^{t^\prime}_{i=1;i\neq t}\left(B_{\mathcal{W}}  + B_{\mathcal{W}} B_{\mathbb{C}}\right)\right) B_{\mathcal{W}}, \label{eq.mid4}
\end{eqnarray}
and
\begin{equation}
    \left\Vert \prod^{t-1}_{i=1}\left({\pmb\beta_{i}^s}^\top + {\pmb\beta_{i}^a}^\top\pmb{x}_i^\top\right)\right\Vert_F \leq \prod^{t-1}_{i=1}\left(\Vert\pmb\beta_{i}^s\Vert_F + \Vert\pmb\beta_{i}^a\Vert_F \Vert\pmb\vartheta_i\Vert_F\right)\leq \prod^{t-1}_{i=1}\left(B_{\mathcal{W}} +B_{\mathcal{W}} B_{\mathbb{C}}\right).
    \label{eq.mid5}
\end{equation}
By plugging (\ref{eq.mid4}) and (\ref{eq.mid5}) into \eqref{eq: squared upper bound for L smooth 2}, we bound
\begin{eqnarray*}
L_t(\pmb\omega) &\leq& (\Vert\pmb{s}_1\Vert +  B_{\mathcal{W}})\Vert\pmb{b}_t\Vert \prod^{t-1}_{i=1}\left(B_{\mathcal{W}} +B_{\mathcal{W}} B_{\mathbb{C}}\right)B_{\mathcal{W}}\\
&\,& + \left(\Vert\pmb{s}_1\Vert +  B_{\mathcal{W}}\right)\left\Vert\pmb{b}_t\right\Vert\left\Vert \sum^H_{t^\prime=t}\left(\prod^{t^\prime}_{i=1;i\neq t}\left(B_{\mathcal{W}}  + B_{\mathcal{W}} B_{\mathbb{C}}\right)\right) B_{\mathcal{W}}\right\Vert_F.
\end{eqnarray*}
Since this bound has no dependence on $\pmb{w}$, the desired property is obtained.

\noindent\textbf{Step 3: $\mathcal{J}$ is L-smooth.}

For any $\pmb{x},\pmb{y} \in \mathbb{C}$, we have
\begin{eqnarray}
\Vert \nabla_{\pmb\theta} J(\pmb{x};\pmb{w}) - \nabla_{\pmb\theta} J(\pmb{y};\pmb{w})\Vert &\leq& \left(\sum^{H-1}_{t=1}\left\Vert \mbox{vec}\left(
 \frac{\partial J(\pmb{x};\pmb{w})}
 {\partial {\pmb\vartheta_t}}
 \right) - \mbox{vec}\left(
 \frac{ \partial J(\pmb{y};\pmb{w})}
 {\partial {\pmb\vartheta_t}}\right)
 \right\Vert^2\right)^{1/2} \nonumber\\
    &=&\left(\sum^{H-1}_{t=1}\left\Vert \frac{\partial J(\pmb{x};\pmb{w})}
 {\partial {\pmb\vartheta_t}}
 - \frac{ \partial J(\pmb{y};\pmb{w})}
 {\partial {\pmb\vartheta_t}}
 \right\Vert_F^2\right)^{1/2} \nonumber\\
    &\leq& \left(\sum^{H-1}_{t=1} L_t^2\Vert \pmb{x}_t - \pmb{y}_t\Vert_F^2\right)^{1/2}\nonumber\\
     &\leq& \max_{t\in \mathcal{T}}\{L_t\}\left(\sum^{H-1}_{t=1}\Vert \pmb{x}_t - \pmb{y}_t\Vert_F^2\right)^{1/2} \nonumber\\
      %&=& \max_{t\in \mathcal{T}}\{L_t\}\left(\sum^{H-1}_{t=1}\Vert \mbox{vec}(\pmb{x}_t) - \mbox{vec}(\pmb{y}_t)\Vert^2\right)^{1/2} \nonumber\\
     &=&\max_{t\in \mathcal{T}}\{L_t\}\Vert\pmb{x}-\pmb{y}\Vert.
     \label{eq : lemma 1 L-smooth}
\end{eqnarray}
Consequently, $\Vert\nabla_{\pmb{\theta}} J(\pmb{x};\pmb{w}) -\nabla_{\pmb{\theta}} J(\pmb{y};\pmb{w})\Vert \leq L\lVert \pmb{x}-\pmb{y}\rVert$, where $L=\max_{t\in \mathcal{T}}\left\{L_t(\pmb{w})\right\}$.

Proceeding along similar lines, one can straightforwardly show the boundedness of the gradient
\begin{equation*}
    \left\Vert\nabla_{\pmb{\theta}} J(\pmb{x}, \pmb{\omega})\right\Vert =\left(\sum_{t=1}^H\left\Vert\mbox{vec}\left( \frac{\partial J(\pmb{x}, \pmb{\omega})}{\partial \pmb{\vartheta}_t}\right)\right\Vert^2\right)^{1/2} =\left(\sum_{t=1}^H\left\Vert \frac{\partial J(\pmb{x}, \pmb{\omega})}{\partial \pmb{\vartheta}_t}\right\Vert_F^2\right)^{1/2}
\end{equation*}
 for any $\pmb{x}\in \mathbb{C}$, where
 \begin{eqnarray*}
  \left\Vert
  \frac{\partial J(\pmb{x};\pmb{w})}
 {\partial {\pmb\vartheta_t}}
 \right\Vert_F &=&\left\Vert \mathbf{R}_{1,t}(\pmb{x})(\pmb{s}_1 - \pmb\mu_1^s))\pmb{b}_t^\top
    +\sum^H_{t^\prime=t} (\pmb{s}_1-\pmb{\mu}_1^s)(\pmb{b}_t^\top\pmb{x}_t^\top+\pmb{c}_t^\top)\frac{\partial \mathbf{R}_{1,t^\prime}(\pmb{x})}{\partial\pmb\vartheta_t}
    \right\Vert_F\\
    &\leq& \prod^{t-1}_{i=1}\left(B_{\mathcal{W}} + B_{\mathcal{W}}B_{\mathbb{C}}\right) (\left\Vert\pmb{s}_1\right\Vert + B_{\mathcal{W}})\Vert\pmb{b}_t\Vert\\
    &\,& +\sum^H_{t^\prime=t} (\left\Vert\pmb{s}_1\right\Vert + B_{\mathcal{W}})(\Vert\pmb{b}_t\Vert B_{\mathbb{C}}+\Vert\pmb{c}_t\Vert)\left(\prod^{t^\prime}_{i=1;i\neq t}\left(B_{\mathcal{W}} + B_{\mathcal{W}} B_{\mathbb{C}}\right)\right)B_{\mathcal{W}}.
    \end{eqnarray*}
Thus by the bounded convergence theorem, we have
\begin{equation}\label{eq: lemma 1 interchange}
    \nabla_{\pmb{\theta}}\E_{\pmb{w}}[{J}(\pmb{y};\pmb{w})]=\E_{\pmb{w}}[\nabla_{\pmb{\theta}} {J}(\pmb{x};\pmb{w})]
\end{equation}
Then, for any $\pmb{x},\pmb{y}\in\mathbb{C}$, we have
\begin{eqnarray}
    \Vert\nabla \mathcal{J}(\pmb x)-\nabla  \mathcal{J}(\pmb y)\Vert_F
    &=&\Vert\nabla _{\pmb{\theta}} \E_{\pmb{w}}[{J}(\pmb{x};\pmb{w})]-\nabla_{\pmb{\theta}}
    \E_{\pmb{w}}[{J}(\pmb{y};\pmb{w})]\Vert \nonumber\\
    &\leq& \Vert\E_{\pmb{w}}[\nabla_{\pmb{\theta}} {J}(\pmb{x};\pmb{w})]-\E_{\pmb{w}}[\nabla_{\pmb{\theta}}
    {J}(\pmb{y};\pmb{w})]\Vert\label{eq: lemma 1 regularity} \\
    &=& \Vert\E_{\pmb{w}}[\nabla_{\pmb{\theta}} {J}(\pmb{x};\pmb{w})-\nabla_{\pmb{\theta}}
    {J}(\pmb{y};\pmb{w})]\Vert \nonumber\\
    &\leq& \E_{\pmb{w}}\left[\Vert\nabla_{\pmb{\theta}} {J}(\pmb{x};\pmb{w})-\nabla_{\pmb{\theta}}
    {J}(\pmb{y};\pmb{w})\Vert\right] \label{eq: lemma1 jensen}\\
    &=& L\lVert \pmb{x}-\pmb{y}\rVert \label{eq: lemma1 L-smooth final}
\end{eqnarray}
where \eqref{eq: lemma 1 regularity} follows from \eqref{eq: lemma 1 interchange},
\eqref{eq: lemma1 jensen} is due to Jensen's inequality and the convexity of the norm, and \eqref{eq: lemma1 L-smooth final} follows from \eqref{eq : lemma 1 L-smooth}.

\subsection{Proof of Corollary~\ref{corollary: bounded gradient}} \label{appendix: corollary proof}

Let $\pmb\theta^\star$ be the unique global optimum. Assumption~\ref{assumption 3} implies $ \nabla \mathcal{J}(\pmb\theta^\star)=0$. By Lemma~\ref{lemma: L-smooth},
\begin{equation*}
\Vert\nabla \mathcal{J}(\pmb{x})\Vert=\Vert\nabla \mathcal{J}(\pmb{x}) - \nabla \mathcal{J}(\pmb\theta^\star)\Vert \leq L\Vert\pmb{x} - \pmb\theta^\star\Vert\leq L\cdot \max_{\pmb{y}\in \mathbb{C}}\Vert\pmb{y} -\pmb\theta^\star\Vert.
\end{equation*}

\subsection{Proof of Lemma~\ref{lemma: p(theta) convergence as}}
% ================

\noindent\textbf{Step 1: For any $\theta\in\mathbb{C}$ and small enough $\eta$, $\pmb\theta + \eta\nabla \mathcal{J}(\pmb\theta)\in \mathbb{C}$.}

Recall that, by Assumption \ref{assumption 1}, there exists $c_0$ such that, for all $\pmb\theta\in\partial\mathbb{C}$, the update $\pmb\theta + \eta\frac{\nabla{J}(\pmb\theta;\pmb{w})}{\Vert\nabla{J}(\pmb\theta;\pmb{w})\Vert}  \in\mathbb{C}$ for $\eta \leq c_0$.

Denote by
\begin{equation*}
\mathbb{U}(c_0)=\left\{\eta\frac{\nabla{J}(\pmb\theta:\pmb{w})}{\Vert\nabla{J}(\pmb\theta;\pmb{w})\Vert}:\pmb{w}\in \mathcal{W} \mbox{ and } \eta\leq c_0\right\}.
\end{equation*}
Let $B_{\pmb\theta}(c_0)$ denote the ball with center $\pmb\theta$ and radius $c_0$. Define the convex cone
% for all unit vector $\frac{\nabla{J}(\pmb\theta;\pmb{w})}{\Vert\nabla{J}(\pmb\theta;\pmb{w})\Vert}$
\begin{equation*}
\mathbb{M}=\left\{\eta (\pmb{v}-\pmb\theta):\pmb{v}\in B_{\pmb \theta}(c_0)\cap \mathbb{C} \mbox{ and } \eta\geq 0\right\}
\end{equation*}
Since $\pmb\theta + \eta \frac{\nabla{J}(\pmb\theta;\pmb{w})}{\Vert\nabla{J}(\pmb\theta;\pmb{w})\Vert}\in B_{\pmb\theta}(c_0)\cap \mathbb{C}$ for $\eta\leq c_0$ and for all $\pmb{w}\in\mathcal{W}$, we have $\mathbb{U}(c_0)\subset \mathbb{M}$. If we choose the stepsize as $\eta= \Vert\nabla{J}(\pmb\theta;\pmb{w})\Vert$, then we can rewrite $\nabla_{\pmb\theta} J(\pmb\theta;\pmb{w}) =\eta\frac{\nabla{J}(\pmb\theta;\pmb{w})}{\Vert\nabla{J}(\pmb\theta;\pmb{w})\Vert}\in \mathbb{M}$. Writing out
\begin{equation*}
\nabla\mathcal{J}(\pmb\theta)=\E[\nabla_{\pmb\theta} J(\pmb\theta;\pmb{w})]=\int \nabla_{\pmb\theta} J(\pmb\theta;\pmb{w})p(\pmb{w}|\mathcal{D})d\pmb{w}=\int_{\mathcal{W}} \nabla_{\pmb\theta} J(\pmb\theta;\pmb{w})p(\pmb{w}|\mathcal{D})d\pmb{w},
\end{equation*}
the convexity of $\mathbb{M}$ implies $\nabla\mathcal{J}(\pmb\theta)\in\mathbb{M}$, by generalizing the definition of convex combination to include integrals and probability distributions \citep{boyd2004convex}. Moreover, by the definition of a cone, we have $\eta\frac{\nabla\mathcal{J}(\pmb\theta)}{\left\Vert\nabla\mathcal{J}(\pmb\theta)\right\Vert}\in\mathbb{M}$ as well.

Since the norm of the vector $\eta\frac{\nabla\mathcal{J}(\pmb\theta)}{\left\Vert\nabla\mathcal{J}(\pmb\theta)\right\Vert}$ equals $\eta$, we have $\eta\frac{\nabla\mathcal{J}(\pmb\theta)}{\left\Vert\nabla\mathcal{J}(\pmb\theta)\right\Vert} \in B_{\pmb 0}(c_0)\cap \mathbb{M}$ for $\eta\leq c_0$, with $\pmb{0}$ denoting the origin. By the definition of $\mathbb{M}$,
\begin{equation*}
\pmb\theta+\eta\frac{\nabla\mathcal{J}(\pmb\theta)}{\left\Vert\nabla\mathcal{J}(\pmb\theta)\right\Vert} \in B_{\pmb \theta}(c_0)\cap \mathbb{C}, \qquad \eta \leq c_0.
\end{equation*}
Let $G$ be the bound on $\Vert\nabla \mathcal{J}(\pmb\theta)\Vert$ obtained from Corollary \ref{corollary: bounded gradient}. Then
\begin{equation*}
\pmb\theta+\frac{\eta}{G}\nabla\mathcal{J}(\pmb\theta) \in B_{\pmb \theta}(c_0)\cap \mathbb{C}, \qquad \eta\leq c_0,
\end{equation*}
whence the desired result follows.

\noindent\textbf{Step 2: Update is in $\mathbb{C}$ when $\pmb{\theta}$ is close to the boundary.}

By Assumption \ref{assumption 1}, for any $\pmb x\in\partial \mathbb{C}$, there exists a neighborhood $B_{\pmb x}(\epsilon_{\pmb x})$ with radius $\epsilon_{\pmb x}>0$ such that
\begin{equation*}
    \pmb\theta + \eta\nabla \mathcal{J}(\pmb\theta)\in\mathbb{C},  \qquad \pmb\theta \in B_{\pmb x}(\epsilon_{\pmb x})\cap \mathbb{C},\quad \eta \leq \frac{c_0}{G}.
    \label{eq:mid7}
\end{equation*}
Define
\begin{equation}
    \rho_{\pmb x} =\sup\left\{\epsilon: \pmb\theta+\eta \nabla \mathcal{J}(\pmb\theta)\in \mathbb{C}, \qquad \pmb\theta\in B_{\pmb x}(\epsilon)\cap\mathbb{C}, \quad\eta \leq \frac{c_0}{G}\right\}.
    \label{eq.mid8}
\end{equation}
Obviously, $\rho_{\pmb x} > \epsilon_{\pmb{x}} > 0$. We wish to show that $\inf_{\pmb x} \rho_{\pmb x} > 0$.

We proceed by contradiction. Suppose that  $\inf_{\pmb x}\rho_{\pmb x}=0$. Then, there must exist $\pmb x\in \partial \mathbb{C}$ and a sequence $\{\pmb x_n\}_{n=1}^\infty \in\partial \mathbb{C}$ satisfying $\pmb x_n \rightarrow\pmb x$, such that $\rho_{\pmb x_n} \rightarrow 0$ as $n\rightarrow \infty$. At the same time, we must also have $\rho_{\pmb x}>0$. For all sufficiently large $n$, we have $\Vert\pmb x_n - \pmb x\Vert < \frac{1}{2}\rho_{\pmb x}$ and at the same time $\rho_{\pmb x_n} <\frac{1}{2}\rho_{\pmb x}$.
For any such $n$, take any $\pmb\theta\in B_{\pmb x_n}(\frac{1}{2}\rho_{\pmb x})$. Then, by the triangle inequality, we obtain
\begin{equation*}
    \Vert\pmb\theta-\pmb x\Vert\leq \Vert \pmb\theta -\pmb x_n\Vert+\Vert \pmb x_n - \pmb x\Vert < \rho_{\pmb x}.
\end{equation*}
Therefore, $\pmb\theta \in B_{\pmb x}\left(\rho_x\right)$, and by the definition of $\rho_{\pmb x}$ in (\ref{eq.mid8}) we have $\theta+\eta \nabla \mathcal{J}(\pmb\theta)\in \mathbb{C}$ for all $\eta \leq \frac{c_0}{G}$. But, by the same definition of $\rho_{\pmb{x}_n}$, it follows that $B_{\pmb x_n}(\frac{1}{2}\rho_{\pmb x}) \subseteq B_{\pmb{x}_n}\left(\rho_{\pmb{x}_n}\right)$, which contradicts the fact that $\rho_{\pmb x_n} <\frac{1}{2}\rho_{\pmb x}$. We conclude that $\inf_{\pmb x} \rho_{\pmb x}>0$.

\noindent\textbf{Step 3: Final result}.

Let $\rho_{\min}=\inf_{\pmb x} \rho_{\pmb x}$ and suppose that $\pmb\theta_k\in B_{\pmb x}(\rho_{\min})\cap \mathbb{C}$. We wish to show that the event of the updated parameters moving a distance more than $\frac{c_0}{G}$ in the direction of $\nabla \widehat{\mathcal{J}}(\pmb{\pi}_{\pmb{\theta}})$ can occur at most finitely many times. That is,
\begin{equation}\label{eq:finitesumofevents}
\sum_k 1_{\left\{\Vert\nabla\widehat{\mathcal{J}}(\pmb\theta_k)\Vert>\frac{c_0}{\eta_kG}\right\}} < \infty
\end{equation}
almost surely. To show this, we derive
\begin{eqnarray}
\sum^\infty_{k=1}P\left(\left\Vert \nabla\widehat{\mathcal{J}}(\pmb\theta_k)\right\Vert>\frac{c_0}{\eta_kG}\right)
 &\leq&\sum^\infty_{k=1} \frac{\E\left[\left\Vert\nabla\widehat{\mathcal{J}}(\pmb\theta_k)\right\Vert^2\right]}{\frac{c_0^2}{\eta_k^2 G^2}} \label{eq: lemma 6 (1-1)}\\
    &\leq&\sum^\infty_{k=1}\eta_k^2\frac{G^2\E\left[\left(2\max\left\{\left\Vert\nabla\widehat{\mathcal{J}}(\pmb\theta_k)-\nabla{\mathcal{J}}(\pmb\theta_k)\right\Vert, \left\Vert\nabla{\mathcal{J}}(\pmb\theta_k)\right\Vert\right\}\right)^2\right]}{c_0^2}
    \label{eq.mid9-1} \\
    &=&4G^2\sum^\infty_{k=1}\eta_k^2\frac{\E\left[\left\Vert\nabla\widehat{\mathcal{J}}(\pmb\theta_k)-\nabla{\mathcal{J}}(\pmb\theta_k)\right\Vert^2+ \left\Vert\nabla{\mathcal{J}}(\pmb\theta_k)\right\Vert^2\right]}{c_0^2} \nonumber\\
    %&=&4\sum^\infty_{k=1}\eta_k^2\frac{\E\left[\max\left\{\left\Vert\nabla\widehat{\mathcal{J}}(\pmb\theta_k)-\nabla{\mathcal{J}}(\pmb\theta_k)\right\Vert^2, \left\Vert\nabla{\mathcal{J}}(\pmb\theta_k)\right\Vert^2\right\}\right]}{\rho_{min}^2} \nonumber\\
    &\leq&4G^2\sum^\infty_{k=1}\eta_k^2\frac{\E\left[\left\Vert\nabla\widehat{\mathcal{J}}(\pmb\theta_k)-\nabla{\mathcal{J}}(\pmb\theta_k)\right\Vert^2\right]+\E\left[\left\Vert\nabla{\mathcal{J}}(\pmb\theta_k)\right\Vert^2\right]}{c_0^2} \nonumber\\
    &\leq&\frac{4G^2(\sigma^2+G^2)}{c_0^2}\sum^\infty_{k=1}\eta_k^2\label{eq: lemma 6 (1-3)}\\
    &<& \infty, \nonumber
\end{eqnarray}
where \eqref{eq: lemma 6 (1-1)} is due to Markov's inequality, (\ref{eq.mid9-1}) follows from the triangle inequality, and \eqref{eq: lemma 6 (1-3)} holds due to Assumption \ref{assumption 2} and Corollary \ref{corollary: bounded gradient}. We then obtain (\ref{eq:finitesumofevents}) by the Borel-Cantelli lemma.

In a very similar manner, we can show $\sum_k 1_{\left\{\Vert\nabla\widehat{\mathcal{J}}(\pmb\theta_k)\Vert>\frac{\rho_{\min}}{\eta_k}\right\}} < \infty$ by deriving
\begin{eqnarray}
 \sum^\infty_{k=1}P\left(\left\Vert \nabla\widehat{\mathcal{J}}(\pmb\theta_k)\right\Vert>\frac{\rho_{\min}}{\eta_k}\right)
 &\leq& \sum^\infty_{k=1} \frac{\E\left[\left\Vert\nabla\widehat{\mathcal{J}}(\pmb\theta_k)\right\Vert^2\right]}{\rho_{\min}^2/\eta_k^2}
 \label{eq: lemma 6 (1)}\\
    &\leq&\sum^\infty_{k=1}\eta_k^2\frac{\E\left[\left(2\max\left\{\left\Vert\nabla\widehat{\mathcal{J}}(\pmb\theta_k)-\nabla{\mathcal{J}}(\pmb\theta_k)\right\Vert, \left\Vert\nabla{\mathcal{J}}(\pmb\theta_k)\right\Vert\right\}\right)^2\right]}{\rho_{min}^2}
    \label{eq.mid9} \\
    &=&4\sum^\infty_{k=1}\eta_k^2\frac{\E\left[\left\Vert\nabla\widehat{\mathcal{J}}(\pmb\theta_k)-\nabla{\mathcal{J}}(\pmb\theta_k)\right\Vert^2+ \left\Vert\nabla{\mathcal{J}}(\pmb\theta_k)\right\Vert^2\right]}{\rho_{min}^2} \nonumber\\
    %&=&4\sum^\infty_{k=1}\eta_k^2\frac{\E\left[\max\left\{\left\Vert\nabla\widehat{\mathcal{J}}(\pmb\theta_k)-\nabla{\mathcal{J}}(\pmb\theta_k)\right\Vert^2, \left\Vert\nabla{\mathcal{J}}(\pmb\theta_k)\right\Vert^2\right\}\right]}{\rho_{min}^2} \nonumber\\
    &\leq&4\sum^\infty_{k=1}\eta_k^2\frac{\E\left[\left\Vert\nabla\widehat{\mathcal{J}}(\pmb\theta_k)-\nabla{\mathcal{J}}(\pmb\theta_k)\right\Vert^2\right]+\E\left[\left\Vert\nabla{\mathcal{J}}(\pmb\theta_k)\right\Vert^2\right]}{\rho_{min}^2} \nonumber\\
    &\leq&\frac{4(\sigma^2+G^2)}{\rho_{min}^2}\sum^\infty_{k=1}\eta_k^2\label{eq: lemma 6 (3)}\\
    &<& \infty, \nonumber
\end{eqnarray}
where, again, \eqref{eq: lemma 6 (1)} is due to Markov's inequality, (\ref{eq.mid9}) follows from the triangle inequality, and \eqref{eq: lemma 6 (3)} holds due to Assumption~\ref{assumption 2} and Corollary~\ref{corollary: bounded gradient}.

Thus, we have
\begin{equation*}
\lim_{k\rightarrow \infty} 1_{\left\{\pmb\theta_{k}+\eta_k\nabla \widehat{\mathcal{J}}(\pmb\theta_k)\notin\mathbb{C}\right\}}=0 \mbox{ a.s.},
\end{equation*}
that is, the updated parameters will always fall within the feasible region for large enough $k$. Applying the dominated convergence theorem, we obtain $p\left(A^c_k\right) \rightarrow 0$, as required.

% ========================

\subsection{Proof of Theorem~\ref{theorem:convergence}}
\label{subsec:proofConvergence}

The proof uses two technical lemmas, which are stated below and proved in separate subsections of this Appendix.

\begin{lemma}
\label{graident mapping upper bound}
For any $\pmb\theta \in \mathbb{C}$, we have
$\E[\Vert \widehat{g}_c(\pmb\theta)\Vert^2]\leq 4\Vert \nabla
\mathcal{J}(\pmb\theta)\Vert^2 +4\sigma^2$.
\end{lemma}

\begin{lemma}\label{lemma: cross grdients}
At iteration $k$, if $\pmb\theta_k\in \mathbb{C}$ and $\pmb\theta_k+\eta_k\nabla\widehat{\mathcal{J}}(\pmb\theta_k)\notin \mathbb{C}$, we have the following inequality,
\begin{equation}
    \E[\langle \nabla \mathcal{J}(\pmb\theta_k),\eta_k \widehat{g}_{c}(\pmb\theta_k)\rangle] \geq -\eta_k(G^2+G\sigma)\left(1-p(A_k)\right)^{1/2} +\eta_k\E\left[\Vert \nabla \mathcal{J}(\pmb\theta_k)\Vert^2\right] \nonumber
\end{equation}
\end{lemma}

Let $\mathcal{J}^\star = \mathcal{J}(\pmb\theta^\star)$, and recall the definition
\begin{equation*}
\widehat{g}_{c}(\pmb\theta_k)=\frac{1}{\eta_k}\left(\Pi_{C}\left(\pmb\theta_k+\eta_k\nabla\widehat{\mathcal{\mathcal{J}}}(\pmb{\theta}_k)\right)-\pmb\theta_k\right).
\end{equation*}
By the property (\ref{eq:nesterov}), which follows from Lemma~\ref{lemma: L-smooth}, we obtain
\begin{equation} \label{eq: theorem equation 0}
\mathcal{J}(\pmb \theta_{k}) - \mathcal{J}(\pmb{\theta}_{k+1}) \leq -\langle \nabla \mathcal{J}(\pmb\theta_k),\eta_k \widehat{g}_{c}(\pmb\theta_k)\rangle + L\Vert \pmb\theta_{k+1}-\pmb\theta_k\Vert^2.
\end{equation}
We will take the expectation of both sides of (\ref{eq: theorem equation 0}). By Lemma \ref{lemma: cross grdients}, we have
\begin{equation*}
    -\E\left[\langle\nabla \mathcal{J}(\pmb\theta_k), \eta_k \widehat{g}_{c}(\pmb\theta_k)\rangle \right] \leq
    \eta_k\left(G^2+G\sigma\right)\left(1-p(A_k)\right)^{1/2} -\eta_k\E\left[\Vert \nabla \mathcal{J}(\pmb\theta_k)\Vert^2\right].
    % \frac{\eta_k(\sigma^4+G^4+1)}{2} \left(1-p(A_k)\right)^{1/2} -\eta_k\E\left[\Vert \nabla \mathcal{J}(\pmb\theta_k)\Vert^2\right]
\end{equation*}
Consequently, \eqref{eq: theorem equation 0} leads to
\begin{eqnarray}
      \E\left[\mathcal{J}(\pmb\theta_k)\right]-\E\left[\mathcal{J}(\pmb\theta_{k+1})\right] &\leq&  -{\eta_k}\E\left[\Vert\nabla \mathcal{J}(\pmb\theta_k)\Vert^2\right] +\eta_k(G^2+G\sigma)\left(1-p(A_k)\right)^{1/2}\nonumber\\
       &\,& + L\eta_k^2 \E\left[\Vert \widehat{g}_{c}(\pmb\theta_k)\Vert^2\right]. \label{eq: law of total expectation 1}
\end{eqnarray}
Rearranging both sides of \eqref{eq: law of total expectation 1} gives
\begin{eqnarray}
   \eta_k\E\left[\Vert\nabla \mathcal{J}(\pmb\theta_k)\Vert^2\right] &\leq&  \left(\E\left[\mathcal{J}(\pmb\theta_{k+1})\right] -\E\left[\mathcal{J}(\pmb\theta_k)\right] \right)+ \eta_k(G^2+G\sigma)\left(1-p(A_k)\right)^{1/2}\nonumber\\
    &\,& + L\eta_k^2 \E\left[\Vert \widehat{g}_{c}(\pmb\theta_k)\Vert^2\right]. \label{eq: law of total expectation 2}
\end{eqnarray}
By applying Lemma~\ref{graident mapping upper bound}, we can rewrite \eqref{eq: law of total expectation 2} as
\begin{eqnarray}
    \eta_k\E\left[\Vert\nabla \mathcal{J}(\pmb\theta_k)\Vert^2\right] &\leq&  \left(\E\left[\mathcal{J}(\pmb\theta_{k+1})\right] -\E\left[\mathcal{J}(\pmb\theta_k)\right] \right)+\eta_k(G^2+G\sigma)\left(1-p(A_k)\right)^{1/2}\nonumber\\
     &\,& + 4L\eta_k^2 \E\left[\Vert \nabla \mathcal{J}(\pmb\theta_k)\Vert^2\right] +4L\eta_k^2\sigma^2\label{eq: law of total expectation 3}
\end{eqnarray}
By moving $4L\eta_k^2\E\left[\Vert \nabla \mathcal{J}(\pmb\theta_k)\Vert^2\right]$ to the left-hand side of \eqref{eq: law of total expectation 3}, we have
\begin{equation*}
    \eta_k\left(1-4\eta_k L\right)\E\left[\Vert\nabla \mathcal{J}(\pmb\theta_k)\Vert^2\right] \leq \left(\E\left[\mathcal{J}(\pmb\theta_{k+1})\right] -\E\left[\mathcal{J}(\pmb\theta_k))\right] \right)+\eta_k(G^2+G\sigma)\left(1-p(A_k)\right)^{1/2}+ 4 L\eta_k^2\sigma^2.
\end{equation*}
When $\eta_k$ small enough that $1 - 4\eta_kL \geq \frac{1}{2}$, or, equivalently, $\eta_k \leq\frac{1}{8L}$, we obtain the simplified inequality
\begin{equation*}
    \eta_k\left(1-4\eta_k L\right)\E\left[\Vert\nabla \mathcal{J}(\pmb\theta_k)\Vert^2\right] \geq\frac{\eta_k}{2}\E\left[\Vert\nabla \mathcal{J}(\pmb\theta_k)\Vert^2\right].
\end{equation*}
Thus, we have
\begin{eqnarray}
\frac{1}{2}\E\left[\Vert\nabla \mathcal{J}(\pmb\theta_k)\Vert^2\right] &\leq&  \left(\frac{1}{\eta_k}\E\left[\mathcal{J}(\pmb\theta_{k+1})\right] -\frac{1}{\eta_k}\E\left[\mathcal{J}(\pmb\theta_k))\right] \right)\nonumber\\
&\,& +\left(G^2+G\sigma\right)\left(1-p(A_k)\right)^{1/2}+ 4L\eta_k \sigma^2.\label{eq.mid7}
\end{eqnarray}
Then, by applying (\ref{eq.mid7}) and $\mathcal{J}(\pmb\theta_k)\leq \mathcal{J}^\star$, and summing over $k = 1,2,\ldots, K$, we obtain
\begin{eqnarray*}
&\,&    \frac{1}{2}\sum_{k=1}^K\E\left[\Vert\nabla \mathcal{J}(\pmb\theta_k)\Vert^2\right]\nonumber \\
&\leq&-\frac{1}{\eta_1}\E\left[\mathcal{J}(\pmb\theta_1)\right] +\frac{1}{\eta_{K}}\E\left[\mathcal{J}(\pmb\theta_{K+1})\right]+ \sum_{k=2}^K\left(\frac{1}{\eta_{k-1}}-\frac{1}{\eta_k}\right)\E\left[\mathcal{J}(\pmb\theta_k)\right] \nonumber\\
    &&+(G^2+G\sigma)\sum_{k=1}^K(1-p(A_k))^{1/2}
     + 4L\sigma^2\sum_{k=1}^K\eta_k  \nonumber\\
    &\leq&  -\frac{1}{\eta_1} \mathcal{J}(\pmb\theta_1)+\frac{1}{\eta_{K}}\mathcal{J}^\star + \sum_{k=2}^K\left(\frac{1}{\eta_{k-1}}-\frac{1}{\eta_k}\right)\mathcal{J}^\star +(G^2+G\sigma)\sum_{k=1}^K(1-p(A_k))^{1/2}+ 4L\sigma^2\sum_{k=1}^K\eta_k \nonumber\\
    &\leq&  \frac{1}{\eta_1} \left(\mathcal{J}^\star -\mathcal{J}(\pmb\theta_1) \right)  +(G^2+G\sigma)\sum_{k=1}^K(1-p(A_k))^{1/2}+ 4L\sigma^2\sum_{k=1}^K\eta_k.
\end{eqnarray*}
%For stepsize $\eta_k = \frac{1}{k^{1/2+\epsilon}}$, $\epsilon>0$,
Multiplying through by $\frac{2}{K}$ gives the result stated in Theorem \ref{theorem:convergence}.

\subsection{Proof of Lemma \ref{graident mapping upper bound}} \label{subsec:proof_lemmas}

%Since $\nabla \widehat{\mathcal{J}}(\pmb\theta)$ is a sample average estimator for $\nabla  \mathcal{J}(\pmb\theta)$, it is unbiased.
%\begin{lemma}\label{unbiasedness}
%The policy gradient is an unbiased estimator, i.e., $\E[\nabla \widehat{\mathcal{J}}(\pmb\theta)] = \nabla  \mathcal{J}(\pmb\theta)$.
%\end{lemma}

% Lemma EC.2

To simplify the notation, let $\nabla\mathcal{J}(\pmb\theta) = \nabla_{\pmb{\theta}}\mathcal{J}(\pmb\theta)$. Observe that
\begin{equation}
\E[\Vert\nabla \widehat{\mathcal{J}}(\pmb{\theta})-{\nabla \mathcal{J}}(\pmb{\theta})\Vert^2]^{1/2} \leq \E[\Vert\nabla \widehat{\mathcal{J}}(\pmb{\theta})-{\nabla \mathcal{J}}(\pmb{\theta})\Vert^4]^{1/4}\leq \sigma,\label{eq.mid0}
\end{equation}
where the first inequality is obtained by applying Jensen's inequality to the convex function $x\mapsto x^2$, and the second inequality follows by Assumption \ref{assumption 2}.

We then expand $\widehat{g}_c(\pmb\theta)$ and derive
\begin{eqnarray}
\E\left[\left\Vert \widehat{g}_c(\pmb\theta)\right\Vert^2\right]&=&
\E\left[\left\Vert \frac{1}{\eta}\left(\Pi_{\mathbb{C}}\left(\pmb\theta+\eta\nabla\widehat{\mathcal{J}}(\pmb\theta)\right)-\pmb\theta\right)\right\Vert^2\right]
\nonumber\\
&=&\E\left[\frac{1}{\eta^2}\left\Vert \Pi_{\mathbb{C}}\left(\pmb\theta+\eta\nabla\widehat{\mathcal{J}}(\pmb\theta)\right)-\Pi_{\mathbb{C}}(\pmb\theta)\right\Vert^2\right]\nonumber\\
&\leq&\E\left[\frac{1}{\eta^2}\eta^2\left\Vert \nabla\widehat{\mathcal{J}}(\pmb\theta)\right\Vert^2\right] \label{lemma2: eq1}\\
&=& \E\left[\left\Vert \nabla\widehat{\mathcal{J}}(\pmb\theta) - \nabla \mathcal{J}(\pmb\theta) + \nabla \mathcal{J}(\pmb\theta)\right\Vert^2\right]\nonumber\\
&\leq& \E\left[\left(\left\Vert \nabla\widehat{\mathcal{J}}(\pmb\theta) - \nabla \mathcal{J}(\pmb\theta)\right\Vert + \left\Vert\nabla \mathcal{J}(\pmb\theta)\right\Vert\right)^2\right]\nonumber\\
&\leq& \E\left[\left(2\max\left\{\left\Vert \nabla\widehat{\mathcal{J}}(\pmb\theta) - \nabla \mathcal{J}(\pmb\theta)\right\Vert, \left\Vert\nabla \mathcal{J}(\pmb\theta)\right\Vert\right\}\right)^2\right]\nonumber\\
&\leq& 4\E\left[\max\left\{\left\Vert \nabla\widehat{\mathcal{J}}(\pmb\theta) - \nabla \mathcal{J}(\pmb\theta)\right\Vert^2, \left\Vert\nabla \mathcal{J}(\pmb\theta)\right\Vert^2\right\}\right]\nonumber\\
&\leq& 4\E\left[\Vert \nabla\widehat{\mathcal{J}}(\pmb\theta) - \nabla \mathcal{J}(\pmb\theta )\Vert^2+\left\Vert \nabla \mathcal{J}(\pmb\theta)\right\Vert^2\right] \nonumber\\
& \leq
& 4\E\left[\left\Vert \nabla \mathcal{J}(\pmb\theta)\right\Vert^2\right]+ 4\sigma^2 \label{lemma2: eq3}
\end{eqnarray}
where \eqref{lemma2: eq1} follows from the contraction property of the projection operator, and \eqref{lemma2: eq3} is obtained by applying (\ref{eq.mid0}).

\subsection{Proof of Lemma~\ref{lemma: cross grdients}}\label{appendix sec: lemma cross grdients}

First, we derive
\begin{eqnarray}
     \left(\E\left[\Vert\nabla\widehat{\mathcal{J}}(\pmb\theta_k)\Vert^4 \right]\right)^{1/4} &=&  \left(\E\left[\left\Vert \nabla\widehat{\mathcal{J}}(\pmb\theta_k) - \nabla \mathcal{J}(\pmb\theta_k) + \nabla \mathcal{J}(\pmb\theta_k)\right\Vert^4\right]\right)^{1/4} \nonumber\\
    %  &\leq& \left(\E\left[\E\left[\left\Vert \nabla\widehat{\mathcal{J}}(\pmb\theta_k) - \nabla \mathcal{J}(\pmb\theta_k) \right\Vert^4+\left\Vert \nabla \mathcal{J}(\pmb\theta_k)\right\Vert^4 \Big| \mathcal{F}_k\right]\right]\right)^{1/4}\label{lemma4: eq1-1} \\
    &\leq& \left(\E\left[\left\Vert \nabla\widehat{\mathcal{J}}(\pmb\theta_k) - \nabla \mathcal{J}(\pmb\theta_k) \right\Vert^4\right]\right)^{1/4}+ \left(\E\left[\left\Vert \nabla \mathcal{J}(\pmb\theta_k)\right\Vert^4 \right]\right)^{1/4} \label{lemma4: eq1-1}\\
     &\leq& \left(\E\left[\Vert \nabla  \mathcal{J}(\pmb\theta_k)\Vert^4\right]\right)^{1/4}+\sigma\label{lemma4: eq1-2}
\end{eqnarray}
where \eqref{lemma4: eq1-1} follows from Minkowski's inequality, and \eqref{lemma4: eq1-2} is due to Assumption~\ref{assumption 2}.

We use the following additional notation. Let $\mathcal{F}_k= \sigma(\pmb\theta_1, \pmb{w}_1,\ldots,\pmb{w}_{k-1})$ represent the $\sigma$-algebra generated by $\pmb\theta_1, \pmb{w}_1,\ldots, \pmb{w}_{k-1}$, where each $\pmb{w}_i = \left\{\pmb{w}^{(b)}_i\right\}^B_{b=1}$. Note that $\mathcal{F}_k$ is based on the process model parameters collected through the $(k-1)$th iteration, while $\widehat{\mathcal{J}}(\pmb\theta_k)$ is estimated by using the new process models $\pmb{w}_k$ obtained in the $k$th iteration.

In the following, we use the form (\ref{eq:sawithg}) of the update. We first consider the conditional expectation with respect to $\mathcal{F}_k$, and calculate
\begin{eqnarray}
&\,& \E\left[\left\langle \nabla \mathcal{J}(\pmb\theta_k),\eta_k \widehat{g}_{c}(\pmb\theta_k)\right\rangle|\mathcal{F}_k\right]\nonumber\\
&=&\E\left[\left\langle \nabla \mathcal{J}(\pmb\theta_k),\eta_k \widehat{g}_{c}(\pmb\theta_k) - \eta_k \nabla \mathcal{J}(\pmb\theta_k)\right\rangle|\mathcal{F}_k\right] + \eta_k\Vert \nabla \mathcal{J}(\pmb\theta_k)\Vert^2\nonumber\\
% &=& \langle \nabla \mathcal{J}(\pmb\theta_k),\eta_k \E[\widehat{g}_{c}(\pmb\theta_k)|\pmb\tau_{[k-1]}] - \eta_k \nabla \mathcal{J}(\pmb\theta_k)\rangle + \eta_k\Vert \nabla \mathcal{J}(\pmb\theta_k)\Vert^2_2 \nonumber\\
&=&\E\left[\left\langle \nabla \mathcal{J}(\pmb\theta_k), \Pi_{\mathbb{C}}\left(\pmb\theta_k+\eta_k\nabla\widehat{\mathcal{J}}(\pmb\theta_k)\right) -\pmb\theta_k -\eta_k\nabla\widehat{\mathcal{J}}(\pmb\theta_k) +\eta_k\nabla\widehat{\mathcal{J}}(\pmb\theta_k) - \eta_k \nabla \mathcal{J}(\pmb\theta_k)\right\rangle\Big|\mathcal{F}_k\right] + \eta_k\Vert \nabla \mathcal{J}(\pmb\theta_k)\Vert^2 \nonumber\\
&=&\E\left[\left\langle \nabla \mathcal{J}(\pmb\theta_k), \Pi_{\mathbb{C}}\left(\pmb\theta_k+\eta_k\nabla\widehat{\mathcal{J}}(\pmb\theta_k)\right) -\pmb\theta_k -\eta_k\nabla\widehat{\mathcal{J}}(\pmb\theta_k)\right\rangle\Big|\mathcal{F}_k \right]
\nonumber\\
&&+\E\left[\langle \nabla \mathcal{J}(\pmb\theta_k),\eta_k\nabla\widehat{\mathcal{J}}(\pmb\theta_k) - \eta_k \nabla \mathcal{J}(\pmb\theta_k)\rangle\Big|\mathcal{F}_k \right]
+\eta_k\Vert \nabla \mathcal{J}(\pmb\theta_k)\Vert^2 \nonumber\\
&=&\E\left[\left\langle \nabla \mathcal{J}(\pmb\theta_k), \Pi_{\mathbb{C}}\left(\pmb\theta_k+\eta_k\nabla\widehat{\mathcal{J}}(\pmb\theta_k)\right) -\pmb\theta_k -\eta_k\nabla\widehat{\mathcal{J}}(\pmb\theta_k)\right\rangle\Big|\mathcal{F}_k \right]
\nonumber\\
&&+\left\langle \nabla \mathcal{J}(\pmb\theta_k),\E\left[\eta_k\nabla\widehat{\mathcal{J}}(\pmb\theta_k) - \eta_k \nabla \mathcal{J}(\pmb\theta_k)\Big|\mathcal{F}_k \right]\right\rangle
+\eta_k\Vert \nabla \mathcal{J}(\pmb\theta_k)\Vert^2
\nonumber\\
&=&\E\left[\left\langle \nabla \mathcal{J}(\pmb\theta_k), \Pi_{\mathbb{C}}\left(\pmb\theta_k+\eta_k\nabla\widehat{\mathcal{J}}(\pmb\theta_k)\right) -\pmb\theta_k -\eta_k\nabla\widehat{\mathcal{J}}(\pmb\theta_k)\right\rangle\Big|\mathcal{F}_k \right] +
\eta_k\Vert \nabla \mathcal{J}(\pmb\theta_k)\Vert^2
\label{lemma 3: eq 0}\\
&=&-\Big\langle \nabla \mathcal{J}(\pmb\theta_k), \E\left[ \left(\pmb\theta_k +\eta_k\nabla\widehat{\mathcal{J}}(\pmb\theta_k)-\Pi_{\mathbb{C}}\left(\pmb\theta_k+\eta_k\nabla\widehat{\mathcal{J}}(\pmb\theta_k)\right) \right)1_{A_k}\Big|\mathcal{F}_k\right]
% \left(1- p(A_k)\right)
\nonumber\\
&& + \E\left[ \left( \pmb\theta_k +\eta_k\nabla\widehat{\mathcal{J}}(\pmb\theta_k)-
\Pi_{\mathbb{C}}\left(\pmb\theta_k+\eta_k\nabla\widehat{\mathcal{J}}(\pmb\theta_k)\right)\right)1_{A_k^c}\Big|\mathcal{F}_k \right]\Big\rangle +
\eta_k\Vert \nabla \mathcal{J}(\pmb\theta_k)\Vert^2 \nonumber\\
&=&-\Big\langle \nabla \mathcal{J}(\pmb\theta_k), \E\left[\left( \pmb\theta_k +\eta_k\nabla\widehat{\mathcal{J}}(\pmb\theta_k)-\Pi_{\mathbb{C}}\left(\pmb\theta_k+\eta_k\nabla\widehat{\mathcal{J}}(\pmb\theta_k)\right)\right)1_{A_k^c}\Big|\mathcal{F}_k\right] \Big\rangle +
\eta_k\Vert \nabla \mathcal{J}(\pmb\theta_k)\Vert^2 \label{lemma 3: eq 0-1}\\
&=&-\E\left[1_{A_k^c}\Big\langle \nabla \mathcal{J}(\pmb\theta_k),  \pmb\theta_k +\eta_k\nabla\widehat{\mathcal{J}}(\pmb\theta_k)-\Pi_{\mathbb{C}}\left(\pmb\theta_k+\eta_k\nabla\widehat{\mathcal{J}}(\pmb\theta_k)\right)\Big\rangle\Big|\mathcal{F}_k  \right] +
\eta_k\Vert \nabla \mathcal{J}(\pmb\theta_k)\Vert^2.
\label{eq.mid6}
\end{eqnarray}
Above, \eqref{lemma 3: eq 0} follows from the fact that $\E\left[\nabla \widehat{\mathcal{J}}(\pmb\theta_k)\right | \mathcal{F}_k] = \nabla \mathcal{J}(\pmb\theta_k)$ because $\widehat{\mathcal{J}}$ is a sample average, and \eqref{lemma 3: eq 0-1} holds because
\begin{equation*}
\E\left[\left( \pmb\theta_k +\eta_k\nabla\widehat{\mathcal{J}}(\pmb\theta_k)-\Pi_{\mathbb{C}}\left(\pmb\theta_k+\eta_k\nabla\widehat{\mathcal{J}}(\pmb\theta_k)\right)\right)1_{A_k}\Big|\mathcal{F}_k\right]= 0.
\end{equation*}
Notice that $1_{A_k^c}=1_{A_k^c}^2$. Then, taking unconditional expectations of the preceding quantities, (\ref{eq.mid6}) yields
\begin{eqnarray}
&\,&\E\left[\left\langle \nabla \mathcal{J}(\pmb\theta_k),\eta_k \widehat{g}_{c}(\pmb\theta_k)\right\rangle\right]\nonumber\\
&=& -\E\left[1_{A_k^c}\Big\langle \nabla \mathcal{J}(\pmb\theta_k),  \pmb\theta_k +\eta_k\nabla\widehat{\mathcal{J}}(\pmb\theta_k)-\Pi_{\mathbb{C}}\left(\pmb\theta_k+\eta_k\nabla\widehat{\mathcal{J}}(\pmb\theta_k)\right)\Big\rangle  \right] +
\eta_k \E\left[\Vert \nabla \mathcal{J}(\pmb\theta_k)\Vert^2\right]\nonumber\\
&=&-\E\left[\Big\langle1_{A_k^c} \nabla \mathcal{J}(\pmb\theta_k), 1_{A_k^c}\left( \pmb\theta_k +\eta_k\nabla\widehat{\mathcal{J}}(\pmb\theta_k)-\Pi_{\mathbb{C}}\left(\pmb\theta_k+\eta_k\nabla\widehat{\mathcal{J}}(\pmb\theta_k)\right) \right)\Big\rangle  \right] +
\eta_k \E\left[\Vert \nabla \mathcal{J}(\pmb\theta_k)\Vert^2\right]\nonumber\\
&\geq&-\left(\E\left[1_{A_k^c}\Vert \nabla \mathcal{J}(\pmb\theta_k)\Vert^2\right]\right)^{1/2}\left(\E\left[1_{A_k^c} \left \Vert\Pi_{\mathbb{C}}\left(\pmb\theta_k+\eta_k\nabla\widehat{\mathcal{J}}(\pmb\theta_k)\right) -\pmb\theta_k -\eta_k\nabla\widehat{\mathcal{J}}(\pmb\theta_k)
\right \Vert^2 \right]\right)^{1/2}  \nonumber\\
&&+
\eta_k \E\left[\Vert \nabla \mathcal{J}(\pmb\theta_k)\Vert^2\right] \label{lemma 3: eq 1}\\
&\geq&-\E[1_{A_k^c}]^{1/4} \left(\E\left[\Vert \nabla \mathcal{J}(\pmb\theta_k)\Vert^4\right]\right)^{1/4}\E[1_{A_k^c}]^{1/4}\left(\E\left[ \left\Vert\Pi_{\mathbb{C}}\left(\pmb\theta_k+\eta_k\nabla\widehat{\mathcal{J}}(\pmb\theta_k)\right) -\pmb\theta_k -\eta_k\nabla\widehat{\mathcal{J}}(\pmb\theta_k) \right\Vert^4 \right]\right)^{1/4}  \nonumber\\
&&+
\eta_k \E\left[\Vert \nabla \mathcal{J}(\pmb\theta_k)\Vert^2\right] \label{lemma 3: eq 1-2}\\
&\geq&- \E[1_{A_k^c}]^{1/2} \left(\E\left[\Vert \nabla \mathcal{J}(\pmb\theta_k)\Vert^4\right]\right)^{1/4} \left(\E\left[\Vert\eta_k\nabla\widehat{\mathcal{J}}(\pmb\theta_k)\Vert^4 \right]\right)^{1/4} +
\eta_k \E\left[\Vert \nabla \mathcal{J}(\pmb\theta_k)\Vert^2\right] \label{lemma 3: eq 2}\\
&\geq&-\left(1- p(A_k)\right)^{1/2}\eta_k \left(\E\left[\Vert \nabla \mathcal{J}(\pmb\theta_k)\Vert^4\right]\right)^{1/4}\left( \left(\E\left[\Vert \nabla \mathcal{J}(\pmb\theta_k)\Vert^4\right]\right)^{1/4} + \sigma\right) +
\eta_k \E\left[\Vert \nabla \mathcal{J}(\pmb\theta_k)\Vert^2\right] \label{lemma 3: eq 3}\\
&\geq&-\left(1-p(A_k)\right)^{1/2}\eta_k(G^2+G\sigma)+
\eta_k \E\left[\Vert \nabla \mathcal{J}(\pmb\theta_k)\Vert^2\right] \label{lemma 3: eq 4}\\
&\geq&
-\eta_k(G^2+G\sigma)\left(1-p(A_k)\right)^{1/2} +\eta_k\E\left[\Vert \nabla \mathcal{J}(\pmb\theta_k)\Vert^2\right].\nonumber
\end{eqnarray}

Above, \eqref{lemma 3: eq 1} and \eqref{lemma 3: eq 1-2} follow by applying Holder's inequality. Then, the inequality \eqref{lemma 3: eq 2} follows by applying the second property of the projection in Proposition~\ref{prop: projection}. Notice that $\pmb\theta_k \in {\mathbb{C}}$.
The inequality \eqref{lemma 3: eq 3} follows from \eqref{lemma4: eq1-2}, and, finally, \eqref{lemma 3: eq 4} is obtained by applying Corollary \ref{corollary: bounded gradient}.

\end{appendices}
\end{document}